\documentclass[9pt,twocolumn,twoside]{ilcss}
\usepackage[
  locale=US,
  separate-uncertainty=true,
  per-mode=symbol-or-fraction,
  group-separator={,},
  group-minimum-digits=4,
]{siunitx}
\usepackage{makecell} 
\usepackage{graphicx,multirow} 
\usepackage{colortbl} 
\usepackage{bm} 
\usepackage{url}  
\usepackage{siunitx} 
\usepackage{caption} 
\usepackage{float} 
\usepackage{threeparttable}
\usepackage{tabularx}
\usepackage{booktabs} 
\usepackage{hyperref}
\usepackage{subcaption}
\usepackage{placeins} 
\usepackage{mwe} 

\DeclareSIUnit{\kgCOeq}  {kg\,CO_2\,eq.}

\templatetype{ilcssworkingpaper} 

\title{CellViT++: Energy-Efficient and Adaptive Cell Segmentation and Classification Using Foundation Models}	

\author[a,b,c,1]{Fabian Hörst}
\author[a,b,c]{Moritz Rempe}
\author[a,b]{Helmut Becker}
\author[a,b]{Lukas Heine}
\author[a,d]{Julius Keyl}
\author[a,b,c,e]{Jens Kleesiek}

\affil[a]{\small Institute for AI in Medicine (IKIM), University Hospital Essen (AöR), Essen, Germany}
\affil[b]{Cancer Research Center Cologne Essen (CCCE), West German Cancer Center Essen, University Hospital Essen (AöR), Essen, Germany}
\affil[c]{Department of Physics, TU Dortmund University, Dortmund, Germany}
\affil[d]{Institute of Pathology, University Hospital Essen (AöR), Essen, Germany}
\affil[e]{German Cancer Consortium (DKTK, Partner site Essen), Heidelberg, Germany}
\leadauthor{Hörst} 

\correspondingauthor{\textsuperscript{1}Corresponding Author: fabian.hoerst@uk-essen.de}

\vspace{-2mm}

\begin{abstract}
Digital Pathology is a cornerstone in the diagnosis and treatment of  diseases. A key task in this field is the identification and segmentation of cells in hematoxylin and eosin-stained images. Existing methods for cell segmentation often require extensive annotated datasets for training and are limited to a predefined cell classification scheme. To overcome these limitations, we propose $\text{CellViT}^{{\scriptscriptstyle ++}}$, a framework for generalized cell segmentation in digital pathology. $\text{CellViT}^{{\scriptscriptstyle ++}}$ utilizes Vision Transformers with foundation models as encoders to compute deep cell features and segmentation masks simultaneously. To adapt to unseen cell types, we rely on a computationally efficient approach. It requires minimal data for training and leads to a drastically reduced carbon footprint. We demonstrate excellent performance on seven different datasets, covering a broad spectrum of cell types, organs, and clinical settings. The framework achieves remarkable zero-shot segmentation and data-efficient cell-type classification. Furthermore, we show that $\text{CellViT}^{{\scriptscriptstyle ++}}$ can leverage immunofluorescence stainings to generate training datasets without the need for pathologist annotations. The automated dataset generation approach surpasses the performance of networks trained on manually labeled data, demonstrating its effectiveness in creating high-quality training datasets without expert annotations. To advance digital pathology, $\text{CellViT}^{{\scriptscriptstyle ++}}$ is available as an open-source framework featuring a user-friendly, web-based interface for visualization and annotation. The code is available under \url{https://github.com/TIO-IKIM/CellViT-plus-plus}. 
\end{abstract}

\doi{\url{https://github.com/TIO-IKIM/CellViT-plus-plus}}

\begin{document}

\maketitle
\thispagestyle{firststyle}
\ifthenelse{\boolean{shortarticle}}{\ifthenelse{\boolean{singlecolumn}}{\abscontentformatted}{\abscontent}}{}
\vspace{-3mm}
Histopathology serves as the "workhorse" of medical diagnostics, playing a crucial role in the identification and classification of diseases. Through the systematic examination of tissue and cellular structures, pathologists generate important insights that inform clinical decision-making and influence treatment plans for various conditions, including cancer, infections, autoimmune disorders, genetic disorders, neurodegenerative diseases, cardiovascular diseases, or transplant rejection. For routine diagnostics, tissue samples are collected and stained with hematoxylin and eosin (H\&E) to visualize cellular and tissue structures. The digitization of these tissue specimens through whole slide imaging (WSI) has introduced a significant advancement in pathology by enabling computational assessments~\citep{Song2023}. This development facilitates the integration of artificial intelligence (AI) into diagnostic processes, enhancing existing workflows and slide quantification while advancing research and biomarker discovery. Consequently, more and more laboratories are shifting from traditional manual workflows to more efficient and scalable digitized workflows~\citep{recommendation_digital_workflow, dp_workflow}. This process will be further expedited by the implementation of reimbursement strategies for medical AI solutions~\citep{reimbursement1, reimbursement2}.

\begin{figure*}[!b]
    \centering
    \includegraphics[width=\linewidth]{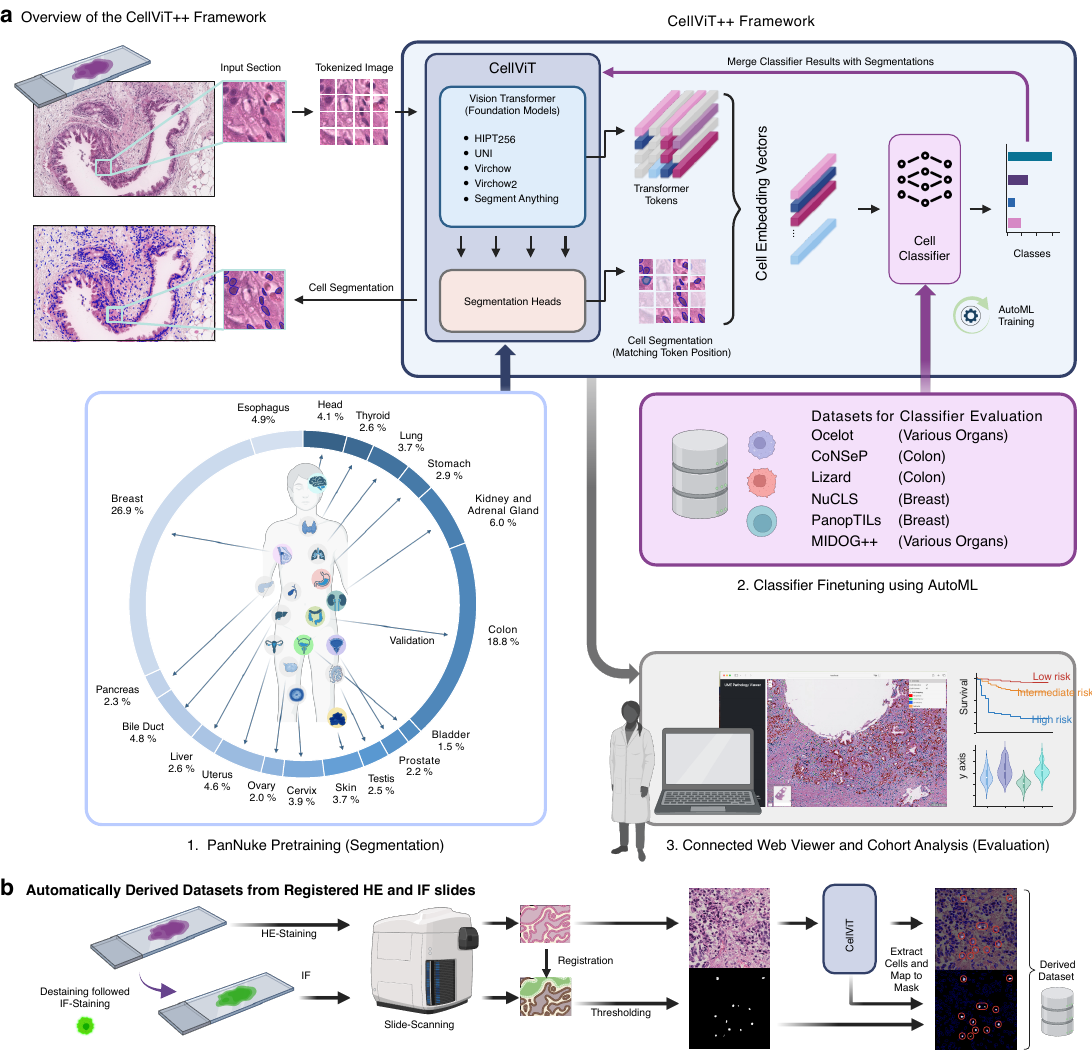}
    \caption{\textbf{Overview of the $\text{CellViT}^{{\scriptscriptstyle ++}}$ Framework}. \textbf{a} Network architecture including the newly introduced cell classification module based on cell embeddings which are equal to the Transformer tokens of the last Transformer Block. Cell embedding vectors can be extracted in the forward pass in conjunction with the segmentation process. The embeddings are subsequently used to train a cell type classification module, to adapt the framework to new cell classes. The segmentation network of $\text{CellViT}^{{\scriptscriptstyle ++}}$ is pretrained using the PanNuke dataset (1). Tissue types highlighted in bold are selected for further analysis in this study. Subsequent classification modules are trained on unseen datasets (2) and the results are combined with the segmentation masks. Cohort analysis and image visualization can be performed with our web-based viewer (3).  \textbf{b} Pipeline to automatically derive labels from registered H\&E and IF scans using $\text{CellViT}^{{\scriptscriptstyle ++}}$, exemplified by the SegPath dataset.}
    \label{fig:overview_image}
    \phantomsubcaption
    \label{fig:overview_image_a}
    \phantomsubcaption
    \label{fig:overview_image_b}
\end{figure*}

Among the various computational tasks in digital pathology, cell detection and classification in WSIs are particularly critical, as they can reveal prognostic factors. For example, tumor-infiltrating lymphocytes and inflammatory cells within the tumor microenvironment are important markers for breast cancer~\citep{tils, report_comp_til_assessment}. This task is not feasible for pathologists at a large scale due to its time-consuming nature and the high degree of intra- and inter-observer variability~\citep{CellViT}. Next to cellular detection, the precise identification of cell boundaries is essential for cellular feature analysis. Due to inconsistencies in cell morphology, staining intensity, and the presence of overlapping or touching nuclei, this task remains challenging. Deep learning methods based on convolutional neural networks (CNNs)~\citep{dist,hovernet, cellpose, stardist} and Vision Transformers (ViTs)~\citep{CellViT} have emerged as powerful tools in this context. In addition to algorithmic advancements, the digitization of pathology has led to the creation of increasingly large-scale slide datasets, predominantly by non-public institutions. This has facilitated the development of "foundation models" that are designed to learn generalizable representations from extensive amounts of WSI. Unlike specialized task-specific models, foundation models aim to serve as generalist models capable of addressing a wide range of tasks based on learned representations. Several foundation models have been proposed for digital pathology, including HIPT/$\text{HIPT}_{256}$~\citep{hipt}, UNI~\citep{uni}, Virchow~\citep{virchow}, $\text{Virchow}_2$~\citep{virchow2}, RudolfV~\citep{rudolfV}, and Gigapath~\citep{gigapath}, with the most promising models utilizing the ViT~\citep{vit16x16} network architecture. Complementing,  the Segment Anything Model (SAM)~\citep{sam} for segmenting natural images, has also shown remarkable performance in medical imaging~\citep{sam_biomed_applications}.
Building on the success of foundation models, we previously developed the \mbox{CellViT} model~\citep{CellViT}, which combines ViT based foundation models with a segmentation head. Using this structure, we outperformed existing methods and achieved state-of-the-art (SOTA) results for cell detection and segmentation~\citep{CellViT}. The \mbox{CellViT} model has been validated in several independent studies. 
An adaption of \mbox{CellViT}, incorporating a tissue segmentation branch, won the Ocelot Challenge~\citep{cellvitocelotwinner}. In CD20+ B cell quantification for lung tissue, \mbox{CellViT} effectively detected nuclei in CD20-stained sections, revealing associations between B cell clusters and granulomas in mice with M. tuberculosis infection~\citep{b-cells-cellvit}. \mbox{CellViT} also outperformed methods like StarDist and CellPose in analyzing acute lymphoblastic leukemia and kidney samples~\citep{cellvit-leukemia, cellvit-kidney}. Additionally, applying \mbox{CellViT} to the HEST1K dataset~\citep{Hest1k} demonstrates its ability to integrate molecular and morphological information, emphasizing the need for robust segmentation algorithms, as shown for the GATA3 concentration in heterogeneous cancer regions~\citep{Hest1k}. \\
\indent Despite advancements in foundation models and cell segmentation algorithms, no model has yet demonstrated generalizability for arbitrary cell types or sufficient adaptability without re-training. Current methods depend on extensive annotated datasets and precise segmentation masks. The PanNuke dataset, with 19 organ types and approximately $\num{190000}$ annotated cells, has been a valuable resource for cell segmentation but its utility is limited to the five specific cell types it includes: neoplastic, inflammatory, epithelial, connective, and dead cells. However, since different tissue and tumor types require diverse cell classification schemes, a dynamic adaptability in class assignment for detected cells is necessary. For instance, the group of inflammatory cells encompasses various immune cell types, such as lymphocytes, neutrophils, eosinophils, and macrophages, with lymphocytes being particularly relevant in breast cancer prognosis~\citep{tils, panoptils}. The standard approach for a new cell-class scheme is to create a corresponding dataset with several thousand annotated cells by pathologists to develop a model for the cell types of interest based on existing methods~\citep{hovernet,cellpose,stardist}. Creating expert-level cell segmentation annotations is more time-consuming and costly than creating a cell detection dataset without cell contours~\citep{Reiss_2021_CVPR}, limiting new network candidates to detection-based models. An out-of-the-box algorithmic solution like the nnUNet for radiological images is still absent for digital pathology~\citep{nnunet}.\\
\indent To solve this problem, we propose $\text{CellViT}^{{\scriptscriptstyle ++}}$, a framework designed for generalized cell segmentation in H\&E-stained images (Fig.~\ref{fig:overview_image}). Building upon the previously published \mbox{CellViT} architecture, our approach leverages the representation learning capacity of foundation models and the inherent structure of the Transformer architecture of the image encoder. During the forward pass through the \mbox{CellViT} model, deep cell features are computed alongside the segmentation masks without additional computational overhead (Fig.~\ref{fig:overview_image_a}). These cell embeddings are then used to build segmentation agnostic cell type classification modules to adapt the network to new cell types. This bypasses the traditional requirement for separate cell cropping and feature extraction stages seen in two-stage models~\citep{nucleiio}.
A key aspect of the $\text{CellViT}^{{\scriptscriptstyle ++}}$ framework is the robust segmentation performance achieved by the segmentation heads, which allows us to focus solely on re-training the lightweight cell classification module to achieve accurate cell classification and segmentation across different cell types. We extend the original \mbox{CellViT} model by (1) incorporating several foundation models as image encoders, (2) extracting cell tokens for each detected cell, (3) introducing a lightweight cell classification module based on the cell tokens to swiftly adapt to new cell classification schemes, (4) optimizing the code for faster inference speed, and (5) providing an entire toolbox including a web-based visualization and annotation tool (Fig.~\ref{fig:overview_image_a}). To evaluate the efficiency of the classification approach, we show that integrating the classification modules at the bottleneck layer (last Transformer layer) of the encoder provides an effective method for adapting the model to new classes. We evaluate $\text{CellViT}^{{\scriptscriptstyle ++}}$ using different foundation models as image encoders to assess their ability to generate discriminative cell embeddings. We included, in order of increasing parameter count, $\text{HIPT}_{256}$, UNI, Virchow , $\text{Virchow}_2$, as well as the domain-agnostic model Segment Anything Model (SAM-H). All $\text{CellViT}^{{\scriptscriptstyle ++}}$ variants, respectively the segmentation models, have been pretrained on the heterogeneous PanNuke dataset (Fig.~\ref{fig:overview_image_a}). We utilize a wide range of cell datasets to test the cell classification module, including two multi-organ datasets (Ocelot~\citep{ocelot}, MIDOG++~\citep{midogpp}), that encompass organs not included in the PanNuke training data. Additionally, we assess CellViT’s performance on two common cancer types: breast cancer (NuCLS~\citep{nucls}, SegPath~\citep{segpath}, PanopTILs~\citep{panoptils}) and colorectal cancer (CoNSeP~\citep{hovernet}, Lizard~\citep{lizard}) and compare our method with current baseline methods on these datasets. Our method not only achieves outstanding performance but also yields remarkable results with significantly fewer samples while requiring substantially less energy resources ($\text{CO}_2$ equivalent) due to its efficient feature computation compared to all other methods. 

Additionally, we also demonstrate that the $\text{CellViT}^{{\scriptscriptstyle ++}}$ framework can achieve exceptional performance on unknown cell types without the need for pathologist-curated datasets. We employ the SegPath dataset~\citep{segpath}, which includes registered pairs of H\&E and immunofluorescent (IF) stainings from the same tissue sections. The IF stainings, which use specific antibodies tagged with fluorescent markers to highlight cellular components, enable the automatic generation of segmentation masks for the corresponding H\&E sections through thresholding techniques. Although these masks initially lack instance-level segmentations, the $\text{CellViT}^{{\scriptscriptstyle ++}}$ model can detect and segment single cells in the H\&E samples and transfer these segmentations to the IF masks. This approach facilitates the creation of large-scale cell datasets without needing expert pathologist annotations. A schematic of this method is illustrated in Fig.~\ref{fig:overview_image_b}. We exemplify this approach by generating datasets for lymphocytes and plasma cells in breast tissue. Classifiers trained on these automatically generated datasets achieved nearly the same performance as those trained on pathologist-annotated datasets for lymphocytes and even surpass them for plasma cells.

To democratize the use of $\text{CellViT}^{{\scriptscriptstyle ++}}$ as a high-throughput pipeline, we have developed and released it as an open-source toolbox that does not require specialized high-performance computing clusters or coding expertise. The toolbox automates hyperparameter tuning, functioning as a type of \mbox{AutoML} tool~\citep{automl_healthcare}. Additionally, we provide a new lightweight, web-based WSI viewer for visualizing detection/segmentation results, which offers greater accessibility and ease of use compared to traditional local software solutions. To minimize annotation efforts for pathologists, the toolbox includes a web-based annotation tool that facilitates quick reclassification and relabeling for cells. All trained classification modules are integrated into our framework and can be used out of the box. The toolbox is available at \href{https://github.com/TIO-IKIM/CellViT-plus-plus}{https://github.com/TIO-IKIM/CellViT-plus-plus}. To avoid confusion, CellViT refers to the segmentation model from our previous work, pretrained exclusively on PanNuke, while $\text{CellViT}^{{\scriptscriptstyle ++}}$ denotes the new framework introduced in this study, which integrates the pretrained CellViT model with a cell classification module.

\section*{Results}

\begin{figure*}[!b]
    \centering
    \includegraphics[width=\linewidth]{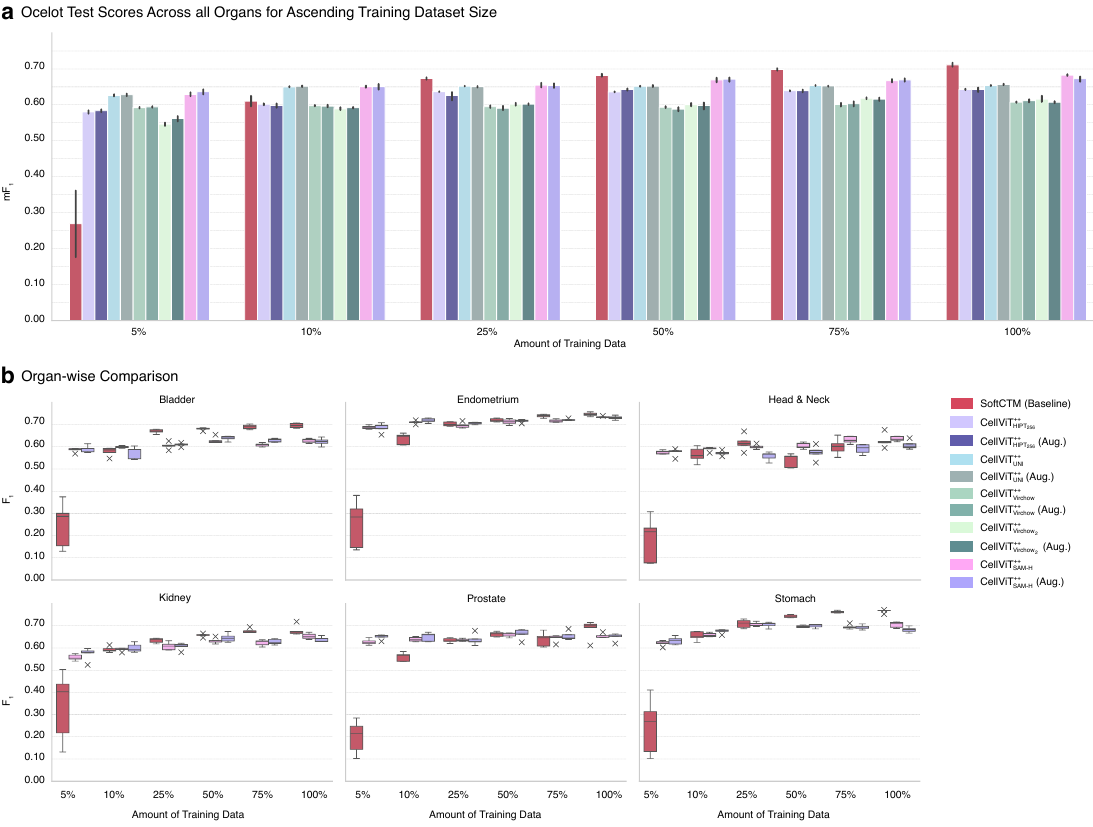}
    \caption{
    \textbf{Ocelot results and comparison with state-of-the-art baseline network SoftCTM}. 
    \textbf{a} Mean $F_1$-Score averaged over all tissue types in the dataset on the official test set for multiple image encoders. 
    Each cell classification module and the baseline SoftCTM model were trained on a limited amount of training data. 
    Results are given for 5 experiments with different seeds. 
    \textbf{b} Organ-wise detection results of the baseline SoftCTM model in comparison to the best performing $\text{CellViT}^{{\scriptscriptstyle ++}}$ model, 
    again trained on limited dataset sizes.}
    \label{fig:ocelot_results}  
    \phantomsubcaption
    \label{fig:ocelot_results_a}
    \phantomsubcaption
    \label{fig:ocelot_results_b}
    \vspace{-3mm}
\end{figure*}

\subsection*{PanNuke Pretraining Results\\}
A fundamental component of our framework is the pretraining of the segmentation models on the PanNuke dataset, which contains $\num{190000}$ extensively annotated cells across 19 tissue types. These pretrained models (CellViT) are then integrated into the $\text{CellViT}^{{\scriptscriptstyle ++}}$ framework without further modification. Before evaluating the $\text{CellViT}^{{\scriptscriptstyle ++}}$ framework, it is essential to demonstrate that the segmentation models achieve satisfactory performance. To this end, we evaluated all models using the official 3-fold cross-validation (CV) split of the PanNuke dataset~\citep{pannuke}. The primary metric employed is the mean Panoptic Quality (mPQ), as it combines segmentation and detection performance into a single score while also considering cell classes.  In addition to the already published $\text{CellViT}_{\text{HIPT}_{256}}$ and $\text{CellViT}_{\text{SAM-H}}$ models, we included three additional histopathological foundation models (UNI, Virchow, and $\text{Virchow}_2$). The results reveal that all \mbox{CellViT} variants with a foundation model encoder exhibited superior performance compared to the baseline model $\text{CellViT}_{\text{ViT-S}}$ (vanilla ViT small) with an average improvement of at least 7.45\% over the baseline ($\text{CellViT-ViT}_\text{s}:0.4417$~\si{mPQ}, $\text{Standard Deviation (SD)}~0.0500$ vs. $\text{CellViT-HIPT}_{256}: 0.4846~\si{mPQ}$, $\text{SD}~0.0503$). Among the evaluated models, the Segment Anything Model ($\text{CellViT}_{\text{SAM-H}}$) achieved the highest performance with $0.49803~\si{mPQ}$ ($\text{SD}~0.041$), benefiting from its extensive segmentation pretraining task. A full overview about all results are given in the~\ref{fig:datset_overview_pannuke_results}. While all models showed promising cell detection and segmentation results, our subsequent analyses will focus on identifying the encoder yielding the most generalistic cell representations. All subsequent \mbox{CellViT} segmentation models included in the $\text{CellViT}^{{\scriptscriptstyle ++}}$ framework have been pretrained on 95\% patches of the PanNuke dataset, with the remaining 5\% used to detect overfitting. If not otherwise stated, each cell classification module consists of one hidden layer with ReLU activation function. As illustrated in Fig.~\ref{fig:overview_image_a}, we assigned each identified cell the corresponding token of the ViT-based image encoder, and trained the classifier solely based on the tokens.

\subsection*{Data-Efficient Cancer Cell Detection Across Multiple Organs Using $\text{CellViT}^{{\scriptscriptstyle ++}}$\\}

Although the PanNuke dataset encompasses a multitude of organs and includes the cell category neoplastic cells, differentiating malignant and benign cells within the neoplastic cell group is a key objective in routine histopathological diagnostics. The Ocelot dataset contains a total of $\num{71691}$ malignant tumor cells and $\num{41335}$ non-tumor cells annotated in 663 $num{1024} \times num{1024}~\si{pixel}~(\si{px})$ image sections including six human organs, namely bladder, endometrium, head/neck, kidney, prostate, and stomach (see~\ref{fig:datset_overview_multi_tissue}). To obtain the mean and standard deviation, each experiment was repeated five times on the train and validation split, with evaluation on the official test set. The metric utilized is the mean $\text{F}_1$-score ($\text{mF}_1$). The dataset also includes area segmentations of tumor tissue, intended to serve as an additional aid for cell classification. Only the subset containing cell annotations is utilized in this current study. As demonstrated by~\citet{cellvitocelotwinner}, the combination of \mbox{CellViT} with a tumor segmentation model (ensemble) currently yielded SOTA results with a $\text{mF}_1$-score of 0.7243 on this data set. Other comparative methods include a ResNet ensemble by~\citet{Lafarge2024ocelot} ($0.6617~\text{mF}_1$), the FC-HarDNet model ($0.6992~\text{mF}_1$)~\citep{fchardnetocelot}, and the model by~\citet{ocelot_millward} ($0.7221~\text{mF}_1$). All these models consist of at least two models for cell and tissue segmentation followed by a merging strategy to fuse tissue segmentations with the cell segmentations. In contrast, the cell-only baseline just reached $0.6444~\text{mF}_1$~\citep{ocelot}. \\
For comparison with our model, we used the CNN-based SoftCTM model (tissue and cell), which has been externally validated and reported to achieve a performance of 0.7172 $\text{mF}_1$-score. We selected this architecture because it does not rely on \mbox{CellViT} as a cell segmentation model. We re-trained the SoftCTM model five times and report the average results along with the standard deviation. On the external test set, SoftCTM achieved $0.7109~\text{mF}_1$ ($\text{SD}~0.0069$), while the best $\text{CellViT}^{{\scriptscriptstyle ++}}$ variant ($\text{CellViT}^{{\scriptscriptstyle ++}}_\text{SAM-H}$) achieved $0.6827~\text{mF}_1$ ($\text{SD}~0.0028$). The results are particularly noteworthy as $\text{CellViT}^{{\scriptscriptstyle ++}}_\text{SAM-H}$ outperformed the cell-only baseline ($0.6444~\text{mF}_1$) by $\Delta=+0.0383~\si{mF}_1$, despite the fact that only the cell classification module, and not the segmentation model, was re-trained. Furthermore, as a cell-only model, $\text{CellViT}^{{\scriptscriptstyle ++}}_\text{SAM-H}$ yielded a performance close to the supervised baseline of SoftCTM ($\Delta=-0.0282~\si{mF}_1$), which utilized additional tumor-microenvironment tissue context. In contrast, all models using histopathological foundation model encoders performed inferior to the $\text{CellViT}^{{\scriptscriptstyle ++}}_\text{SAM-H}$ model. A table with all results can be found in the Supplement (Table \ref{tab:appendix_ocelot_all_networks}). \\
\indent Data augmentation is a common regularization technique to enhance the generalizability and performance of an algorithm. To evaluate its impact, we augmented the input patches using common augmentation techniques (see Supplementary Table \ref{tab:appendix_data_augmentation}). The results, given in Fig.~\ref{fig:ocelot_results_a}, indicate that data augmentation has a negligible effect on classification performance, with marginal improvements for $\text{CellViT}^{{\scriptscriptstyle ++}}_\text{UNI}$ ($\Delta = + 0.0028~\si{mF}_1$) and $\text{CellViT}^{{\scriptscriptstyle ++}}_\text{Virchow}$ ($\Delta = + 0.0040~\si{mF}_1$), but reduced performance for $\text{CellViT}^{{\scriptscriptstyle ++}}_{\text{Virchow}_2}$ ($\Delta = - 0.0087~\si{mF}_1$) and $\text{CellViT}^{{\scriptscriptstyle ++}}_\text{UNI}$ ($\Delta = - 0.0101~\si{mF}_1$). \\
\indent 
Given the challenge of acquiring labeled WSI data and the difficulty of obtaining thousands of annotated cells with expert-level precision, we also investigated the data efficiency. We sequentially sampled subsets of the training data, comprising 5\%, 10\%, 25\%, 50\%, and 75\% of the training dataset (see \ref{tab:appendix_ocelot_cell_amount} in the Supplement for dataset overview). The results, illustrated in Fig.~\ref{fig:ocelot_results_a} and \ref{fig:ocelot_results_b}, demonstrate that all $\text{CellViT}^{{\scriptscriptstyle ++}}$ variants achieved significantly better performance with only 5\% of the data compared to the reference method. At least 25\% of the training data were necessary (equals to $\num{18573}$ cells) such that SoftCTM was on par. In contrast, the performance of the $\text{CellViT}^{{\scriptscriptstyle ++}}$ models saturated, with a tripling of the training dataset size resulting in only a modest increase in $\text{F}_1$-score for $\text{CellViT}^{{\scriptscriptstyle ++}}_\text{SAM-H}$ (25\% data $0.6550~\text{mF}_1$, $\text{SD}~0.0073$ vs. 75\% data with $0.6669~\text{mF}_1$, $\text{SD}~0.0054$). Additionally, the comparison reveals that the $\text{HIPT}_{256}$, UNI, and SAM-H variants generated embeddings that allow for competitive classification performance, whereas the models of the Virchow series (Virchow, $\text{Virchow}_2$) consistently yielded inferior results on this dataset. There is no demonstrable benefit from using data augmentation even when training data is limited. Additional results for the $\text{CellViT}^{{\scriptscriptstyle ++}}_\text{SAM-H}$ model, along with SoftCTM, are shown in Fig.~\ref{fig:ocelot_results_b}, stratified by tissue type. The box plots indicate that $\text{CellViT}^{{\scriptscriptstyle ++}}$ achieved better and more stable performance with limited training data, as the segmentation decoder serves as a strong backbone for cell detection. For underrepresented tissue types in the Ocelot dataset, such as head and neck, $\text{CellViT}^{{\scriptscriptstyle ++}}$ performed comparably or better than SoftCTM even with all available training data (e.g., Head/Neck: SoftCTM $0.6267~\text{mF}_1$, $\text{SD}~0.0268$ vs. $\text{CellViT}^{{\scriptscriptstyle ++}}_\text{SAM-H}$ $0.6359~\text{mF}_1$, $\text{SD}~0.0095$). These results highlight the exceptional data efficiency of our algorithm, making it particularly well-suited for exploratory analysis and establishing a strong benchmark in low-data regimes. A detailed breakdown is provided in the Supplementary Table \ref{tab:appendix_ocelot_amount_results}.

\begin{figure*}[!htb]
    \centering
    \includegraphics[width=\linewidth]{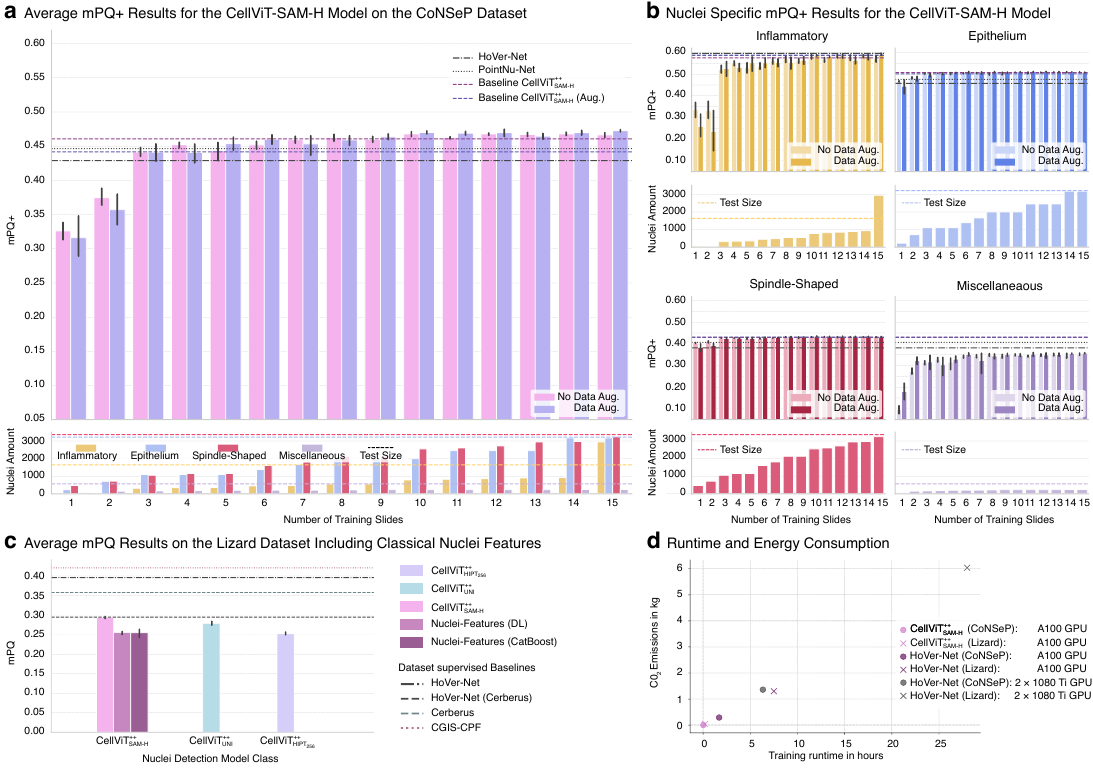}
    \caption{\textbf{Experimental evaluation on colon tissue cell datasets}. \textbf{a} Performance comparison of the $\text{CellViT}^{{\scriptscriptstyle ++}}_\text{SAM-H}$ network, with and without data augmentation, across varying amounts of training slides, besides baseline results from HoVer-Net and PointNu-Net (SOTA) on the CoNSeP dataset.  The upper panel presents the mean panoptic quality (mPQ), while the lower panel depicts the number of nuclei in the datasets. Training data is incrementally increased from a single crop of 1 slide to 15 crops across 15 slides.  \textbf{b} Nuclei specific performance comparison on CoNSeP. \textbf{c} Average mPQ on the Lizard dataset compared to top-performing networks. Additionally, $\text{CellViT}^{{\scriptscriptstyle ++}}_\text{SAM-H}$ is evaluated using both ViT token embeddings as cell features and classical nuclei features with deep learning and CatBoost classifiers. The HoVer-Net Cerberus is a re-trained version by~\citet{cerberus}. \textbf{d} Runtime and energy efficiency comparison of our network, trained on the CoNSeP and Lizard datasets, against HoVer-Net.}
    \label{fig:colon_results}
    \phantomsubcaption
    \label{fig:colon_results_a}
    \phantomsubcaption
    \label{fig:colon_results_b}
    \phantomsubcaption
    \label{fig:colon_results_c}
    \phantomsubcaption
    \label{fig:colon_results_d}
\end{figure*}

\subsection*{Data Efficient Learning for Cell Classification in Colorectal Cancer\\}
Colorectal cancer (CRC) is the third most common malignancy worldwide and a leading cause of cancer-related mortality, with global cases projected to rise from 1.9 million in 2020 to 3.2 million by 2040, and deaths expected to increase from 0.9 million to 1.6 million over the same period~\citep{colorectal_stats,worldcancerreport}. Enhanced segmentation models support research efforts by providing a deeper understanding of the tumor microenvironment, ultimately aiding in the development of more targeted and effective therapeutic strategies. \\
\indent Recent releases of two comprehensive datasets, CoNSeP and Lizard, have significantly advanced the field by providing extensive cell annotations and segmentation masks. These resources serve as crucial benchmarks for developing and validating cell segmentation models in colorectal cancer research. The CoNSeP dataset includes 41 annotated tiles with a size of $\num{1000} \times \num{1000}~\si{px}$ from University Hospitals Coventry and Warwickshire, UK, comprising a total of $\num{24332}$ cells divided into a fixed training set ($\num{15555}$ cells) and test set ($\num{8777}$ cells), as shown in the~\ref{fig:datset_overview_single_tissue}. The cells are categorized into four classes: inflammatory cells, epithelial cells, spindle-shaped cells, and miscellaneous cells. According to~\citet{hovernet}, we followed the official split using 27 tiles for training and 14 for testing. A 5-fold CV strategy was employed on the training slides to assess mean model performance. \\
\indent We compare $\text{CellViT}^{{\scriptscriptstyle ++}}$ with the original HoVer-Net publication, a self-trained HoVer-Net model, and the PointNu-Net~\citep{pointnunet}, the current SOTA network on this dataset. The evaluation metrics includes the $\text{mPQ+}$ score, a variant of the mPQ score for multiclass cell segmentation, as well as the binary $\text{F}_1$ (binary cell detection) and Dice score (binary cell segmentation). Among all $\text{CellViT}^{{\scriptscriptstyle ++}}$ variants, the SAM-H model performed best, achieving an $\text{mPQ+}$ of $0.461 (\text{SD}~0.014$). The baseline models achieved $0.429$ $\text{mPQ+}$, (HoVer-Net), and 0.446 $\text{mPQ+}$ (PointNu-Net). In terms of $\text{mPQ+}$, we set a new benchmark with $\text{CellViT}^{{\scriptscriptstyle ++}}_\text{SAM-H}$. \\
\indent In a zero-shot evaluation setting using models pretrained on the PanNuke dataset (without cell classification module), HoVer-Net achieved an $\text{F}_1$-score of $0.691$ and a Dice score of $0.802$. In comparison, $\text{CellViT}^{{\scriptscriptstyle ++}}_{\text{SAM-H}}$ reached an $\text{F}_1$-score of $0.772$ and a Dice score of $0.845$, demonstrating remarkable zero-shot performance. A detailed overview of the results for all $\text{CellViT}^{{\scriptscriptstyle ++}}$ models and the comparison methods is provided in Supplementary Table~\ref{tab:appendix_consep_comparison_all_networks_global}.\\
To further explore data efficiency, we simulated an active labeling approach, incrementally increasing the number of training tiles, starting with just one. Tiles were selected to ensure a balanced representation of cell classes. We repeated the experiments five times, without cross-validation, both with and without data augmentation. Remarkably, using only three fully annotated tiles with approximately $2,500$ cells, $\text{CellViT}^{{\scriptscriptstyle ++}}$ achieved performance comparable to the PointNu-Net network (trained on 27 tiles), and with four tiles, we exceeded it (see Fig.~\ref{fig:colon_results_a}). The analysis by cell class presented in Fig.~\ref{fig:colon_results_b} supports this hypothesis, demonstrating that careful selection of regions of interest (ROIs) with diverse cell composition and distinct cell nuclei can produce robust results even with limited data. For example, increasing the number of annotated inflammatory cells beyond 750 showed minimal impact, and a similar plateau effect was observed for epithelial and spindle-shaped cells. Our results on the CoNSeP dataset confirm that data augmentation does not improve performance and, in some cases, may reduce it (see Fig.~\ref{fig:colon_results_a} and ~\ref{fig:colon_results_b} for inflammatory cells). This finding aligns with the analyses on the Ocelot dataset.
In conclusion, we achieved superior performance in cell classification and excellent zero-shot segmentation on the CoNSeP dataset. Due to the excellent performance of $\text{CellViT}^{{\scriptscriptstyle ++}}_{\text{HIPT}_{256}}$, $\text{CellViT}^{{\scriptscriptstyle ++}}_{\text{UNI}}$, and $\text{CellViT}^{{\scriptscriptstyle ++}}_{\text{SAM-H}}$, we focus our analysis on these networks to evaluate which network yields to most representative cell embeddings.

\subsection*{Slide Resolution Generalizability\\}
Variability in lab protocols and scanning devices results in WSIs being acquired at different resolutions than the $\text{CellViT}^{{\scriptscriptstyle ++}}$ training standard of $0.25~\si{\text{\textmu} m \per px}$. Consequently, it is crucial to validate $\text{CellViT}^{{\scriptscriptstyle ++}}$ across a range of image resolutions. For this purpose, we employed the Lizard dataset, which provides ROIs from colorectal cancer samples at a resolution of $0.50~\si{\text{\textmu} m \per px}$. This dataset is among the largest available, featuring over $\num{418000}$ cell segmentation masks categorized into epithelium, lymphocytes, plasma cells, neutrophils, eosinophils, and connective tissue cells (see~\ref{fig:datset_overview_single_tissue}, and Supplementary Table \ref{tab:appendix_lizard_amount}). In a previous study, we observed that directly applying $0.50~\si{\text{\textmu} m \per px}$ without preprocessing image resolutions resulted in a significant distribution shift~\citep{CellViT}, degrading performance. To mitigate this issue, we applied a resampling method using a Lanczos filter to upscale the images to $0.25~\si{\text{\textmu} m \per px}$, process them with $\text{CellViT}^{{\scriptscriptstyle ++}}$, and then downscale the segmentation masks back to $0.50~\si{\text{\textmu} m \per px}$. We compare the performance of $\text{CellViT}^{{\scriptscriptstyle ++}}$ with resampling to several other segmentation models specifically tailored for this dataset, including HoVer-Net, Cerberus~\citep{cerberus}, and CGIS-CPF~\citep{cgis-cpf}. These models were trained directly on the Lizard dataset without any resampling or resolution adaptation. For evaluation, we use the PQ metric and report the average 3-fold CV results including standard deviation. We note that we removed the PanNuke samples of the Lizard dataset in our experiments, as they have been used in pretraining the $\text{CellViT}^{{\scriptscriptstyle ++}}$ segmentation models. The best $\text{CellViT}^{{\scriptscriptstyle ++}}$ model with SAM-H encoder achieved a binary PQ score of $0.536$ ($\text{SD}~0.009$), which is lower than of the comparison methods (HoVer-Net-Cerberus: $0.584$ ($\text{SD}~0.014$), HoVer-Net-Baseline: $0.624$ ($\text{SD}~0.139$), Cerberus: $0.612$ ($\text{SD}~0.009$), CGIS-CPF: $0.660$ ($\text{SD}~0.009$)). Incorporating the token-based cell classification module led to a mPQ-score of 0.294 ($\text{SD}~0.002$), which comes close to the performance of the supervised models with scores between 0.295 ($\text{SD}~0.018$) for HoVer-Net (Cerberus) and 0.421 ($\text{SD}~0.013$) for CGIS-CPF. These results highlight two key points: Even though $\text{CellViT}^{{\scriptscriptstyle ++}}$ was developed and optimized for a resolution of $0.25~\si{\text{\textmu} m \per px}$, it showcases remarkable adaptability across different resolutions, whereas the models evaluated here were trained directly on the Lizard dataset with resolution of $0.50~\si{\text{\textmu} m \per px}$. Secondly, $\text{CellViT}^{{\scriptscriptstyle ++}}$'s approach of utilizing a pre-trained segmentation network in a zero-shot setting, while only fine-tuning the classifier, offers a unique advantage in scenarios where annotated data may be limited or expensive to obtain. This efficient transfer learning strategy contrasts with the more data-intensive approach of training models from scratch, which, while beneficial for the baseline models in this specific case, may not always be feasible or desirable in real-world applications. 

\subsection*{Comparison with Conventional Feature Engineering Approaches\\}
To address the performance gap observed between $\text{CellViT}^{{\scriptscriptstyle ++}}$ and supervised models on the Lizard dataset, we examined if the quality of our token embeddings contributed to the discrepancy. For this, we compared the $\text{CellViT}^{{\scriptscriptstyle ++}}$ embeddings of $\text{HIPT}_{256}$, UNI and SAM-H image encoder with conventional handcrafted cell features (histomics). For each cell identified by $\text{CellViT}^{{\scriptscriptstyle ++}}_\text{SAM-H}$, we extracted 128 pre-defined nuclear features~\citep{nucleiio} (e.g., color, texture, shape, spatial, morphology, orientation) and trained a classifier similar to the token based classification module, as well as various classical machine learning models, including SVMs and CatBoost. The results, depicted in Fig.~\ref{fig:colon_results_c}, reveal that deep learning features led to an mPQ score of $0.294$ ($\text{SD}~0.002$), while hand-crafted features with a deep learning classifier scored $0.255$ ($\text{SD}~0.002$). The best classical machine learning model (CatBoost) using these handcrafted features achieved an mPQ of $0.255$($\text{SD}~0.009$), all lower than the foundation models cell features of $\text{CellViT}^{{\scriptscriptstyle ++}}_\text{SAM-H}$ and $\text{CellViT}^{{\scriptscriptstyle ++}}_\text{UNI}$ ($0.279~\text{mPQ}$, $\text{SD}~0.004$), but on par with $\text{CellViT}^{{\scriptscriptstyle ++}}_{\text{HIPT}_{256}}$ ($0.252~\text{mPQ}$, $\text{SD}~0.003$). These findings demonstrate that the token embeddings from $\text{CellViT}^{{\scriptscriptstyle ++}}$ are generally more informative than traditional handcrafted histomics features and remove human bias. Furthermore, generating token embeddings is more computationally efficient, as it avoids the additional overhead of calculating Histomics features, which took an average of 18.71 $\pm$ 37.55 ms per cell, contributing to a substantial time cost. More results are given in the Supplement (Tables~\ref{tab:appendix_consep_cell_amount_table}-\ref{tab:appendix_lizard_results}).

\begin{figure*}[!htb]
    \centering
    \includegraphics[width=\linewidth]{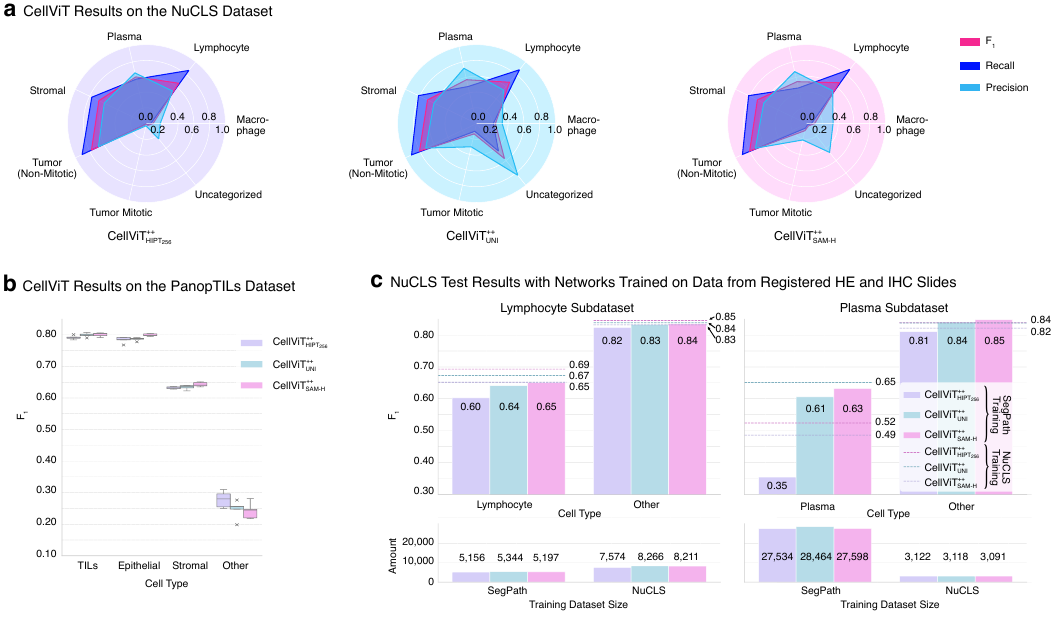}
    \caption{\textbf{Experimental evaluation on breast cancer tissue datasets, including training on automatically derived lymphocytes and plasma cells from the SegPath dataset}. \textbf{a} Comparison of $\text{F}_1$-score, precision, and recall for different $\text{CellViT}^{{\scriptscriptstyle ++}}$ models on the NuCLS dataset with all cell types included. \textbf{b} $\text{CellViT}^{{\scriptscriptstyle ++}}$ performance on the PanopTILs dataset for analyzing the tumor microenvironment in breast cancer. \textbf{c} Detection performance comparison of our network on the NuCLS test set, trained with automatically derived cells from the SegPath dataset versus fully supervised training on the NuCLS training dataset for lymphocytes and plasma cells. The lower panel shows the number of training cells in both datasets. For SegPath, $\text{CellViT}^{{\scriptscriptstyle ++}}$ was applied to HE-slides, with the resulting cell contours mapped to the IHC mask to derive cell classes.}
    \label{fig:breast_results}
    \phantomsubcaption
    \label{fig:breast_results_a}
    \phantomsubcaption
    \label{fig:breast_results_b}
    \phantomsubcaption
    \label{fig:breast_results_c}
\end{figure*}

\subsection*{Characterization of the Tumor Microenvironment in Breast Cancer\\}
Among various cancer types, breast cancer is notably significant as the most frequently diagnosed cancer among women~\citep{worldcancerreport}. For research on this tumor type, two complementary cell datasets have been released: the NuCLS dataset and the PanopTILs dataset. The NuCLS dataset includes multiple classification levels to distinguish between tumor cells, stromal cells, and inflammatory cells, while the PanopTILs dataset is specifically designed to assess tumor-infiltrating lymphocytes (TILs) and additional cellular components within the tumor microenvironment. Tumor-infiltrating lymphocytes are a critical component of the immune response within the tumor microenvironment and have been shown to correlate with patient prognosis and response to immunotherapy in several cancer types, including breast cancer~\citep{tils}. 
Due to the novelty of these datasets, advanced network architectures beyond Mask R-CNN models have yet to be developed. In this study, we establish a robust baseline on both the NuCLS (see Fig.~\ref{fig:breast_results_a}) and PanopTILs datasets (see Fig.~\ref{fig:breast_results_b}). Our analysis of $\text{F}_1$, recall, and precision scores for the $\text{CellViT}^{{\scriptscriptstyle ++}}$ models on NuCLS showed strong overall performance (see Fig.~\ref{fig:breast_results_a}). However, for rare cell types such as macrophages and mitotic tumor cells, recall is lower while precision is higher. This suggests that although these cells are identified less frequently, when they are detected, the detections are more accurate. In line with this, our models demonstrated strong performance on the PanopTILs dataset, achieving high $\text{F}_1$-scores for TILs ($\text{CellViT}^{{\scriptscriptstyle ++}}_\text{SAM-H}: 0.801~\text{F}_1$, $\text{SD}~0.006$). Additionally, for epithelial cells we achieved consistent $\text{F}_1$-scores of 0.800, and for stromal cells of 0.643, as depicted in Fig.\ref{fig:breast_results_b}). We anticipate that releasing these models will facilitate quantitative analysis of the tumor microenvironment in breast cancer tissue, replacing manual counting in ROIs by pathologists with automated methods~\citep{tils_2015, report_comp_til_assessment}.

\subsection*{Reducing $\textbf{CO}_{\bm{2}}$ Emissions and Training Time through Energy-Efficient Fine-tuning\\}
Training large models typically demands considerable computational resources, leading to significant $\text{CO}_2$ emissions~\citep{c02calc_source}. Our method mitigates this issue by utilizing pretrained domain-specific segmentation models, enabling efficient fine-tuning on new tasks and thereby reducing the carbon footprint associated with model training. We evaluate the energy efficiency of our approach using the metrics from~\citet{c02calculator}. Compared to traditional models such as HoVer-Net, our method demonstrated substantial reductions in both training time and $\text{CO}_2$ emissions. For instance, training HoVer-Net on the CoNSeP dataset with two Nvidia 1080 Ti GPUs took 6.33 hours, and 27.98 hours for the Lizard dataset, with an estimated power consumption of $\num{3170}~\si{WH}$ and $\num{14000}~\si{WH}$, respectively. \\
\indent To reduce training runtime, we performed experiments with an Nvidia A100 GPU, which reduced the training times to 1.70 hours ($680~\si{WH}$) and 7.5 hours ($\num{3000}~\si{WH}$) for both datasets (see Fig.~\ref{fig:colon_results_d}). Our $\text{CellViT}^{{\scriptscriptstyle ++}}$ model, on the other hand, required only 81 seconds for training on the CoNSeP ($9.23~\si{WH}$) and 12 minutes for training on the Lizard dataset ($92.16~\si{WH}$). Figure \ref{fig:colon_results_d} illustrates the superior resource efficiency, highlighting lower training times and $\text{CO}_2$ emissions compared to HoVer-Net on various hardware and datasets. Even with a hyperparameter search involving 100 runs, our $\text{CO}_2$ footprint remained lower than that of HoVer-Net. This efficiency arises because approximately 90\% of the training time in our AutoML pipeline is devoted to the initial caching of $\text{CellViT}^{{\scriptscriptstyle ++}}$ results and cell tokens. Once this step is completed, hyperparameter tuning becomes inexpensive, provided that data augmentation is not employed. In contrast, methods requiring full training of a segmentation model for each experiment incur higher computational and environmental costs. 

\subsection*{Automated Cell Dataset Generation from IF Stainings: Reducing Pathologist Involvement\\}

Our previous analyses have demonstrated that models like HoVer-Net and SoftCTM require large datasets with precise segmentation masks to achieve optimal performance, which creation is often performed from or at least supervised by pathologists. Manual annotation, particularly of segmentation masks, is resource-intensive and costly, posing a significant bottleneck in translational research. This limitation makes it impractical to create datasets at the necessary scale. Automated generation of cell datasets from IF staining presents a promising approach to reduce the reliance on pathologists for fine-grained segmentation annotations. \\
\indent To address this challenge, we propose an automated workflow for generating cell-specific datasets with minimal manual intervention. In this method, tissue samples are initially stained with H\&E, then destained and subsequently re-stained using immunohistochemistry (IHC) or IF markers to specifically target the desired cell types. After scanning the H\&E and IHC/IF samples, cell-level registration is performed to align the images, and a binary mask of positively stained cells is created through thresholding of the IF channel. A $\text{CellViT}^{{\scriptscriptstyle ++}}$ model is then applied to the H\&E-stained images, to extract cell segmentation masks. These mask can be re-mapped to the registered IF masks, marking positive and negative cells. The whole dataset creation workflow is depicted in Fig.~\ref{fig:overview_image_b}. This approach significantly reduces the need for manual annotation, requiring only a small pathologist-approved validation set to ensure accuracy.
In this study, we demonstrate the method using two exemplar cohorts from the SegPath dataset~\citep{segpath}, focusing on breast tissue. The dataset employs CD3/CD20 IF staining to specifically identify lymphocytes and MIST1 staining to highlight plasma cells along with registered H\&E slides. Training was conducted on cells automatically extracted from this dataset, following the previously described strategy. Final validation of the classifiers was performed on the pathologist-approved subset of the NuCLS dataset, in a one-vs.-all setting (lymphcoytes/plasma cells vs. all other types). As baselines, we additionally trained lymphocyte and plasma cell classifiers on the NuCLS training dataset. We successfully extracted around $\num{5000}$ lymphocytes and over $\num{27000}$ plasma cells from the SegPath dataset (see Fig.~\ref{fig:breast_results_c}). In comparison, the NuCLS training dataset contained approximately $\num{7500}$ lymphocytes and around 3,000 plasma cells. For training on the automatically generated cell datasets, the $\text{CellViT}^{{\scriptscriptstyle ++}}_\text{SAM-H}$ model achieved an $\text{F}_1$-score of 0.651 for the detection of lymphocytes and 0.632 for plasma cells. In contrast, training on the NuCLS dataset resulted in an $\text{F}_1$-score of 0.693 (-0.042) for lymphocytes and 0.524 (+0.108) for plasma cells (see Fig~\ref{fig:breast_results_c}). Notably, we trained a classifier for plasma cells using $\text{CellViT}^{{\scriptscriptstyle ++}}_\text{UNI}$ tokens on the NuCLS data, achieving an $\text{F}_1$-score of 0.651, slightly surpassing the results obtained with our automatically generated dataset. On average, the performance of classification modules trained on the automatically generated SegPath cell dataset approached that of those trained on the expert-level annotated NuCLS datasets (lymphocytes) and, with a sufficient amount of annotated cell data, even exceeded it (plasma cells).

\subsection*{Dealing with Low Prevalence when Detecting Mitotic Figures\\}
As demonstrated on the NuCLS dataset, recall for rare cell types, such as mitotic figures, is often low, while precision tends to be higher. To further evaluate this effect, we applied $\text{CellViT}^{{\scriptscriptstyle ++}}$ to mitosis detection using the MIDOG++ dataset~\citep{midogpp}, which includes human and canine tissue samples. The MIDOG++ dataset presents a challenging task due to the scarcity of mitotic figures, comprising only 0.16\% of the cell population. It contains $\num{11937}$ annotated mitotic figures and $\num{14351}$ hard negatives (non-mitotic figures that could falsely be recognized as mitotic figures) across 503 tissue sections.
Notably, the dataset is partially annotated, with non-mitotic cells left unannotated unless labeled as hard negatives. For dataset preparation, we extracted all cells using \mbox{CellViT} and labelled them as as either mitotic or non-mitotic. This resulted in a dataset of $\num{7398795}$ cells, used for $\text{CellViT}^{{\scriptscriptstyle ++}}_\text{SAM-H}$ training. Given the extreme class imbalance, we adjusted the training data to include mitotic to non-mitotic cell ratios of 1:1, 1:20, and 1:200.
The results are presented in Table \ref{tab:midog_results}. Our findings indicate that the model with 200 additional non-mitotic cells per mitotic figure performs best. 
A detailed analysis of the precision and recall scores (Supplementary Tables~\ref{tab:appendix_midog_precision} and~\ref{tab:appendix_midog_recall}) confirms that for rare events such as mitosis, recall remained lower than precision. On average, the model achieved a precision of $0.66$ ($\text{SD}~0.14$), which is higher than the recall of $0.54$ ($\text{SD}~0.09$). Performance also varies by tissue: for example, in human melanoma, the model reached a precision of $0.83$ ($\text{SD}~0.03$) and a recall of $0.62$ ($\text{SD}~0.03$), whereas in human breast cancer tissue, the precision was 0.77 ($\text{SD}~0.04$) and the recall $0.49$ ($\text{SD}~0.03$). Despite the applicability of our approach, it falls short of the baseline detection model RetinaNet. Although our cell classification module had an $\text{F}_1$-score of 0.998 on the validation dataset, its performance degraded when applied at the WSI level due to the large number of cells evaluated. This degradation occurs because even a small false positive rate or a small amount of missed mitotic figures can lead to a significant number of misclassifications when applied to million of cells. A visualization is given in~\ref{fig:midog_example}.

\begin{table}[]
    \centering
    \caption{Results ($\text{F}_\textbf{1}$) on the MIDOG++ test dataset, categorized by organ and sample origin. For the $\text{CellViT}^{{\scriptscriptstyle ++}}$ models, we also incorporated additional non-annotated cells to enhance the distinction between mitotic and non-mitotic cells during training. Specifically, 0, 20, or 200 (ratios 1:1, 1:20 and 1:200) additional non-mitotic figures were added per annotated mitotic figure.}
    \label{tab:midog_results}
    \resizebox{\linewidth}{!}{%
    \begin{tabular}{@{}lllrrrr@{}}
    \toprule
\multirow{2}{*}{\large Models}   &                         & \multirow{2}{*}{\large RetinaNet} &                      & \multicolumn{3}{l}{\multirow{2}{*}{\large $\text{CellViT}^{{\scriptscriptstyle ++}}_\text{SAM-H}$}} \\
                          &                         &                            & \multicolumn{1}{l}{} & \multicolumn{3}{l}{}                                                                         \\ \hline
    Organs                    & Origin                  & Baseline                         &    & Ratio 1:1           & Ratio 1:20           & Ratio 1:200      \\ \midrule
    Breast Cancer             & \multirow{3}{*}{Human}  & 0.71 ($\text{SD}$ 0.02)  &    & 0.50 ($\text{SD}$ 0.01)       & 0.55 ($\text{SD}$ 0.01)       & 0.60 ($\text{SD}$ 0.01)   \\
    Neuroendocrine Tumor      &                         & 0.59 ($\text{SD}$ 0.01)  &    & 0.36 ($\text{SD}$ 0.08)       & 0.45 ($\text{SD}$ 0.02)       & 0.50 ($\text{SD}$ 0.00)   \\
    Melanoma                  &                         & 0.81 ($\text{SD}$ 0.01)  &    & 0.61 ($\text{SD}$ 0.09)       & 0.67 ($\text{SD}$ 0.03)       & 0.71 ($\text{SD}$ 0.02)   \\ \midrule
    Cutaneous Mast Cell       & \multirow{4}{*}{Canine} & 0.82 ($\text{SD}$ 0.01)  &    & 0.63 ($\text{SD}$ 0.03)       & 0.66 ($\text{SD}$ 0.02)       & 0.70 ($\text{SD}$ 0.01)   \\
    Lung Cancer               &                         & 0.68 ($\text{SD}$ 0.02)  &    & 0.34 ($\text{SD}$ 0.03)       & 0.41 ($\text{SD}$ 0.01)       & 0.43 ($\text{SD}$ 0.02)   \\
    Lymphoma                  &                         & 0.73 ($\text{SD}$ 0.01)  &    & 0.47 ($\text{SD}$ 0.04)       & 0.51 ($\text{SD}$ 0.01)       & 0.58 ($\text{SD}$ 0.01)   \\
    Soft Tissue Sarcoma       &                         & 0.69 ($\text{SD}$ 0.01)  &    & 0.53 ($\text{SD}$ 0.04)       & 0.53 ($\text{SD}$ 0.01)       & 0.57 ($\text{SD}$ 0.02)   \\ \bottomrule
    \end{tabular}%
    }
\end{table}
\begin{figure*}[!h]
    \centering
    \includegraphics[width=\linewidth]{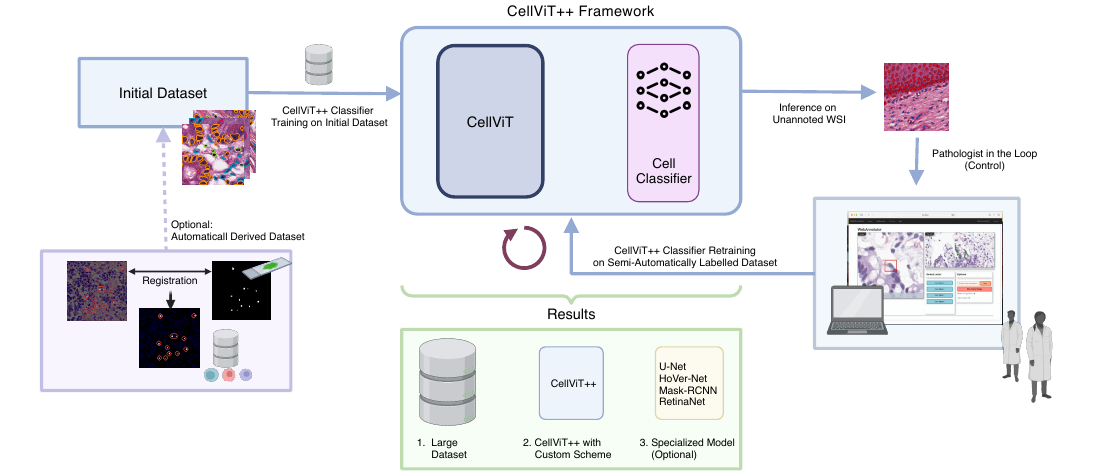}
    \caption{\textbf{Sugested workflow for minimal human intervention training}. }
    \label{fig:workflow_auto_ml}
\end{figure*}

\section*{Discussion}

Current approaches for cell segmentation and classification in digital pathology rely on large, fine-grained annotated datasets and precise segmentation masks, which are both cost- and time-intensive to acquire. Additionally, different tissue and tumor types require distinct cell classification schemes (e.g., TILs instead of generic inflammatory cells), adding further complexity.  Traditional methods are limited in their generalizability and lack flexibility for re-training and adaptation. This limitation necessitates the creation of a new dataset for each application, followed by the training of models such as HoVer-Net or SoftCTM, which is again resource-intensive and time-consuming. \\
\indent The key innovation of our work is the development of \mbox{$\text{CellViT}^{{\scriptscriptstyle ++}}$}, a generalized framework for  cell segmentation in H\&E images that allows lightweight adaptability without computational overhead. In our approach, deep cell features are extracted directly alongside segmentation masks, allowing for the creation of cell classification modules for new classification schemes without the need for re-training the segmentation model. \mbox{$\text{CellViT}^{{\scriptscriptstyle ++}}$} does not require data augmentation, which permits caching of cell segmentations and features during the first epoch of our AutoML hyperparameter search, further accelerating the training process. This feature is particularly beneficial for hyperparameter tuning, as caching only needs to be performed once per dataset. \\
\indent To evaluate the effectiveness of the $\text{CellViT}^{{\scriptscriptstyle ++}}$ framework, we conducted a series of experiments using datasets with varying properties, including differences in scanning resolution, tissue type, hospital origin, scanning devices, and cell types. Parts of our study focused on the clinically relevant cancer types colorectal cancer and breast cancer, both of which have high incidence and mortality rates, posing substantial challenges to healthcare systems worldwide. Our approach excels in scenarios with limited labeled data, highlighting its suitability for low-data environments. Importantly, the segmentation decoders in $\text{CellViT}^{{\scriptscriptstyle ++}}$ were not fine-tuned in this study, demonstrating its robust zero-shot segmentation capabilities and efficient classification performance, even when confronted with new cell types or tissues not represented in the training data. The versatility of our model suggests that it is a suitable foundational tool for future research. We envision significant applications, particularly in the discovery and validation of biomarkers, by providing adaptable and precise cellular features for in-depth cohort analysis. The deep cell features extracted from various tissue samples can enhance disease-specific pattern recognition, thereby contributing to personalized treatment strategies and patient stratification. Quantitative analyses of cell shape information are also conceivable.  \\
\indent Despite these advancements, limitations remain. A major concern is the absence of a quality control (QC) module in our current pipeline. As shown by~\citet{tolkach_quality_control}, segmentation model performance deteriorates significantly when working with suboptimal WSI such as those affected by blurriness or tissue folding. This highlights the importance of incorporating QC during preprocessing to ensure consistent model performance. This is not a model development issue but rather a pipeline optimization task. Additionally, while $\text{CellViT}^{{\scriptscriptstyle ++}}$ performs well in low-data scenarios, its performance can be surpassed by models specifically trained on large, task-specific datasets, such as Lizard. When sufficient training data is available, our performance saturates, whereas specialized networks have an advantage because their segmentation heads are also fine-tuned. However, the availability of such large datasets remains a significant constraint. For application of the $\text{CellViT}^{{\scriptscriptstyle ++}}$ framework, we propose the following workflow, as illustrated in Fig.~\ref{fig:workflow_auto_ml}: Initially, a cell dataset with limited annotations should either be manually curated or automatically generated. This dataset is then used to finetune $\text{CellViT}^{{\scriptscriptstyle ++}}$ on the relevant cell classes from the annotated initial training set, establishing a baseline model. As demonstrated in this study, this baseline already is likely to yield strong performance. Building on the trained model, inference is performed on an unlabeled dataset, followed by active labeling in which a pathologist verifies and corrects the model's predictions~\citep{nucleiio}. These semi-automatically generated annotations can be used to create a significantly larger training dataset, which can then serve as input for the training of specialized models, such as Cerberus~\citep{cerberus} or RetinaNet~\citep{retinanet}. In this workflow, $\text{CellViT}^{{\scriptscriptstyle ++}}$ operates as a task-agnostic foundation model, enabling the efficient development of task-specific models tailored to particular applications. \\
\indent To address the problem of data generation and labelling, we have also introduced a workflow (Fig.~\ref{fig:overview_image_b} and ~\ref{fig:workflow_auto_ml}) for the automatic creation of cell datasets, overcoming a significant bottleneck in this field. This automated dataset generation can perform comparably to manually annotated datasets, representing a major step forward. By utilizing specific antibodies for the target cell type, cell level datasets could be created with minimal expert intervention, significantly reducing the time and resources needed for dataset development. The proposed workflow (Fig.~\ref{fig:workflow_auto_ml}) outlines how an automatic training dataset can be created through the careful selection of markers, followed by the generation of a manual validation dataset using a web-based annotation tool. However, it is worth noting that technical expertise is required for image registration, and the selection of IF/IHC stainings still necessitates domain expertise. Only nucleus-staining markers are suitable for this workflow, and further analyses are needed to determine which stains are most appropriate for which cell type. \\
\indent A further limitation of our approach is its reliance on high-resolution WSI with a spatial resolution of $0.25~\si{\text{\textmu} m \per px}$, which may not be available in all laboratories. Although lower-resolution WSIs can be artificially upscaled to $0.25~\si{\text{\textmu} m \per px}$, as demonstrated in our experiments on the Lizard dataset, this upscaling may introduce artifacts that affect model performance. Whereas the model's token size of $16\times16~\si{px}$ ($\text{HIPT}_{256}$, UNI, SAM-H) or $14\times14~\si{px}$ (Virchow, $\text{Virchow}_2$) is generally well-suited for capturing most cellular structures, in edge cases where adjacent cells are smaller than the token size, there may be challenges in distinguishing between cells from different classes that are represented by the same token. Another limitation is the reduced effectiveness of $\text{CellViT}^{{\scriptscriptstyle ++}}$ when handling rare cell types, such as mitotic figures, which represent only 0.16\% of all cells in the MIDOG++ dataset. In such cases, the classification module must achieve a high level of accuracy, as errors in identifying these rare cell types can disproportionately influence overall performance. \\
\indent Future research will focus on improving $\text{CellViT}^{{\scriptscriptstyle ++}}$ inference time for real-time diagnostic applications, where speed and efficiency are crucial. While the inference time of $\text{CellViT}^{{\scriptscriptstyle ++}}$ has already been reduced by 40.39\% compared to the original \mbox{CellViT} publication~\citep{CellViT}, gigapixel WSIs still require 10-15 minutes for processing, which may be a limitation in clinical environments. \\
\indent While these limitations need further investigation, the broader implications of our work are significant. Unlike many medical AI studies, which often focus solely on task-speficic evaluation and methods, we aim to facilitate widespread adoption by releasing all training code, algorithms, models and derived datasets. This ensures transparency and reproducibility while enabling other researchers to build upon our framework. Additionally, we have developed a web-based viewer and annotation tools (see Fig.~\ref{fig:workflow_auto_ml} and~\ref{fig:extended_viewer}) that eliminate the need for local software installation, allowing for seamless integration into clinical environments. By leveraging cloud computing, our approach is designed to be scalable and accessible.\\
\indent In conclusion, $\text{CellViT}^{{\scriptscriptstyle ++}}$ marks a significant advancement in digital pathology, providing a robust, adaptable, and efficient framework for high-throughput cell segmentation and classification. By leveraging foundation models and automated dataset generation, $\text{CellViT}^{{\scriptscriptstyle ++}}$ accelerates research and enhances diagnostic precision. This work underscores the potential of AI, particularly deep learning, in improving diagnostic workflows and, ultimately, patient outcomes. With $\text{CellViT}^{{\scriptscriptstyle ++}}$, we give researchers and pathologists a tool that streamlines the analysis of complex tissue samples, reducing manual labor and minimizing errors in cell detection.
Integrating such models into clinical practice holds the promise of not only enhancing diagnostic accuracy but also offering deeper insights into tumor biology, thereby enabling more data-driven and evidence-based therapeutic decision-making.


\section*{CRediT authorship contribution statement} 
\textbf{Fabian Hörst:} Conceptualization, Methodology, Software, Formal Analysis, Investigation, Data Curation, Writing - Original Draft, Writing - Review \& Editing, Visualization.
\textbf{Moritz Rempe:} Methodology, Writing - Original Draft, Writing - Review \& Editing.
\textbf{Helmut Becker:} Methodology, Writing - Original Draft, Writing - Review \& Editing.
\textbf{Lukas Heine:} Software, Methodology, Writing - Original Draft, Writing - Review \& Editing.
\textbf{Julius Keyl:} Validation, Writing - Review \& Editing.
\textbf{Jens Kleesiek:} Resources, Writing - Review \& Editing, Supervision, Project administration, Funding acquisition.

\acknow{This work received funding from \lq KITE' (Plattform für KI-Translation Essen) from the REACT-EU initiative (\url{https://kite.ikim.nrw/}, EFRE-0801977) and the Cancer Research Center Cologne Essen (CCCE). The work of Helmut Becker was funded by a PhD grant from the DFG Research Training Group 2535 "Knowledge-and data-based personalization of medicine at the point of care (WisPerMed)".
Moritz Rempe and Jens Kleesiek received funding from the Bruno \& Helene Jöster Foundation. We acknowledge support by the Open Access Publication Fund of the University of Duisburg-Essen. The authors acknowledge that this manuscript was edited with the assistance of LLMs. The authors declare no competing interests.\\
The following Figures have been created with BioRender:
\begin{itemize}
    \item Figure 1: Created in BioRender. Hörst, F. (2025) \url{https://BioRender.com/j63e764}
    \item Figure 5: Created in BioRender. Hörst, F. (2025) \url{https://BioRender.com/u10j553}
    \item Extended Data Fig. 1: Created in BioRender. Hörst, F. (2025) \url{https://BioRender.com/v55z265}
\end{itemize}
}
\showacknow{} 


\bibliography{bibliography}
\section*{Methods}

\subsection*{Network Architecture}
\subsubsection*{CellViT.}
\mbox{CellViT} is a deep learning architecture designed specifically for nuclei segmentation and classification in histopathological images. The architecture is inspired by the UNETR~\citep{hatamizadeh2022unetr} network, adapted for 2D images, resulting in a U-shaped structure consisting of an image encoder and segmention decoders. To facilitate the task of nuclei segmentation and to allow the network to separate overlapping nuclei, \mbox{CellViT} employs three distinct multi-task decoder branches~\citep{hovernet}. The first branch predicts the binary segmentation map of all nuclei, the second generates horizontal and vertical distance maps to provide spatial information to delineate the nuclei and the third branch predicts the nuclei type map, enabling the classification into the PanNuke nuclei classes~\citep{CellViT}, which we removed in this study. To integrate the outputs from these branches, we use the postprocessing strategy of HoVer-Net to split up overlapping nuclei. The core of \mbox{CellViT} leverages the Vision Transformer architecture to encode images, which is an adaption of the Transformer model originally developed for natural language processing tailored for image analysis. Given an input image $\bm{x}\in \mathbb{R}^{H\times W\times 3}$ with height $H$, width $W$ and 3 input channels, the image is divided into a sequence of flattened tokens $\bm{x}_\text{p} \in \mathbb{R}^{N \times \left(P^2 \cdot 3 \right)}$, $N=HW/P^2$. Each of these token is a squared image section with the dimension $P \times P$, mapped by a linear projection layer into a $D$ dimensional latent space vector $z_0 \in \mathbb{R}^{D}$ whose size remains constant through all of the Transformer layers $l=1 \cdots L$, resulting in a token matrix $\bm{z}_0\in\mathbb{R}^{(N+k)\times D}$. Additional tokens such like the class token [CLS] or register tokens [REG] are summarized in the scalar $k$. Each of the Transformer layers $l$ consists of the self-attention mechanism with $H$ heads, updating the tokens $\bm{z}_l=f(\bm{z}_{l-1}),~l=1 \cdots L,$ successively. Using this approach, ViTs are able to capture complex spatial relationships across the image as each token attends to all remaining tokens. Additionally, five skip connections are added between the ViT encoder at different depth levels $l$ and the corresponding upsampling path for information passing to the upsamling decoder branches. \\
Using this method, we achieved outstanding results on the PanNuke dataset and demonstrated the generalizability of our model to new datasets with significant distributional differences. This underscores the robustness and adaptability of our approach to distribution shifts. The most important factor contributing to this perfromance improvement was that the ViT encoder architecture allowed us to use large pretrained networks, so-called foundation models.

\subsubsection*{Inference.}Although all \mbox{CellViT} models were initially trained using input images of $256 \times 256~\si{px}$, WSI-wise inference is performed on image sections of $\num{1024} \times \num{1024}~\si{px}$ size with an overlap of $64~\si{px}$. To accelerate postprocessing, we optimized the HoVer-Net postprocessing strategy to run on CUDA-enabled GPUs by utilizing libraries such as Numba~\citep{numba}, CuPY~\citep{cupy} and Ray~\citep{ray}. This optimization enables efficient, independent parallel processing of each patch within a batch on GPUs. The final step involves merging overlapping cells, a process that incurs minimal computational overhead due to the relatively small number of overlapping cells compared to the total number of cells in a WSI. By optimizing the pipeline, we achieved a 40.38\% reduction in runtime (SD 4.77\%) on a test dataset consisting of 10 diverse WSIs, compared to our original publication of CellViT~\citep{CellViT}.

\subsubsection*{Token based Cell Classifcation Module and $\text{CellViT}^{{\textbf{++}}}$.}
As outlined in the introduction to the \mbox{CellViT} model, our approach primarily relies on a ViT-based image encoder, where the input data is represented and processed as tokens $\bm{z}_l\in\mathbb{R}^{(N+k)\times D}$ at each layer $l$. Given that WSI adhere to a fixed physical optical scale - defined by a consistent magnification through a microscope, typically with a resolution of $0.25~\si{\text{\textmu} m \per px}$ at $\times~40$ magnification or $0.50~\si{\text{\textmu} m \per px}$ at $\times~20$ magnification - a consistent mapping between the objects within the image and the corresponding tokens can be established, which remains constant across WSIs with identical scanning setups. Given an input resolution of $0.25~\si{\text{\textmu} m \per px}$ of the images for the \mbox{CellViT} segmentation algorithm and a token size of $P=14~\si{px}$ or $P=16~\si{px}$, each token corresponds to approximately $4~\si{\text{\textmu} m}$, which is in the range of the size of a human cell nuclei~\citep{hipt}.  Excluding the [CLS] token and any potential register tokens [Reg], the token matrices $\bm{z}_l$ of the respective Transformer layers $l$ can be rearranged into a three-dimensional tensor $\bm{Z}_j \in \mathbb{R}^{\frac{H}{P} \times \frac{W}{P}\times D}$, where the first two dimensions correspond to the spatial arrangement analogous to the input WSI sections. This spatial arrangement, coupled with the fact that the size of each token roughly matches that of a nucleus, allows for the assignment of tokens $z^{\hat{y}_j}_l \in \mathbb{R}^{D}$ to each detected cell $\hat{y}_j$. Since each layer $l$ further abstracts and enriches the tokens with additional information, we only assign each cell the token $z^{\hat{y}_j}_L$ of the output of the last Transformer $L$, thereby generating an embedding vector for each cell. Consequently, in the \mbox{CellViT} model, tokens from the last Transformer layer of the ViT encoder can be directly mapped to individual cells detected at the output of the segmentation head. This structure allows for efficient computation of deep cell features directly within the forward pass of the model. If a nucleus is associated with multiple tokens, we average over all token embeddings in which the nucleus is located.

A significant advantage of our method is that the cell embedding vectors can be directly extracted during a forward pass of or \mbox{CellViT} segmentation models. In contrast, existing two-stage models consists of a segmentation model and a separate feature extraction method. For this, visual crops of the segmented cells serves as inputs for a second model that either extract classical image features, like by~\citet{nucleiio}, or computes an embedding vector using another deep learning model. This two-stage approach introduces considerable overhead regarding computational time and memory, as the number of cells in a typical WSI can range from several hundred thousand to millions. By extracting cell embeddings jointly with the segmentation process, our method eliminates this overhead, as the tokens are inherently present within the ViT encoder. \\

In this work, cell tokens are utilized to extend the \mbox{CellViT} model to include new cell classes. We achieve this by extracting cell embeddings from annotated datasets, where each cell is assigned a class label. This process results in a dataset comprising cell embedding vectors paired with their corresponding cell labels. We then train a classifier to predict the cell class based on these tokens. This information can be leveraged in the post-processing stage to adaptively adjust the cell classes that \mbox{CellViT} can recognize, which are initially limited to the PanNuke classes. Each cell at the output is then assigned the prediction of our cell classification module. This approach allows for the inclusion of more fine-grained cell categories or the definition of organ-specific classes. We denote this combination of segmentation models and the lightweight cell classification module as $\textbf{CellViT}^{{\textbf{++}}}$.
\\
For the classifier, we employ a fully connected feedforward network with one hidden layer and a ReLU activation function. To avoid negatively impacting downstream segmentation performance and for efficiency reasons, only the classifier is re-trained based on the tokens and cell labels, while the $\text{CellViT}^{{\scriptscriptstyle ++}}$ segmentation network remains unchanged.

\subsection*{Foundation Models\\}\label{sec:foundation_models} 
As previously mentioned, the primary motivation behind the development of the \mbox{CellViT} architecture was to create a network structure capable of leveraging domain-specific and task-specific foundation models for fine-grained cell segmentation and detection, based on the ViT architecture. A list of the included foundation models and their specifications is provided in this section, with the networks organized in ascending order based on the number of parameters. The models $\text{HIPT}_{256}$, UNI, Virchow and $\text{Virchow}_2$ are all histopathological foundation models, while SAM is a task specific foundation model for image segmentation.

\subsubsection*{$\text{HIPT}_{256}$.} In 2022~\citet{hipt} introduced one of the first foundation models for histopathological slide analysis, called Hierarchical
Image Pyramid Transformer (HIPT). The HIPT model is a hierachrical network designed for slide-level representation extraction by stacking multiple ViT-stages on top of each other. The architecure employs a three-stage hierarchical structure performing aggregation from cell-level over micro-tissue structure up to whole slide representation. In total, their network consists of three stages, with each stage independently pretrained using DINO~\citep{dino}. The $\text{HIPT}_{256}$ model refers specifically to the first stage, which processes $16 \times 16~\si{px}$-sized visual tokens extracted from $256 \times 256~\si{px}$ patches to create a local cell-cluster token~\citep{CellViT}, employing a vanilla ViT-Small (ViT-S) architecture with 21.7 million parameters. The stage was pretrained on 104 million patches extracted from $\num{10678}$ slides of TCGA. By analyzing the attention heatmaps of the Transfomer layers within the ViT,~\citet{hipt} showed that the network effectively learned visual concepts specific to histopathological images, including cell-locations. This makes $\text{HIPT}_{256}$ a powerful pretrained backbone for our $\text{CellViT}^{{\scriptscriptstyle ++}}$ model. Notably, it is the smallest encoder we have examined in this study, with only $3.3~\%$ parameter count of our largest model with ViT-H backbone. The ViT-S specifications are $P=16, D=384, L=12, H=6$.

\subsubsection*{UNI.} UNI, developed by the same group as  $\text{HIPT}_{256}$, was specifically tailored to be a general-purpose model for computational pathology to generate patch embeddings. To increase the versatility of UNI, the authors used a more diverse dataset for pretraining, utilizing an internal dataset of more than $\num{100000}$ diagnostic WSIs across 20 major tissue types~\citep{uni}. This dataset is one of the largest available collections of WSI and one magnitude larger than the commonly used TCGA dataset. The model was pretrained using DinoV2~\citep{dinov2} self-supervised training algorithm with more than 100 million patches and is based on a ViT-L architecture with 307 million parameters. UNI excels in weakly supervised slide-level classification tasks, outperforming competitive pretrained encoders. However, as~\citet{uni} note, UNI lacks vision-specific biases for solving dense prediction tasks even though they showed competitive segmentation results using Mask2Former approach~\citep{mask2former}, but without the capability to perform instance-wise segmentation. The ViT-L backbone has the following specification: $P=16, D=1024, L=24, H=12$. 

\subsubsection*{Virchow.} 
The Virchow~\citep{virchow} model represents an advancement over UNI with slightly better performance in a cancer detection comparison, outperforming UNI on 13 out of 16 cancer types. Similarly to UNI, the model was trained with DinoV2~\citep{dinov2} on an internal dataset comprising $\num{1488550}$ WSIs from $\num{119629}$ patients. For this, 2 billion tiles were randomly sampled from these WSIs. The architecture employs a ViT-H model with 632 million parameters, but differing from all the previous networks, with a token size of $14 \times 14~\si{px}$. Thus, to align all models, we reshaped the PanNuke training patches from $256 \times 256~\si{px}$ to $252 \times 252~\si{px}$ and performed inference on rescaled $\num{1022} \times \num{1022}~\si{px}$ instead of $\num{1024} \times \num{1024}~\si{px}$ patches. The model's parameters include an embedding dimension of $D=1280$, $L=32$ Transformer layers, and $H=16$ attention heads.

\subsubsection*{$\text{Virchow}_{\textit{2}}$.} $\text{Virchow}_2$ builds upon the Virchow model by incorporating additional register tokens into the ViT structure and further training on an expanded dataset of 3.1 million histopathology WSIs, derived from $\num{225401}$ patients across multiple magnifications ($\times 5$, $\times 10$, $\times 20$, and $\times 40$) and staining techniques (H\&E and IHC)~\citep{virchow2}. This dataset extends the original Virchow dataset and includes nearly 200 tissue types, with 15\% of the WSIs and 57\% of the patients sourced from diverse institutions worldwide. Trained with an adapted DINOv2 self-supervised learning algorithm, $\text{Virchow}_2$ outperforms both UNI and Virchow on average across eight public tile-level classification benchmarks such as CRC-100k (colorectal cancer morphology classification), Camelyon16 (lymph node metastasis detection in breast cancer) or the challenging Hest-1k dataset (spatial transcriptomics)~\citep{Hest1k}, demonstrating its robustness across a wide range of applications. On average, $\text{Virchow}_2$ achieved an $F_1$-score of 0.885, Virchow of 0.877, and UNI of 0.866. The model's parameters remain consistent with those of Virchow (ViT-H) but with the addition of 4 register tokens $\left(P=14, D=1280, L=32, H=16\right)$.

\subsubsection*{Segment Anything.}
In the field of natural image processing,~\citet{sam} introduced the Segment Anything Model (SAM), an open-source segmentation model designed for semantic segmentation. SAM's architecture consists of an image encoder based on the ViT model and a lightweight mask decoder network. The model is available in three different sizes - SAM-B, SAM-L, and SAM-H - corresponding to ViT-Base, ViT-Large, and ViT-Huge architectures, respectively. The final backbone, SAM-H, was trained in a supervised manner on 1.1 billion segmentation masks from 11 million images. According to our previous study~\citep{CellViT}, \mbox{CellViT} with SAM-H backbone outperforms its smaller counterparts in cell segmentation, making it the preferred backbone for this task, with 632 million parameters. Recently, the model was updated with a memory module to capture temporal consistencies in videos, termed SAM-2~\citep{sam2}. However, for applications not involving time series image data, the original SAM model performs equivalently. This work is limited to the ViT-H backbone (SAM-H), with $P=16, D=1280, L=32,$ and $ H=16$.
Abstract von~\cite{sam_biomed_applications} Recent advancements in biomedical image analysis have been significantly driven by the Segment Anything Model (SAM). This transformative technology, originally developed for general-purpose computer vision, has found rapid application in medical image processing. Within the last year, marked by over 100 publications, SAM has demonstrated its prowess in zero-shot learning adaptations for medical imaging. The fundamental promise of SAM lies in its capability to segment or identify objects in images without prior knowledge of the object type or imaging modality.

\subsection*{Experimental Setup\\}
The primary aim of this study is to assess how effectively cell embeddings generated by $\text{CellViT}^{{\scriptscriptstyle ++}}$ in a single forward pass can be leveraged to train classifiers for newly defined cell classes. Additionally, the study explores variations in cell embeddings across different foundation models and evaluates their efficacy for cell classification tasks. To this end, we employed a variety of datasets, including two multi-organ datasets (Ocelot and MIDOG++) featuring organs not present in the PanNuke training dataset. Furthermore, we assessed $\text{CellViT}^{{\scriptscriptstyle ++}}$ on two prevalent cancer types: breast cancer, the most commonly diagnosed cancer in women (2.1 million new cases in 2018), using the NuCLS, SegPath, and PanopTILs datasets, and colorectal cancer, the third most common cancer overall (1.8 million new cases in 2018), using the CoNSeP and Lizard datasets.

Despite differences in scope, cell types, and dataset composition, all datasets follow a consistent structure. Each consists of a training set and an externally defined test set, with original publication splits adhered to for comparability. Some datasets, such as Ocelot, include validation splits, while others, like Lizard, use a cross-validation strategy (3-fold in this case). For datasets lacking explicit validation data definitions, we conducted 5-fold CV within the training data and applied models from each fold to the external test dataset. For datasets with validation splits but without CV strategies, experiments were performed using five different random seeds, resulting in five models. Reported results represent the mean scores across these models (except Lizard, which uses 3-fold CV). Any deviations from this methodology are specified in the respective experimental sections.

For each experiment and CellViT-variant, hyperparameter tuning was initially performed with 100 random runs on the training-validation split or the first CV fold, optimizing the classifier's hyperparameters based on validation performance. The tuned hyperparameters included the hidden dimension of the classifier, the learning rate, and the weight decay of the optimizer. Additionally, we examined the impact of exponential learning rate scheduling with a factor of $\gamma = 0.95$ versus a constant learning rate reduced by half after half the training epochs. This tuning was conducted without data augmentation to minimize runtime. Due to the lack of data augmentation, training and validation cells could be cached after the initial epoch, allowing hyperparameter tuning to be completed within a few hours.

\subsubsection*{Training setup}
We used the AdamW optimizer with weight decay and \(\beta_1 = 0.85\) and \(\beta_2 = 0.9\). Training was conducted with a batch size of 256 over 50 epochs, applying early stopping after 10 epochs without improvement of the area under receiver operating characteristics curve of the classifier. Additionally, we used dropout with a rate of 0.1 as regularization technique. As the \mbox{CellViT} model was trained with mixed precision (AMP training), we also train the classification module with mixed precision. Data caching of $\text{CellViT}^{{\scriptscriptstyle ++}}$ cell detections along with the tokens was enabled after the first epoch when no data augmentation was used, which is the default configuration. Images were normalized according to the normalization parameters of the foundation models. All experiments were conducted on a NVIDA A100 GPU with $80~\si{GB}$ VRAM on a workstation equipped with 24CPUs and 76GB RAM. However, most of the computations here could be reasonably performed on smaller GPUs (e.g., NVIDIA RTX A5000 with $24~\si{GB}$ VRAM) as well.

\subsection*{Classical Nuclei Feature Calculation\\}
To evaluate the effectiveness of embeddings generated by $\text{CellViT}^{{\scriptscriptstyle ++}}$, we compared them against traditional two-stage models, where $\text{CellViT}^{{\scriptscriptstyle ++}}$ was employed specifically for cell segmentation which are subsequently used to extract classical features accordingly to~\citet{nucleiio}. The feature set used in this comparison consists of 128 distint features, encompassing the 122 classical features of Nuclei.io~\citep{nucleiio}. We introduced six additional features related to cytoplasmic region and texture analysis. A list of all features can be found in the publication of~\citet{nucleiio}. These features capture a range of color, morphological, textural, and spatial properties of the nuclei and cytoplasm. The cytoplasmic region was defined as a 40-by-40 pixel area centered on the nucleus at a resolution of $0.25~\si{\text{\textmu} m \per px}$, allowing for a consistent analysis across samples. \\
For the development of cell classifiers using classical cell features, we compared the three-layer fully connected classification module we use with the $\text{CellViT}^{{\scriptscriptstyle ++}}$ token approach with traditional machine learning algorithms, including logistic regression, decision trees, XGBoost, and CatBoost (see Supplementary Table~\ref{tab:appendix_pycaret_models}). We incorporated the auto-ML tool PyCaret into our pipeline for systematic evaluation and selection of the optimal classification method based on the training data. Through this process, CatBoost demonstrated superior performance across multiple metrics, thereby establishing it as the most effective classical algorithm for our experiments. Consequently, we limited our analysis to CatBoost as a representative classical machine learning technique in our study.

\subsection*{Datasets}

\subsubsection*{Ocelot.}
The Ocelot dataset~\citep{ocelot} is a tumor cell detection dataset to advance the understanding of cell-tissue interaction across multiple human organ systems. It contains a total of $\num{113026}$ cells, annotated to differentiate between tumor and non-tumor cells. These cells were derived from human specimens from the TCGA dataset of six organs: Bladder, endometrium, head/neck, kidney, prostate, and stomach. In total, the datasets consists of 664 patches extracted from 303 WSI, with each patch measuring $\num{1024} \times \num{1024}~\si{px}$. Each patch was acquired at $0.20~\si{\text{\textmu} m \per px}$ at a magnification of $\times 40$. While the original dataset also includes region-level fields of view for each cell-level patch to assess whether context regions can improve tumor cell detection, this study focuses exclusively on the annotated cell-level dataset. The dataset is divided into three subsets, training (400 patches), validation (137 patches), and test (126 patches), with a 6:2:2 split ratio at WSI level to prevent leakage~\citep{ocelot}. The tumor cell distribution is consistent among all three subsets, with around $65\%$ tumor cells and $35\%$ non-tumor cells. In addition to the full training dataset, we further splitted the dataset into smaller training subsets to assess model performance under limited data conditions. These subsets consist of 5\% (18 patches), 10\% (44 patches), 25\% (103 patches), 50\% (201 patches) and 75\% (300 patches) of the total training data, allowing for an analysis in regard of data efficiency. A table with detailed information about the split and the cell amount is given in the Supplement (Tab. \ref{tab:appendix_ocelot_cell_amount}). Annotations were perfomed by board-certified pathologists with a census strategy among three independent pathologists for each patch. The dataset does not provide segmentation masks of cell contours. 

For a fair comparison, all baseline experiments using the SoftCTM model have been conducted on the Ocelot dataset including the region-wise tumor segmentation masks. 

\subsubsection*{MIDOG++.}
The MIDOG++ dataset is an extended version of the original MIDOG challenge dataset, designed for mitotic figure detection in histopathology images~\citep{midogpp}. It represents the most extensive multi-domain mitotic figure dataset to date, incorporating images from multiple centers, including UMC Utrecht, VMU Vienna, FU Berlin, and AMC New York. The dataset was acquired using multiple scanners to evaluate domain generalization, with images captured at a resolution of 0.23 to $0.25~\si{\text{\textmu} m \per px}$, comprising a total of 503 images. The dataset includes 111 test images, representing a 20\% stratified split. For each experiment, we performed a stratified 5-fold cross-validation, as suggested by the authors. 
Annotations in MIDOG++ were generated through a multi-expert consensus strategy, involving expert-level reviews and validation by an automated algorithm. The dataset contains a total of $\num{11937}$ annotated mitotic figures and $\num{14351}$ hard negatives. Utilizing the $\text{CellViT}^{{\scriptscriptstyle ++}}_\text{SAM-H}$ model, we extracted a total of $\num{7398795}$ cells, with mitotic figures representing only 0.16\% of all cells. The image tiles have varying sizes, with an average width of $\num{6804}$ pixels and an average height of $\num{5102}$ pixels, covering larger sections than any other comparable datasets in this study. The primary evaluation metric for this dataset is the $\text{F}_1$-score. We compared $\text{CellViT}^{{\scriptscriptstyle ++}}$ with the object detection model RetinaNet, trained by~\citet{midogpp}.

\subsubsection*{CoNSeP (Colorectal Nuclear Segmentation and Phenotypes).} This dataset was introduced as part of the HoVer-Net publication~\citep{hovernet} and comprises 41 image tiles derived from colon cancer patients at the University Hospitals Coventry and Warwickshire, UK. Each tile measures $num{1000} \times num{1000}~\si{px}$, with a resolution of $0.25~\si{\text{\textmu} m \per px}$ ($\times 40$). The dataset includes various cell types such as inflammatory cells, healthy epithelial cells, dysplastic/malignant epithelial cells, fibroblasts, muscle cells, and endothelial cells. The annotations were initially performed by one pathologist for each slide, with a second pathologist reviewing the annotations to reach a consensus. Aligned with HoVer-Net, we group normal and dysplastic/malignant epithelial cells under a single epithelial category, and fibroblasts, muscle cells, and endothelial cells under a spindle-shaped nuclei category. The dataset is divided into 27 training tiles and 14 test tiles. To facilitate algorithmic comparisons, we conducted 5-fold CV on the training images, with the final evaluation carried out on the test images.
Additionally, we mimicked an iterative labeling approach to assess the cell classification module performance with limited training data. Starting with a single annotated tile, we incrementally increased the number of annotated tiles up to 15, evaluating the impact on model performance at each step. A detailed overview of the number of tiles used at each stage is provided in Supplementary Table~\ref{tab:appendix_consep_cell_amount_table}. The dataset provides segmentation masks such that in addition to the detection capability, also the segmentation performance can be evaluated. 

\subsubsection*{Lizard.} The Lizard dataset~\citep{lizard} extends the CoNSeP dataset by integrating colon cancer tissue from multiple sources, including PanNuke~\citep{pannuke}, GlaS~\citep{glas}, CRAG~\citep{crag}, DigestPath, TCGA~\citep{tcga}, and CoNSeP~\citep{hovernet}. The dataset includes contributions from University Hospitals Coventry and Warwickshire, UK, multiple centers in the USA, and four hospitals in China~\citep{lizard}. Unlike other datasets used in this study, the Lizard dataset was acquired at a resolution of $0.50~\si{\text{\textmu} m \per px}$ using $20\times$ magnification. An iterative labeling process was employed, beginning with automatic labeling, followed by semi-automatic refinement, and concluding with manual boundary refinement. This process resulted in the annotation of $\num{418935}$ nuclei (we excluded TCGA and PanNuke), classified into neutrophils, lymphocytes, plasma cells, eosinophils, epithelial cells, and connective tissue cells (see \ref{tab:appendix_lizard_amount}). The dataset comprises 270 tiles of varying sizes, with an average dimension of $\num{1016} \times 917~\si{px}$~\citep{lizard}. To process the Lizard dataset with $\text{CellViT}^{{\scriptscriptstyle ++}}$, the input images were rescaled to $0.25~\si{\text{\textmu} m \per px}$ using Lanczos resampling, and the output was subsequently rescaled back to $0.50~\si{\text{\textmu} m \per px}$ prior to performance metric calculations. This procedure enabled the evaluation of the network's ability to generalize across different magnifications. The evaluation was conducted using 3-fold CV based on the splits provided in the original paper, with the PanNuke images excluded for the $\text{CellViT}^{{\scriptscriptstyle ++}}$ evaluation, due to prior network pretraining. Since the TCGA test set is not publicly available, no external evaluation could be conducted. Similar to CoNSeP, this dataset provides cell-level segmentation masks.

\subsubsection*{NuCLS.}
The NuCLS dataset is a crowdsourced collection of nuclei annotations from breast cancer tissue (source TCGA), annotated by medical students and pathologists~\citep{nucls}. It contains over 220,000 annotations, with varying levels of quality depending on the  extent of pathologist involvement. Specifically, some parts of the dataset were annotated by medical students who had access to pathologist feedback, though the final annotations were not always reviewed by a pathologist (lower quality). Other subsets of the dataset were annotated by multiple non-experts and experts to enhance reliability.
For this study, we utilized the corrected single-rater subset, which represents the highest quality control level within the NuCLS dataset~\citep{nucls}. This subset includes annotations that have been corrected under the direct supervision of a pathologist. The training data comprises 124 whole slide images (WSIs), divided into 5 folds, with 15 WSIs reserved for testing. For each WSI, multiple crops have been annotated. Each crop has a resolution of $0.20~\si{\text{\textmu} m \per px}$. Annotations in the crops are provided within a field of view area as either fine-grained cell contours or bounding boxes. If a nucleus extends beyond the FOV boundary, its boundary or bounding box is extended into the surrounding area of the crop. Each FOV has an average width and height of 320 pixels, corresponding to a field size of approximately $65 \times 65~\si{\text{\textmu} m}$. The FOVs were annotated at high magnification to accurately indicate the location and classification of all nuclei within the area. For our analysis, we cut out the FOV area ($320~\si{px}$) from the crops and resized them to $256 \times 256~\si{px}$, resulting in a resolution of  $0.25~\si{\text{\textmu} m \per px}$.  We included only those cells whose center of mass fell within the FOV region. Additionally, we removed ambiguous nuclei labels. This process reduced the original dataset from $\num{59485}$ nuclei to $\num{48365}$ nuclei. An overview of the final dataset size and the distribution of nuclei among the main and super annotation classes is provided in the Supplementary Table \ref{tab:nucls_amount}. \\
The primary classes analyzed in our study include lymphocytes and plasma cells (superclass: sTILs), macrophages and stromal cells (superclass: stromal cells), mitotic and non-mitotic tumor cells (superclass: tumor cells), and miscellaneous cells. Our evaluation focuses on the performance of classifiers on these main classes within the NuCLS dataset. Additionally, we use the NuCLS test set to compare the performance of $\text{CellViT}^{{\scriptscriptstyle ++}}$ classifiers trained on the automatically derived SegPath dataset described below. For this comparison, we establish baseline results by training classifiers specifically on the lymphocyte and plasma cell classes, changing the tasks to a binary classification problem for each cell class, determining whether a cell belongs to the class or not. 

\subsubsection*{PanopTILs.} 
Introduced by~\citet{panoptils}, the PanopTILs dataset provides region and cell-level annotations for $\num{859759}$ nuclei from 151 breast cancer patients within the TCGA cohort. The dataset is particularly significant due to its focus on TILs, which hold substantial prognostic and predictive value in breast cancer. Despite their clinical importance, the visual assessment of TILs is highly subjective. Computational TIL scores have been shown to possess a higher prognostic value than traditional visual assessments, independent of TNM stage and patient age. This underscores the clinical relevance of the PanopTILs dataset and highlights its potential as a valuable resource for analyzing the breast tumor microenvironment~\citep{panoptils}. The dataset comprises annotations for several cell types, including TILs, stromal cells, epithelial cells, and miscellaneous cells, grouped according to the classification scheme suggested by~\citet{panoptils}. Although the dataset also includes region-wise annotations, our analysis focused solely on the nuclei labels. 

The PanopTILs dataset is divided into two parts: one for training and validation, and the other for testing. Training data includes manually defined regions with algorithmically bootstrapped nuclei labels, while the test set features manual regions with pathologist-approved nuclei annotations. In total, the training dataset consists of $\num{1709}$ image crops, each measuring $\num{1024} \times \num{1024}~\si{px}$, containing $\num{814886}$ annotated cells.  The test set, derived from the NuCLS dataset, includes only cells within a field of view (FOV), which is smaller than the $\num{1024} \times \num{1024}~\si{px}$ crop size used in the training data. Consequently, the evaluation of the test data is performed only on these FOV segments, not on the entire crops. The test set consists of $\num{1317}$ crops with $\num{44873}$ annotated cells. All images have a resolution of $0.25~\si{\text{\textmu} m \per px}$ at $\times 40$ magnification.
As a newly introduced dataset, PanopTILs provides a novel benchmark for the evaluation of cell detection performance, specifically within the context of breast cancer microenvironment analysis. Unlike prior studies~\citep{panoptils} that primarily focused on the classification accuracy of detected cells, our work contributes by establishing baseline results for TIL detection using PanopTILs.

\subsubsection*{SegPath.} 
The SegPath~\citep{segpath} dataset has a substantial difference to the previously listed datasets, as it is not a conventional annotated cell dataset. Unlike traditional datasets, where ground truth segmentation masks are manually derived from pathologists' annotations, SegPath’s segmentation masks were automatically generated through the registration of IF stainings. Indeed, the dataset does not provide cell level segmentations, but rather region-wise segmentation masks. It was developed for the semantic segmentation of H\&E stained images, specifically focusing on eight major cell types within tumor tissues. The dataset acquisition process was fully automated. First, tissue sections were stained with H\&E and subsequently digitized using a slide scanner. These sections were then destained through alcohol and autoclave processing, followed by IF staining with 4',6-diamidino-2-phenylindole dihydrochloride (DAPI) nuclear staining~\citep{segpath}. Specific antibodies were used to identify each cell type. The slides were digitized again after IF staining. To ensure precise alignment of the hematoxylin component in the H\&E images with the DAPI channel in the IF images, multiresolution rigid and non-rigid registration steps were applied. Binary segmentation masks were then generated based on the IF intensity, followed by morphological erosion~\citep{segpath}.
With this procedure, the authors were able to create a dataset consisting of $\num{158687}$ registered H\&E and IF patches, each with a size of $984 \times 984$ pixels at a resolution of $0.22~\si{\text{\textmu} m \per px}$ at $\times 40$ magnification. The dataset includes segmentation masks for epithelium (antibody: Pan-cytokeratin), smooth muscle/myofibroblast (antibody: $\alpha\text{SMA}$), lymphocytes (antibodies: CD3/CD20), leukocytes (antibody: CD45RB), blood/lymphatic vessels (antibody: ERG), plasma cells (antibody: MIST1), myeloid cells (antibody: MNDA), and red blood cells (antibody: CD235a) across 20 different organs. However, it does not contain cell-level instance segmentations. To derive a cell-level dataset of the H\&E-slides from the antibody IF segmentation masks, we applied $\text{CellViT}^{{\scriptscriptstyle ++}}$ to the H\&E patches and transferred the segmentation results onto the IF masks. A cell was considered positive if the detected cell had more than 15\% overlap with the antibody mask; otherwise, it was assigned the negative class label. This allows for the rapid creation of large-scale cell datasets. A critical aspect of this method is the selection of antibodies with high specificty. The chosen antibody must effectively stain the nuclei while avoiding excessive staining of cytoplasm areas. The antibodies CD3/CD20 (for lymphocytes) and MIST1 (for plasma cells) proved particularly suitable, as they distinctly marked the nuclei and provided clear nuclear staining. \\
To assess the performance of classifiers trained on these automatically derived cell datasets, we focused on the breast cancer subset of SegPath, specifically targeting lymphocyte and plasma cell data. The classifiers were trained on this subset and evaluated using the test set from the NUCLS dataset, which includes manual annotations supervised by pathologists. Using the $\text{CellViT}^{{\scriptscriptstyle ++}}$ approach, we extracted over $\num{5000}$ lymphocyte cells from 220 H\&E patches and $\num{27000}$ plasma cells from $\num{2054}$ patches within the SegPath training dataset for breast cancer, on which the classifiers can be trained.

\subsection*{Metrics}
\subsubsection*{Nuclei Detection.} 
To evaluate the performance of the model for the case of cell detection, we applied the metrics precision, recall, and $\text{F}_1$-score. Given that the detection of cells also involves classifying the cells into the correct cell class, our calculations account for the correct assignment of classes. The scores are first computed separately for each class and then averaged across them. Following the approach of~\citet{ocelot}, these metrics are referred to as mean precision (mPrec), mean recall (mRec), and mean $\text{F}_1$-score ($\text{mF}_1$). In the context of cell detection, the metrics can be interpreted as follows. Precision measures the accuracy of the detected cells relative to the ground truth. It is defined as the ratio of true positive (TP) detections (i.e., correclty detected cells with corresponding ground-truth (GT)) to the total number of cells detected by the model including both TP and false positives (FP). False positive cells are cells that do not match any GT cell. A high precision indicates that most of the detected cells are correct, with a low amount of FP cells. Conversely, if precision is low, the model detects many FP cell are indeed not present in the specimen. Recall, on the other hand, measures the model's ability to detect cells in the specimen. It is defined as the ratio of the TP cells to the the total number of cells in the ground-truth of the slide indicated by the sum of TP and false negatives (FN). Models with a high recall detect most of the present cells, whereas models with a low recall miss many cells. The $\text{F}_1$-Score combines precision and recall by calculating the harmonic mean of both metrics. It is a preferred metric in many classification and detection tasks because it effectively balances the trade-off between precision and recall to capture both aspects. \\
In the case of cell detection, the challenge remains in determining when a cell can be considered detected. To ensure consistency with previous works, we adopt the approach of~\citet{ocelot}. A predicted cell is considered as TP if an annotated cell of the same class is within a distance of 15 pixels, i.e., $3-4~\si{\text{\textmu} m}$ depending on the baseline resolution ($0.20-0.27~\si{\text{\textmu} m \per px}$ at $\times 40$ magnification). Otherwise, the detected cell is considered FP. If an annotated cell cannot be matched to a prediction, it is considered a FN cell. To calculate the $\text{mF}_1$-score, each cell class score is calculated with
\begin{align*}
    \text{F}_1 &= 2 \times \frac{\text{Precision} \times \text{Recall}}{\text{Precision} + \text{Recall}} \\
    \text{Precision} &= \frac{\text{TP}}{\text{TP} + \text{FP}} \\
    \text{Recall} &= \frac{\text{TP}}{\text{TP} + \text{FN}}
\end{align*}
Further details on the algorithmic implementation can be found in the original publication on the Ocelot dataset~\citep{ocelot}.

\subsubsection*{Nuclei Instance Segmentation.}
The detection metrics described above, while well-suited for comparing the classification and detection quality of networks, are insufficient for assessing the segmentation quality of cells. For many applications, it is crucial that the contours of cells are accurately captured, particularly for algorithms that rely on classical cell features. Therefore, in addition to the precision, recall, and $\text{F}_1$-score metrics, it is necessary to include metrics that reflect segmentation quality. In this study, we use the Panoptic Quality (PQ), specifically the binary PQ (bPQ) and mean PQ (mPQ) introduced by~\citet{hovernet}. These metrics consider both the segmentation quality of each nucleus and the quality of distinguishing between nuclei, including their class assignment. It is defined as
\begin{align*}
    PQ=\underbrace{\frac{\left | TP \right |}{\left | TP \right | + \frac{1}{2}\left | FP \right | + \frac{1}{2}\left | FN \right |}}_{\text{Detection Quality (DQ)}} \times \underbrace{\frac{\sum_{(y,\hat{y})\in{TP}} IoU(y,\hat{y})}{\left |  TP \right |}}_{\text{Segmentation Quality (SQ)}}. 
\end{align*}
In this equation $IoU(y,\hat{y})$ denotes the intersection-over-union, with $y$ beeing a GT segment $\hat{y}$ a predicted segment~\citep{kirillov_pq, CellViT}.  The calculation of true positives (TP), false positives (FP), and false negatives (FN) follows the same principles as for the detection metrics. \\
For bPQ, it is assumed that all cells belong to the same class, allowing for an assessment of how well nuclei are detected and segmented. The mPQ, in contrast, also considers the class assignment of the nuclei. We utilize the implementation provided in the original HoVer-Net publication~\citep{hovernet, CellViT}. \\
Since the mPQ score is calculated by first determining the PQ value for each image and each class before averaging, this approach can lead to certain classes being excluded from the results if they were predicted but not present in image annotations. This limitation arises from the inherent calculation method described by~\citet{hovernet}. To address this issue, they introduced the $\text{mPQ+}$ metric, which calculates the metrics across all images first and then averages them by class~\citep{cerberus}. This modification ensures that all classes are accounted for, regardless of their presence in the ground truth annotations. To avoid confusion, we explicitly specify whether the mPQ or $\text{mPQ+}$ metrics is being reported in our results. We consistently use the metric employed in the original dataset publication to ensure the comparability of our results with comparative methods.
\vspace{+5mm}
\subsection*{Carbon footprint calculation\\}The Carbon footprint and energy consumption estimations were conducted using the \href{https://mlco2.github.io/impact#compute}{Machine Learning Impact calculator} presented in~\cite{c02calculator}. 
Experiments were conducted on a private infrastructure with an estimated carbon efficiency of $0.432~\si{kg\,CO_2\,eq./kWH}$ (OECD's 2014 yearly average). For HoVer-Net on the CoNSeP dataset, we used the baseline configuration of 2 NVIDIA GeForce 1080 Ti GPUs, trained for 380 minutes (6.3 hours), resulting in an energy consumption of $3,170~\si{WH}$ (equivalent to $1.37~\si{kg\,CO_2\,eq.}$). This setup was scaled for the Lizard dataset, leading to a runtime of 28 hours and an energy consumption of approximately $14,000~\si{WH}$ ($6.04~\si{kg\,CO_2\,eq.}$). Using our hardware, for training on the CoNIC dataset, a cropped subset of the Lizard dataset prepared for HoVer-Net, we conducted training for 7.5 hours, consuming $3,000 ~\si{WH}$ ($1.30~\si{kg\,CO_2\,eq.}$), while training on the CoNSeP dataset took 102 minutes (1.7 hours), resulting in $680 ~\si{WH}$ ($0.29~\si{kg\,CO_2\,eq.}$). \\
Pretraining the largest \mbox{CellViT} model (ViT-H encoder) on the PanNuke dataset with a single NVIDIA A100 80GB GPU for 30 hours resulted in an energy consumption of  $12,000~\si{WH}$ ($5.18~\si{kg\,CO_2\,eq.}$). However, as demonstrated in our study, the model is a generalized cell segmentation model that can be adapted for new datasets by fine-tuning only the cell classification module based on the tokens. On average across all dataset, fine-tuning the cell classification module under the same setup took 20 minutes, including caching during the first epoch, consuming  $120~\si{WH}$ ($0.05~\si{kg\,CO_2\,eq.}$). Since the initial caching costs are incurred only during the first epoch in the first run, energy-efficient hyperparameter tuning can be performed, with an accumulated $\si{CO_2}$ equivalent of $0.92~\si{kg}$ ($2,120~\si{WH}$) for 100 experiments. We also determined specific values for the CoNSeP and Lizard datasets. Training the cell classification module took only 81 seconds on an NVIDIA A100 GPU, consuming $9.23~\si{WH}$ with a $\si{CO_2}$ equivalent of $4~\si{gr}$, which is negligible. For the Lizard dataset, the process took 12 minutes, consuming $92.16~\si{WH}$ ($40~\si{gr\,CO_2\,eq.}$). As~\citet{c02calc_source} highlighted, the development of a machine learning model consumes far more resources than the final training run, leading to a significantly higher overall energy consumption - a factor we attempt to address by simulating 100 training runs. This analysis shows that, even compared to a parameter-efficient cell segmentation method like HoVer-Net, our energy consumption is substantially reduced. \\
However, the carbon footprint of foundation models should not be overlooked. Thus, for transparency, we have also calculated the respective carbon footprint of the original publications. The SAM-H model was trained for 68 hours on 256 NVIDIA A100 80GB GPUs, consuming $6,963~\si{kWH}$ ($3,008~\si{kg\,CO_2\,eq.}$). Unfortunately, data for the Virchow and $\text{Virchow}_2$ models are not available, but we estimate that at least 48 NVIDIA V100 32GB GPUs were used for 200 hours (based on~\cite{ctranspath}), leading to a lower-bound estimate of  $3,600~\si{kWH}$ ($1,555~\si{kg\,CO_2\,eq.}$). The UNI model was trained with $4 \times 8$ Nvidia A100 80GBGPUs for 32 hours, consuming $410~\si{kWH}$ ($177~\si{kg\,CO_2\,eq.}$).  
Despite the significant carbon footprint, foundation models provide substantial advantages in terms of efficiency and versatility. When these models are fine-tuned and reused across multiple tasks, the overall carbon footprint can be significantly reduced compared to re-training specialized models from scratch for each new application.

\subsection*{Software\\}
The \mbox{CellViT} code, as initially described in~\cite{CellViT}, was modified to enhance both inference speed and the flexibility of the encoder network, allowing for the integration of the Virchow and $\text{Virchow}_{2}$ ViT architectures. During inference, the tool for slide processing is Pathopatch (v. 1.0.4b0), which integrates OpenSlide (v. 4.0.0), wsidicom (v. 0.20.4), and CUCIM (v. 24.04.00). This combination of tools allowed us to preprocess slides from a variety of open-source and vendor formats, including DICOM.
All experiments and analyses were conducted using Python (v. 3.10.14). The reproducibility of these experiments is ensured through the use of open-source libraries, with the most important libraries specified as follows: PyTorch (v. 2.2.1) along with its associated libraries torchvision (v. 0.17.1) and torchmetrics (v. 0.11.4), numpy (v. 1.23.5), numba (v. 0.59.0), CuPY (v. 13.0.0), CUCIM (v. 24.04.00), OpenSlide (v. 4.0.0), pandas (v. 1.4.3), scikit-learn (v. 1.3.0), PyCaret (v. 3.3.2) and CatBoost (v. 1.2.5). The implementations of the foundation Models used in this study were sourced from the following repositories: UNI, available at \href{https://huggingface.co/MahmoodLab/UNI}{https://huggingface.co/MahmoodLab/UNI}; Virchow, available at \href{https://huggingface.co/paige-ai/Virchow}{https://huggingface.co/paige-ai/Virchow}; Virchow2, available at \href{https://huggingface.co/paige-ai/Virchow2}{https://huggingface.co/paige-ai/Virchow2}; HIPT, available at \href{https://github.com/mahmoodlab/HIPT}{https://github.com/mahmoodlab/HIPT}; and Segment Anything, available at \href{https://github.com/facebookresearch/segment-anything}{https://github.com/facebookresearch/segment-anything}. A full list is given in the Supplementary Table~\ref{tab:appendix_identifier_overview}.

\subsection*{Data availability\\}
The study is based on public available datasets. The datasets have been acquired from the following sources: Ocelot~\citep{ocelot}, available at \href{https://ocelot2023.grand-challenge.org/}{https://ocelot2023.grand-challenge.org/}; Midog $++$~\citep{midogpp}, available at \href{https://doi.org/10.6084/m9.figshare.c.6615571.v1}{https://doi.org/10.6084/m9.figshare.c.6615571.v1}; CoNSeP~\citep{hovernet}, available at \href{https://warwick.ac.uk/fac/cross_fac/tia/data/}{https://warwick.ac.uk/fac/cross\_fac/tia/data/}; Lizard~\citep{lizard}, available at \href{https://warwick.ac.uk/fac/cross_fac/tia/data/}{https://warwick.ac.uk/fac/cross\_fac/tia/data/}; NuCLS ~\citep{nucls},  available at \href{https://sites.google.com/view/nucls/}{https://sites.google.com/view/nucls/}; SegPath~\citep{segpath} available at \href{https://dakomura.github.io/SegPath/}{https://dakomura.github.io/SegPath/}; and PanopTILs~\citep{panoptils} available at \href{https://sites.google.com/view/panoptils/}{https://sites.google.com/view/panoptils/}.  A full list is given in the Supplementary Table~\ref{tab:appendix_cell_dataset_overview}. 

\subsection*{Code availability\\}
The source code of $\text{CellViT}^{{\scriptscriptstyle ++}}$ is available at \href{https://github.com/TIO-IKIM/CellViT-plus-plus}{https://github.com/TIO-IKIM/CellViT-plus-plus}. 

\clearpage
\appendix
\newcommand{\customcaption}[2]{\noindent\textbf{#1:} #2}
\captionsetup[figure]{name=}
\renewcommand{\thefigure}{Extended Data Fig. \arabic{figure}}
\renewcommand{\thetable}{S\arabic{table}}  
\setcounter{figure}{0}
\setcounter{table}{0}

\onecolumn
\section*{Extended Data Figures}
\begin{figure}[!h]
    \centering
    \includegraphics[width=\linewidth]{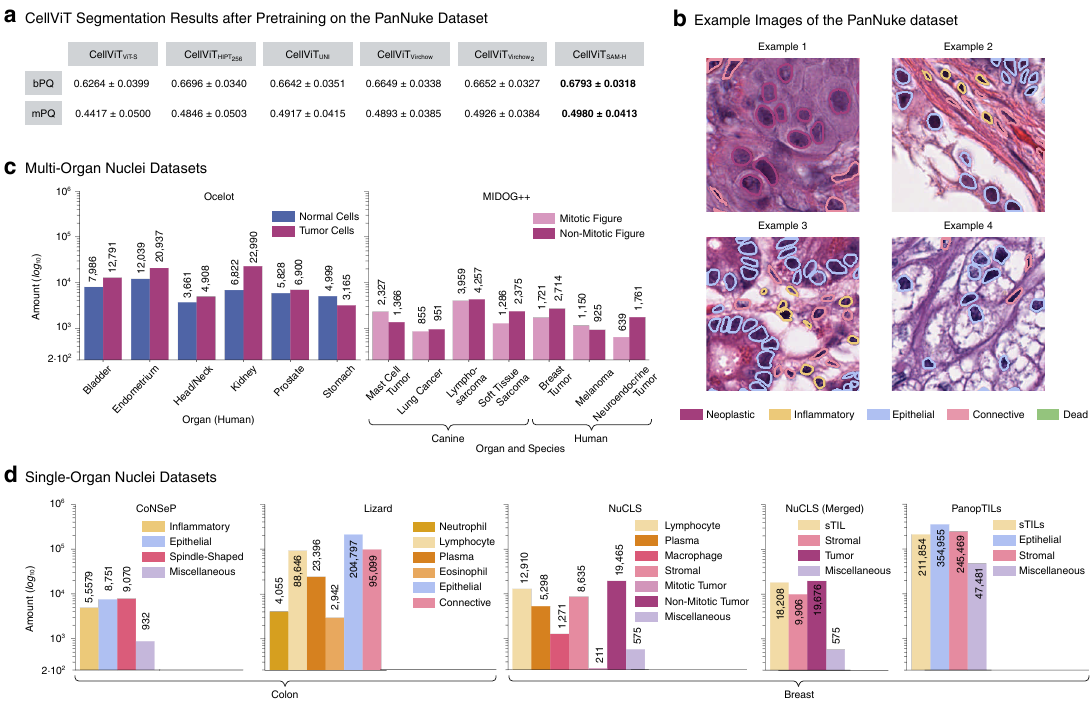}
    \caption{\textbf{Pretraining results and overview of the datasets used in our work}. \textbf{a} Panoptic quality (PQ) of each $\text{CellViT}^{{\scriptscriptstyle ++}}$ Variant on the PanNuke dataet using different foundation models as image encoders. Metrics are splitted into binary PQ (bPQ) and mean PQ (mPQ) taking the PanNuke nuclei classification into account. \textbf{b} Example training patches ($256 \times 256~\si{px}$) for the segmentation heads of the PanNuke dataset. \textbf{c} Multi organ nuclei datasets for tumor cell detection (Ocelot) and mitosis detection (MIDOG++). \textbf{d} Single organ datasets. We compared our models extensively on cells from two organs: Breast and colon.
    }
    \label{fig:datset_overview}
    \phantomsubcaption
    \label{fig:datset_overview_pannuke_results}
    \phantomsubcaption
    \label{fig:datset_overview_pannuke_images}
    \phantomsubcaption
    \label{fig:datset_overview_multi_tissue}
    \phantomsubcaption
    \label{fig:datset_overview_single_tissue}
\end{figure}

\begin{figure}[!h]
    \centering
    \includegraphics[width=\linewidth]{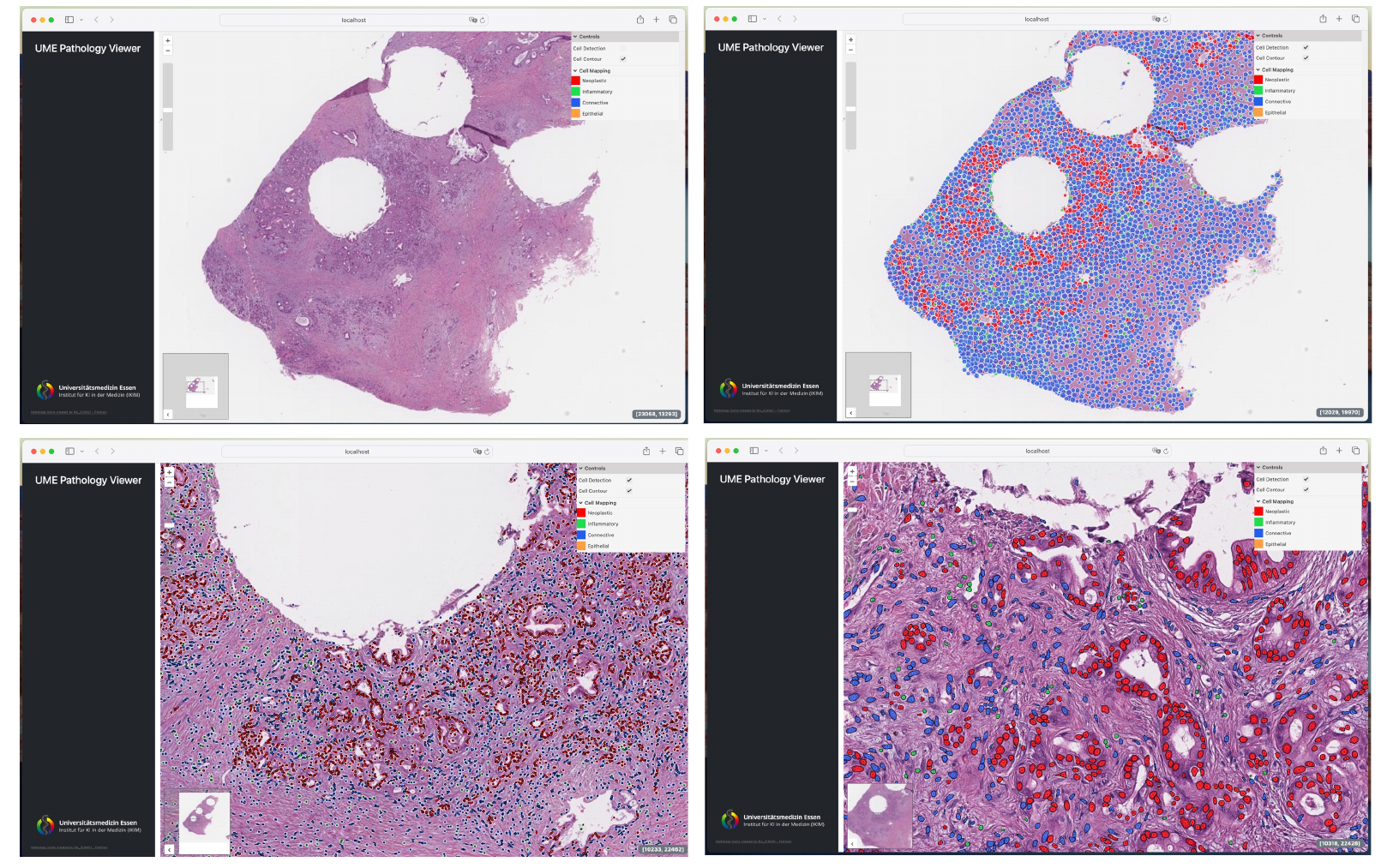}
    \caption{New web-based WSI viewer demonstrating visualization capabilities without local software installation using modern web technologies. (Upper left) General overview of the WSI. (Upper right) Overview with cell detections, clustered to depict spatial distribution. (Lower left) Zoomed-in view of a selected region with detailed cell detections. (Lower right) High-resolution view with cell contour overlays.}
    \label{fig:extended_viewer}
\end{figure}

\begin{figure}
    \centering
    \includegraphics[width=\linewidth]{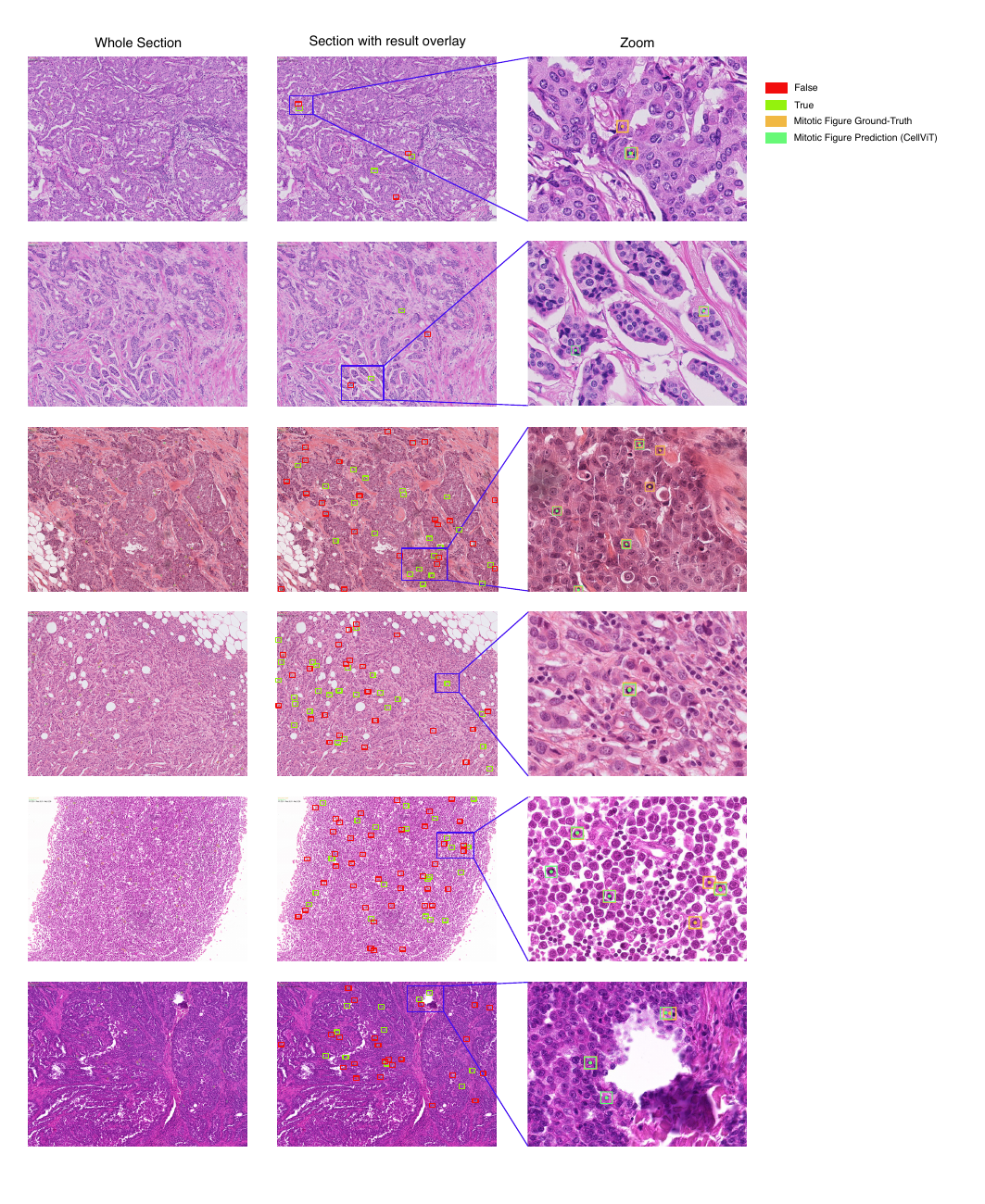}
    \caption{Representative tissue sections from the MIDOG++ dataset, illustrating the challenge of mitotic figure detection as a "needle in a haystack" problem. Each WSI section is annotated with ground truth (GT) mitotic figures and corresponding predictions generated by the $\text{CellViT}^{{\scriptscriptstyle ++}}_\text{SAM-H}$ model.}
    \label{fig:midog_example}
\end{figure}
\clearpage

\section*{Supplementary Tables}

\begin{table}[!h]
\centering
\label{tab:appendix_pannuke_mpq_organ}
\caption{Average mPQ and bPQ scores across the 19 tissue types of the PanNuke dataset for 3-fold CV for all \mbox{CellViT} variants. The standard deviation (SD) of the splits is provided in the final row. Best results are marked bold, second best underlined.}
\resizebox{\textwidth}{!}{%
\begin{tabular}{@{}llllllllllllllll@{}}
\toprule
     &  \multicolumn{2}{l}{$\text{CellViT}_{\text{HIPT}_{256}}$} &  & \multicolumn{2}{l}{$\text{CellViT}_\text{UNI}$} &  & \multicolumn{2}{l}{$\text{CellViT}_\text{Virchow}$} &  & \multicolumn{2}{l}{$\text{CellViT}_{\text{Virchow}_2}$} &  & \multicolumn{2}{l}{$\text{CellViT}_\text{SAM-H}$} \\ \midrule
           
Tissue            & $mPQ$                       & $bPQ$               &  & $mPQ$                & $bPQ$                      &  & $mPQ$                   & $bPQ$                    &  & $mPQ$                & $bPQ$                     &  & $mPQ$                    & $bPQ$                   \\ \midrule
Adrenal           & 0.4950                      & \underline{0.7009}  &  & 0.5033               & 0.6939                     &  & \underline{0.5087}      & 0.6979                   &  & 0.5028               & 0.6960                    &  & \textbf{0.5134}          & \textbf{0.7086}         \\
Bile Duct         & 0.4721                      & \underline{0.6705}  &  & 0.4736               & 0.6592                     &  & 0.4763                  & 0.6643                   &  & \underline{0.4820}   & 0.6729                    &  & \textbf{0.4887}          & \textbf{0.6784}         \\
Bladder           & 0.5756                      & \underline{0.7056}  &  & \underline{0.5789}   & 0.6980                     &  & 0.5599                  & 0.6958                   &  & 0.5595               & 0.6898                    &  & \textbf{0.5844}          & \textbf{0.7068}         \\
Breast            & \underline{0.5089}          & \underline{0.6641}  &  & 0.5006               & 0.6547                     &  & 0.4999                  & 0.6562                   &  & 0.5043               & 0.6563                    &  & \textbf{0.5180}          & \textbf{0.6748}         \\
Cervix            & 0.4893                      & \underline{0.6862}  &  & \underline{0.4911}   & 0.6815                     &  & 0.4908                  & 0.6766                   &  & 0.4849               & 0.6724                    &  & \textbf{0.4984}          & \textbf{0.6872}         \\
Colon             & 0.4245                      & 0.5700              &  & 0.4392               & 0.5729                     &  & \underline{0.4398}      & \underline{0.5730}       &  & 0.4377               & 0.5728                    &  & \textbf{0.4485}          & \textbf{0.5921}         \\
Esophagus         & \underline{0.5373}          & \underline{0.6619}  &  & 0.5267               & 0.6525                     &  & 0.5352                  & 0.6573                   &  & 0.5325               & 0.6544                    &  & \textbf{0.5454}          & \textbf{0.6682}         \\
Head \& Neck      & 0.4901                      & \underline{0.6472}  &  & 0.4827               & 0.6316                     &  & 0.4709                  & 0.6361                   &  & \underline{0.4904}   & 0.6395                    &  & \textbf{0.4913}          & \textbf{0.6544}         \\
Kidney            & 0.5409                      & 0.6993              &  & \textbf{0.5610}      & \underline{0.7094}         &  & 0.5483                  & 0.6940                   &  & \underline{0.5602}   & 0.6975                    &  & 0.5366                   & \textbf{0.7092}         \\
Liver             & 0.5065                      & 0.7160              &  & \underline{0.5140}   & \underline{0.7180}         &  & 0.5078                  & \underline{0.7180}       &  & 0.5111               & 0.7158                    &  & \textbf{0.5224}          & \textbf{0.7322}         \\
Lung              & 0.4102                      & \underline{0.6317}  &  & \underline{0.4235}   & 0.6250                     &  & 0.4228                  & 0.6205                   &  & \textbf{0.4314}      & 0.6308                    &  & \textbf{0.4314}          & \textbf{0.6426}         \\
Ovarian           & \underline{0.5260}          & 0.6596              &  & 0.5218               & 0.6564                     &  & 0.5185                  & 0.6529                   &  & 0.5207               & \underline{0.6620}        &  & \textbf{0.5390}          & \textbf{0.6722}         \\
Pancreatic        & \underline{0.4769}          & \underline{0.6643}  &  & 0.4723               & 0.6530                     &  & 0.4700                  & 0.6604                   &  & \textbf{0.4924}      & 0.6626                    &  & 0.4719                   & \textbf{0.6658}         \\
Prostate          & 0.5164                      & \underline{0.6695}  &  & \underline{0.5247}   & 0.6666                     &  & 0.5169                  & 0.6682                   &  & 0.5182               & 0.6640                    &  & \textbf{0.5321}          & \textbf{0.6821}         \\
Skin              & 0.3661                      & 0.6400              &  & 0.4342               & 0.6320                     &  & \textbf{0.4442}         & \underline{0.6425}       &  & \underline{0.4384}   & 0.6327                    &  & 0.4339                   & \textbf{0.6565}         \\
Stomach           & \underline{0.4475}          & \underline{0.6918}  &  & 0.4463               & 0.6896                     &  & 0.4354                  & 0.6827                   &  & 0.4405               & 0.6851                    &  & \textbf{0.4705}          & \textbf{0.7022}         \\
Testis            & \underline{0.5091}          & \underline{0.6883}  &  & 0.5024               & 0.6760                     &  & 0.5046                  & 0.6794                   &  & 0.5044               & 0.6758                    &  & \textbf{0.5127}          & \textbf{0.6955}         \\
Thyroid           & 0.4412                      & 0.7035              &  & 0.4596               & 0.7018                     &  & \textbf{0.4630}         & \underline{0.7080}       &  &\underline{0.4598}    & 0.7056                    &  & 0.4519                   & \textbf{0.7151}         \\
Uterus            & 0.4737                      & 0.6516              &  & \underline{0.4863}   & 0.6481                     &  & 0.4838                  & 0.6497                   &  & \textbf{0.4873}      & \underline{0.6526}        &  & 0.4737                   & \textbf{0.6625}         \\ \midrule
Average           & 0.4846                      & \underline{0.6696}  &  & 0.4917               & 0.6642                     &  & 0.4893                  & 0.6649                   &  & \underline{0.4926}   & 0.6652                    &  & \textbf{0.4980}          & \textbf{0.6793}         \\
SD                & 0.0503                      & 0.0340              &  & 0.0415               & 0.0351                     &  & 0.0385                  & 0.0338                   &  & 0.0384               & 0.0327                    &  & 0.0413                   & 0.0318                  \\ \bottomrule
\end{tabular}%
}
\end{table}

\begin{table}[]
    \centering
    \caption{CellViT-Backbone comparison (mPQ) on the PanNuke dataset, split by cell-type. Average results of the official 3-fold CV split. Best results are marked bold, second best underlined.
    }
    \label{tab:appendix_pannuke_mpq_nuclei}
    \resizebox{0.75\textwidth}{!}{%
        \begin{tabular}{lccccccccccccccc}
        \toprule
          {Model}                             & & & Neoplastic          &  & Epithelial        &  & Inflammatory         & & Connective           & &  Dead             \\ \midrule
          $\text{CellViT}_{\text{HIPT}_{256}}$              & & & 0.567               &  & 0.559             &  & \underline{0.405}                & & 0.405                & & 0.144             \\ 
          $\text{CellViT}_\text{UNI}$                         & & & 0.573               &  & 0.579             &  & 0.403                & & 0.408                & & \underline{0.152}             \\ 
          $\text{CellViT}_\text{Virchow}$                     & & & 0.577               &  & \underline{0.580}             &  & 0.393                & & 0.409                & & 0.147             \\ 
          $\text{CellViT}_{\text{Virchow}_2}$        & & & \underline{0.578}               &  & \underline{0.580}             &  & 0.403                & & \underline{0.410}                & & \textbf{0.154}    \\ 
          $\text{CellViT}_\text{SAM-H}$                       & & & \textbf{0.581}      &  & \textbf{0.583}    &  & \textbf{0.417}       & & \textbf{0.423}       & & 0.149             \\ \bottomrule
        \end{tabular}
    }
\end{table}
\twocolumn

\clearpage
\begin{table*}[!t]
    \centering
    \caption{Overview of the Ocelot dataset splits and amount of nuclei for the training, validation and test split.}
    \label{tab:appendix_ocelot_cell_amount}
    \resizebox{0.7\linewidth}{!}{%
        \begin{tabular}{@{}lrrrrrrrr@{}}
        \toprule
                        & \multicolumn{6}{r}{\textbf{Training}}       & \textbf{Validation} & \textbf{Test} \\ \cmidrule(lr){2-7}
        Amount          & 5\%  & 10\% & 25\%  & 50\%  & 75\%  & 100\% &                     &               \\ \midrule
        Tumor Cells     & 1,571 & 5,056 & 12,607 & 22,994 & 32,766 & 42,505 & 16,312               & 12,874         \\
        Non-Tumor Cells & 1,644 & 2,579 & 5,966  & 10,695 & 17,982 & 23,332 & 8,400                & 9,603          \\ \midrule
        \textbf{Total}  & 3,215 & 7,635 & 18,573 & 33,689 & 50,748 & 65,837 & 24,712               & 22,477         \\ \bottomrule
        \end{tabular}
    }
\end{table*}

\begin{table*}[!t]
    \centering
    \caption{Average $\text{F}_1$-score of the baseline model (SoftCTM) and the best-performing $\text{CellViT}^{{\scriptscriptstyle ++}}$ model ($\text{CellViT}^{{\scriptscriptstyle ++}}_\text{SAM-H}$) across different training data sizes, averaged over 5 runs. The performance of $\text{CellViT}^{{\scriptscriptstyle ++}}_\text{SAM-H}$ is evaluated with and without data augmentation, with results provided for each organ along with the standard deviation.}
    \label{tab:appendix_ocelot_amount_results}
    \resizebox{\textwidth}{!}{%
    \begin{tabular}{@{}lrccccccc@{}}
    \toprule
                           & Organ                  & Bladder                    & Endometrium                & Head \& Neck               & Kidney                     & Prostate                   & Stomach                    & Average                     \\ \midrule
    Amount                 & Network                & $\text{F}_1 \pm \text{SD}$ & $\text{F}_1 \pm \text{SD}$ & $\text{F}_1 \pm \text{SD}$ & $\text{F}_1 \pm \text{SD}$ & $\text{F}_1 \pm \text{SD}$ & $\text{F}_1 \pm \text{SD}$ & $\text{mF}_1 \pm \text{SD}$ \\ \midrule
    \multirow{3}{*}{5\%}   & SoftCTM                & 0.2488 $\pm$ 0.0928        & 0.2532 $\pm$ 0.0969        & 0.1820 $\pm$ 0.0916        & 0.3388 $\pm$ 0.1403        & 0.1987 $\pm$ 0.0668        & 0.2458 $\pm$ 0.1144        & 0.2692 $\pm$ 0.1105         \\
                           & $\text{CellViT}^{{\scriptscriptstyle ++}}_\text{SAM-H}$          & 0.5854 $\pm$ 0.0084        & 0.6875 $\pm$ 0.0070        & 0.5758 $\pm$ 0.0072        & 0.5575 $\pm$ 0.0120        & 0.6269 $\pm$ 0.0116        & 0.6224 $\pm$ 0.0108        & 0.6291 $\pm$ 0.0065         \\ 
                           & $\text{CellViT}^{{\scriptscriptstyle ++}}_\text{SAM-H}$ (Aug)    & 0.5901 $\pm$ 0.0140        & 0.6870 $\pm$ 0.0186        & 0.5750 $\pm$ 0.0152        & 0.5731 $\pm$ 0.0256        & 0.6499 $\pm$ 0.0104        & 0.6329 $\pm$ 0.0156        & 0.6362 $\pm$ 0.0082	        \\ \midrule
    \multirow{3}{*}{10\%}  & SoftCTM                & 0.5783 $\pm$ 0.0170        & 0.6355 $\pm$ 0.0223        & 0.5639 $\pm$ 0.0303        & 0.5932 $\pm$ 0.0115        & 0.5599 $\pm$ 0.0178        & 0.6543 $\pm$ 0.0182        & 0.6100 $\pm$ 0.0173         \\
                           & $\text{CellViT}^{{\scriptscriptstyle ++}}_\text{SAM-H}$          & 0.5988 $\pm$ 0.0039        & 0.7104 $\pm$ 0.0061        & 0.5894 $\pm$ 0.0088        & 0.5931 $\pm$ 0.0069        & 0.6392 $\pm$ 0.0090        & 0.6589 $\pm$ 0.0080        & 0.6507 $\pm$ 0.0035         \\ 
                           & $\text{CellViT}^{{\scriptscriptstyle ++}}_\text{SAM-H}$ (Aug)    & 0.5735 $\pm$ 0.0244        & 0.7181 $\pm$ 0.0086        & 0.5707 $\pm$ 0.0094        & 0.6005 $\pm$ 0.0177        & 0.6503 $\pm$ 0.0176        & 0.6749 $\pm$ 0.0084        & 0.6507 $\pm$ 0.0092         \\ \midrule
    \multirow{3}{*}{25\%}  & SoftCTM                & 0.6693 $\pm$ 0.0084        & 0.7019 $\pm$ 0.0084        & 0.6176 $\pm$ 0.0312        & 0.6339 $\pm$ 0.0099        & 0.6340 $\pm$ 0.0093        & 0.7090 $\pm$ 0.0169        & 0.6736 $\pm$ 0.0037         \\
                           & $\text{CellViT}^{{\scriptscriptstyle ++}}_\text{SAM-H}$          & 0.6046 $\pm$ 0.0133        & 0.6964 $\pm$ 0.0111        & 0.5997 $\pm$ 0.0087        & 0.6081 $\pm$ 0.0157        & 0.6356 $\pm$ 0.0053        & 0.7080 $\pm$ 0.0084        & 0.6550 $\pm$ 0.0073         \\ 
                           & $\text{CellViT}^{{\scriptscriptstyle ++}}_\text{SAM-H}$ (Aug)    & 0.6091 $\pm$ 0.0066        & 0.7056 $\pm$ 0.0042        & 0.5574 $\pm$ 0.0176        & 0.6074 $\pm$ 0.0142        & 0.6371 $\pm$ 0.0223        & 0.7047 $\pm$ 0.0108        & 0.6539 $\pm$ 0.0073         \\ \midrule
    \multirow{3}{*}{50\%}  & SoftCTM                & 0.6797 $\pm$ 0.0051        & 0.7195 $\pm$ 0.0067        & 0.5385 $\pm$ 0.0272        & 0.6573 $\pm$ 0.0060        & 0.6619 $\pm$ 0.0096        & 0.7425 $\pm$ 0.0058        & 0.6820 $\pm$ 0.0048         \\
                           & $\text{CellViT}^{{\scriptscriptstyle ++}}_\text{SAM-H}$          & 0.6287 $\pm$ 0.0132        & 0.7123 $\pm$ 0.0110        & 0.6041 $\pm$ 0.0121        & 0.6327 $\pm$ 0.0109        & 0.6594 $\pm$ 0.0083        & 0.6968 $\pm$ 0.0043        & 0.6695 $\pm$ 0.0072         \\ 
                           & $\text{CellViT}^{{\scriptscriptstyle ++}}_\text{SAM-H}$ (Aug)    & 0.6378 $\pm$ 0.0093        & 0.7160 $\pm$ 0.0060        & 0.5734 $\pm$ 0.0271        & 0.6457 $\pm$ 0.0170        & 0.6635 $\pm$ 0.0208        & 0.6994 $\pm$ 0.0078        & 0.6716 $\pm$ 0.0073         \\ \midrule
    \multirow{3}{*}{75\%}  & SoftCTM                & 0.6899 $\pm$ 0.0094        & 0.7387 $\pm$ 0.0065        & 0.6001 $\pm$ 0.0333        & 0.6775 $\pm$ 0.0092        & 0.6384 $\pm$ 0.0286        & 0.7623 $\pm$ 0.0046        & 0.6986 $\pm$ 0.0038         \\
                           & $\text{CellViT}^{{\scriptscriptstyle ++}}_\text{SAM-H}$          & 0.6089 $\pm$ 0.0069        & 0.7168 $\pm$ 0.0058        & 0.6307 $\pm$ 0.0143        & 0.6235 $\pm$ 0.0121        & 0.6437 $\pm$ 0.0142        & 0.6950 $\pm$ 0.0089        & 0.6669 $\pm$ 0.0054         \\ 
                           & $\text{CellViT}^{{\scriptscriptstyle ++}}_\text{SAM-H}$ (Aug)    & 0.6293 $\pm$ 0.0071        & 0.7214 $\pm$ 0.0036        & 0.5898 $\pm$ 0.0192        & 0.6280 $\pm$ 0.0108        & 0.6544 $\pm$ 0.0173        & 0.6940 $\pm$ 0.0093        & 0.6694 $\pm$ 0.0053         \\ \midrule
    \multirow{3}{*}{100\%} & SoftCTM                & 0.6955 $\pm$ 0.0094        & 0.7459 $\pm$ 0.0074        & 0.6267 $\pm$ 0.0268        & 0.6779 $\pm$ 0.0207        & 0.6850 $\pm$ 0.0377        & 0.7660 $\pm$ 0.0066        & 0.7109 $\pm$ 0.0069         \\
                           & $\text{CellViT}^{{\scriptscriptstyle ++}}_\text{SAM-H}$          & 0.6276 $\pm$ 0.0085        & 0.7338 $\pm$ 0.0028        & 0.6359 $\pm$ 0.0095        & 0.6532 $\pm$ 0.0120        & 0.6542 $\pm$ 0.0091        & 0.7049 $\pm$ 0.0120        & 0.6827 $\pm$ 0.0028         \\ 
                           & $\text{CellViT}^{{\scriptscriptstyle ++}}_\text{SAM-H}$ (Aug)    & 0.6225 $\pm$ 0.0148        & 0.7311 $\pm$ 0.0076        & 0.6078 $\pm$ 0.0173        & 0.6374 $\pm$ 0.0106        & 0.6493 $\pm$ 0.0152        & 0.6825 $\pm$ 0.0110        & 0.6726 $\pm$ 0.0080	         \\ \bottomrule
    \end{tabular}%
    }
\end{table*}

\begin{table*}[]
    \centering
    \caption{Average $\text{F}_1$-score and standard deviation (SD) of the baseline model (SoftCTM) and all $\text{CellViT}^{{\scriptscriptstyle ++}}$ variants averaged over 5 runs when 100\% of the training data is used for the Ocelot dataset.}
    \label{tab:appendix_ocelot_all_networks}
    \resizebox{\textwidth}{!}{%
    \begin{tabular}{@{}lrccccccc@{}}
    \toprule
                                                    & Organ            & Bladder                    & Endometrium                & Head \& Neck               & Kidney                     & Prostate                   & Stomach                    & Average                     \\ \midrule
    Network                                         & Data Augmentation     & $\text{F}_1 \pm \text{SD}$ & $\text{F}_1 \pm \text{SD}$ & $\text{F}_1 \pm \text{SD}$ & $\text{F}_1 \pm \text{SD}$ & $\text{F}_1 \pm \text{SD}$ & $\text{F}_1 \pm \text{SD}$ & $\text{mF}_1 \pm \text{SD}$ \\ \midrule
    SoftCTM                                         & Yes              & 0.6955 $\pm$ 0.0094        & 0.7459 $\pm$ 0.0074        & 0.6267 $\pm$ 0.0268        & 0.6779 $\pm$ 0.0207        & 0.6850 $\pm$ 0.0377        & 0.7660 $\pm$ 0.0066        & 0.7109 $\pm$ 0.0069         \\ \midrule
    \multirow{2}{*}{$\text{CellViT}^{{\scriptscriptstyle ++}}_{\text{HIPT}_{256}}$}    & No               & 0.5832 $\pm$ 0.0023        & 0.7129 $\pm$ 0.0028        & 0.6053 $\pm$ 0.0041        & 0.5809 $\pm$ 0.0026        & 0.6158 $\pm$ 0.0030        & 0.6291 $\pm$ 0.0039        & 0.6426 $\pm$ 0.0024         \\
                                                    & Yes              & 0.5855 $\pm$ 0.0059        & 0.7109 $\pm$ 0.0065        & 0.6069 $\pm$ 0.0143        & 0.5820 $\pm$ 0.0108        & 0.6134 $\pm$ 0.0055        & 0.6281 $\pm$ 0.0164        & 0.6425 $\pm$ 0.0078	       \\ \midrule
    \multirow{2}{*}{$\text{CellViT}^{{\scriptscriptstyle ++}}_\text{UNI}$}                    & No               & 0.6337 $\pm$ 0.0020        & 0.6659 $\pm$ 0.0037        & 0.6399 $\pm$ 0.0040        & 0.6537 $\pm$ 0.0021        & 0.6158 $\pm$ 0.0054        & 0.6586 $\pm$ 0.0019        & 0.6537 $\pm$ 0.0019         \\
                                                    & Yes              & 0.6362 $\pm$ 0.0054        & 0.6673 $\pm$ 0.0037        & 0.6426 $\pm$ 0.0027        & 0.6581 $\pm$ 0.0032        & 0.6217 $\pm$ 0.0047        & 0.6598 $\pm$ 0.0059        & 0.6565 $\pm$ 0.0022	       \\ \midrule
    \multirow{2}{*}{$\text{CellViT}^{{\scriptscriptstyle ++}}_\text{Virchow}$}                & No               & 0.5775 $\pm$ 0.0085        & 0.6301 $\pm$ 0.0048        & 0.5939 $\pm$ 0.0095        & 0.6058 $\pm$ 0.0046        & 0.5477 $\pm$ 0.0056        & 0.5920 $\pm$ 0.0041        & 0.6073 $\pm$ 0.0023         \\
                                                    & Yes              & 0.5930 $\pm$ 0.0154        & 0.6386 $\pm$ 0.0071        & 0.5669 $\pm$ 0.0190        & 0.6091 $\pm$ 0.0076        & 0.5458 $\pm$ 0.0131        & 0.6034 $\pm$ 0.0048        & 0.6113 $\pm$ 0.0050	       \\ \midrule
    \multirow{2}{*}{CellViT-$\text{Virchow}_2$}     & No               & 0.5773 $\pm$ 0.0136        & 0.6541 $\pm$ 0.0095        & 0.5966 $\pm$ 0.0141        & 0.6303 $\pm$ 0.0134        & 0.5477 $\pm$ 0.0144        & 0.5827 $\pm$ 0.0051        & 0.6158 $\pm$ 0.0095         \\
                                                    & Yes              & 0.5561 $\pm$ 0.0061        & 0.6457 $\pm$ 0.0074        & 0.5958 $\pm$ 0.0200        & 0.6186 $\pm$ 0.0067        & 0.5438 $\pm$ 0.0066        & 0.5807 $\pm$ 0.0092        & 0.6071 $\pm$ 0.0032	       \\ \midrule
    \multirow{2}{*}{$\text{CellViT}^{{\scriptscriptstyle ++}}_\text{SAM-H}$}                  & No               & 0.6276 $\pm$ 0.0085        & 0.7338 $\pm$ 0.0028        & 0.6359 $\pm$ 0.0095        & 0.6532 $\pm$ 0.0120        & 0.6542 $\pm$ 0.0091        & 0.7049 $\pm$ 0.0120        & 0.6827 $\pm$ 0.0028         \\
                                                    & Yes              & 0.6225 $\pm$ 0.0148        & 0.7311 $\pm$ 0.0076        & 0.6078 $\pm$ 0.0173        & 0.6374 $\pm$ 0.0106        & 0.6493 $\pm$ 0.0152        & 0.6825 $\pm$ 0.0110        & 0.6726 $\pm$ 0.0080	       \\  \bottomrule
    \end{tabular}%
    }
\end{table*}

\begin{table*}[h]
    \centering
    \begin{minipage}[t]{0.48\linewidth}
        \caption{Overview of the CoNSeP dataset splits and amount of nuclei for the entire training and test split as well as the tile level training subset.}
        \label{tab:appendix_consep_cell_amount_table}
        \resizebox{\linewidth}{!}{%
            \begin{tabular}{@{}rrrrr@{}}
            \toprule
            \multicolumn{1}{l}{} & \multicolumn{4}{c}{Nuclei Amount}                          \\ \cmidrule(l){2-5} 
            Tile Amount          & Inflammatory & Epithelial & Spindle-Shaped & Miscellaneous \\ \midrule
            1                    & 26           & 223        & 464            & 54            \\
            2                    & 30           & 711        & 714            & 151           \\
            3                    & 308          & 1,095       & 1,047           & 153           \\
            4                    & 340          & 1,095       & 1,145           & 168           \\
            5                    & 351          & 1,095       & 1,145           & 197           \\
            6                    & 446          & 1,382       & 1,604           & 206           \\
            7                    & 476          & 1,660       & 1,798           & 206           \\
            8                    & 550          & 1,993       & 2,128           & 210           \\
            9                    & 550          & 1,993       & 2,128           & 230           \\
            10                   & 782          & 1,993       & 2,542           & 231           \\
            11                   & 819          & 2,445       & 2,592           & 231           \\
            12                   & 852          & 2,445       & 2,719           & 232           \\
            13                   & 895          & 2,445       & 2,938           & 232           \\
            14                   & 927          & 3,173       & 2,974           & 241           \\
            15                   & 2,945         & 3,173       & 3,239           & 241           \\ \midrule
            All Training Tiles   & 3,941         & 5,537       & 5,706           & 371           \\
            All Test Tiles       & 1,638         & 3,214       & 3,364           & 561           \\ \bottomrule
            \end{tabular}
        }
    \end{minipage}%
    \hfill
    \begin{minipage}[t]{0.48\linewidth}
        \caption{Summary of cell annotations present in the Lizard dataset split by cell type and data source. Adopted from~\cite{lizard}.\\
        }
        \label{tab:appendix_lizard_amount}
        \resizebox{\linewidth}{!}{%
            \begin{tabular}{@{}lrrrrr@{}}
            \toprule
            Type/Dataset   & DigestPath & CRAG    & GlaS   & CoNSeP & \textbf{Total} \\ \midrule
            Epithelial     & 70,789     & 99,124  & 31,986 & 2,898  & 204,797        \\
            Lymphocyte     & 49,932     & 27,634  & 9,763  & 1,317  & 88,646         \\
            Plasma         & 11,352     & 9,363   & 2,349  & 332    & 23,396         \\
            Neutrophil     & 2,262      & 1,673   & 90     & 30     & 4,055          \\
            Eosinophil     & 1,349      & 1,255   & 286    & 52     & 2942           \\
            Connective     & 32,826     & 49,994  & 10,890 & 1,389  & 95,099         \\ \midrule
            \textbf{Total} & 168,510    & 189,043 & 55,364 & 6,018  & 418,935        \\ \bottomrule
            \end{tabular}
        }
    \end{minipage}
\end{table*}

\begin{table*}[]
    \centering
    \caption{Comparison of multiple baseline models on the CoNSeP dataset. All baseline models have been trained on the training set, with subsequent validation on the test set. We report the original publication scores if available, but we also re-trained the networks with the setup described in the original publication. The $\text{CellViT}^{{\scriptscriptstyle ++}}$ variants have all been trained and validated with 5 fold CV, with final evaluations of each fold on the test set. To estimate the average perfomance and distribution, we report the mean and standard deviation for our models.}
    \label{tab:appendix_consep_comparison_all_networks_global}
    \resizebox{\textwidth}{!}{%
        \begin{tabular}{@{}lllllllllllllll@{}}
        \toprule
        Score                        & \multicolumn{7}{c}{Binary Scores}                        &  & \multicolumn{6}{c}{Class-Averaged-Scores}                                                                                                                                      \\ \cmidrule(lr){2-8} \cmidrule(l){10-15} 
        Model                        & $\text{F}_1$-Score & DICE  & AJI   & AJI+  & bPQ   & bDQ   & bSQ   &  & $\text{mPQ} \pm \text{SD}$ & $\text{mDQ} \pm \text{SD}$ & $\text{mSQ} \pm \text{SD}$ & $\text{mPQ}+ \pm \text{SD}$ & $\text{mDQ+} \pm \text{SD}$ & $\text{mSQ+} \pm \text{SD}$ \\ \midrule
        HoVer-Net (Orig-Publication) & -        & 0.853 & 0.571 & -     & 0.547 & 0.702 & 0.778 &  & -                          & -                          & -                          & -                           & -                           & -                           \\
        HoVer-Net (PanNuke Baseline) & 0.691    & 0.802 & 0.492 & 0.524 & 0.461 & 0.609 & 0.755 &  & -                          & -                          & -                          & -                           & -                           & -                           \\
        HoVer-Net (self-trained)     & 0.731    & 0.836 & 0.535 & 0.563 & 0.505 & 0.656 & 0.767 &  & 0.364                      & 0.463                      & 0.712                      & 0.429                       & 0.550                       & 0.773                       \\
        Pointnu-Net (self-trained)   & 0.737    & 0.782 & 0.525 & 0.561 & 0.522 & 0.686 & 0.759 &  & 0.383                      & 0.495                      & 0.722                      & 0.446                       & 0.588                       & 0.752                       \\ \midrule
        $\text{CellViT}^{{\scriptscriptstyle ++}}_{\text{HIPT}_{256}}$  & 0.752    & 0.815 & 0.527 & 0.556 & 0.504 & 0.663 & 0.758 &  & 0.344 $\pm$ 0.014          & 0.447 $\pm$ 0.017          & 0.663 $\pm$ 0.007          & 0.398 $\pm$ 0.039           & 0.519 $\pm$ 0.053           & 0.761 $\pm$ 0.001           \\
        $\text{CellViT}^{{\scriptscriptstyle ++}}_\text{UNI}$                  & 0.720    & 0.818 & 0.525 & 0.552 & 0.492 & 0.649 & 0.756 &  & 0.341 $\pm$ 0.009          & 0.444 $\pm$ 0.011          & 0.669 $\pm$ 0.019          & 0.377 $\pm$ 0.026           & 0.491 $\pm$ 0.036           & 0.722 $\pm$ 0.073           \\
        $\text{CellViT}^{{\scriptscriptstyle ++}}_\text{Virchow}$              & 0.722    & 0.808 & 0.493 & 0.511 & 0.451 & 0.607 & 0.741 &  & 0.329 $\pm$ 0.009          & 0.432 $\pm$ 0.011          & 0.667 $\pm$ 0.006          & 0.386 $\pm$ 0.027           & 0.511 $\pm$ 0.038           & 0.749 $\pm$ 0.002           \\
        CellViT-$\text{Virchow}_2$   & 0.725    & 0.811 & 0.493 & 0.513 & 0.457 & 0.612 & 0.743 &  & 0.331 $\pm$ 0.013          & 0.436 $\pm$ 0.016          & 0.675 $\pm$ 0.018          & 0.359 $\pm$ 0.023           & 0.474 $\pm$ 0.032           & 0.750 $\pm$ 0.007           \\
        $\text{CellViT}^{{\scriptscriptstyle ++}}_\text{SAM-H}$                & 0.772    & 0.845 & 0.578 & 0.608 & 0.548 & 0.709 & 0.771 &  & 0.397 $\pm$ 0.004          & 0.507 $\pm$ 0.006          & 0.675 $\pm$ 0.008          & 0.461 $\pm$ 0.014           & 0.596 $\pm$ 0.018           & 0.768 $\pm$ 0.001           \\ \bottomrule
        \end{tabular}%
    }
\end{table*}

\begin{table*}[]
    \centering
    \caption{$\text{CellViT}^{{\scriptscriptstyle ++}}_\text{SAM-H}$ performance with and without data augmentation on the CoNSeP dataset. In this experiment, the models have been trained on limited training data, starting from one ROI up to 15 ROIs (compare Tab. \ref{tab:appendix_consep_cell_amount_table}). We report the performance among all cell types. For comparison, we include the baseline result with 100 \% training data in the bottom part.}
    \label{tab:appendix_consep_comparison_data_amount}
    \resizebox{\textwidth}{!}{%
        \begin{tabular}{@{}rrllllllllll@{}}
        \toprule
                            &                                       & Average                       & Inflammatory                  & Epithelium                   & Spindle-Shaped                & Miscellaneous                 \\ \midrule
        Num-Files           & \multicolumn{1}{l}{Data Augmentation} & $\text{mPQ}+ \pm \text{SD}$    & $\text{mPQ}+ \pm \text{SD}$    & $\text{mPQ}+ \pm \text{SD}$   & $\text{mPQ}+ \pm \text{SD}$    & $\text{mPQ}+ \pm \text{SD}$ \\ \midrule
        \multirow{2}{*}{1}  & No                                    & 0.326 $\pm$ 0.015                  & 0.336 $\pm$ 0.040                  & 0.467 $\pm$ 0.003                  & 0.404 $\pm$  0.002                  & 0.098  $\pm$ 0.020                  \\
                            & Yes                                   & 0.316 $\pm$ 0.033                  & 0.258 $\pm$ 0.059                  & 0.444 $\pm$ 0.037                  & 0.383 $\pm$  0.019                  & 0.180  $\pm$ 0.048                  \\
        \multirow{2}{*}{2}  & No                                    & 0.375 $\pm$ 0.015                  & 0.331 $\pm$ 0.048                  & 0.484 $\pm$ 0.005                  & 0.410 $\pm$  0.006                  & 0.273  $\pm$ 0.016                  \\
                            & Yes                                   & 0.357 $\pm$ 0.026                  & 0.235 $\pm$ 0.111                  & 0.479 $\pm$ 0.013                  & 0.392 $\pm$  0.010                  & 0.322  $\pm$ 0.020                  \\
        \multirow{2}{*}{3}  & No                                    & 0.441 $\pm$ 0.007                  & 0.521 $\pm$ 0.021                  & 0.503 $\pm$ 0.002                  & 0.427 $\pm$  0.002                  & 0.314  $\pm$ 0.012                  \\
                            & Yes                                   & 0.441 $\pm$ 0.015                  & 0.527 $\pm$ 0.042                  & 0.496 $\pm$ 0.010                  & 0.425 $\pm$  0.005                  & 0.317  $\pm$ 0.039                  \\
        \multirow{2}{*}{4}  & No                                    & 0.452 $\pm$ 0.005                  & 0.551 $\pm$ 0.013                  & 0.505 $\pm$ 0.000                  & 0.427 $\pm$  0.002                  & 0.326  $\pm$ 0.012                  \\
                            & Yes                                   & 0.441 $\pm$ 0.015                  & 0.531 $\pm$ 0.020                  & 0.503 $\pm$ 0.006                  & 0.426 $\pm$  0.003                  & 0.304  $\pm$ 0.052                  \\
        \multirow{2}{*}{5}  & No                                    & 0.443 $\pm$ 0.016                  & 0.529 $\pm$ 0.027                  & 0.503 $\pm$ 0.005                  & 0.425 $\pm$  0.004                  & 0.315  $\pm$ 0.034                  \\
                            & Yes                                   & 0.453 $\pm$ 0.011                  & 0.552 $\pm$ 0.038                  & 0.507 $\pm$ 0.003                  & 0.426 $\pm$  0.004                  & 0.329  $\pm$ 0.017                  \\
        \multirow{2}{*}{6}  & No                                    & 0.452 $\pm$ 0.007                  & 0.537 $\pm$ 0.017                  & 0.504 $\pm$ 0.002                  & 0.425 $\pm$  0.003                  & 0.342  $\pm$ 0.008                  \\
                            & Yes                                   & 0.460 $\pm$ 0.007                  & 0.552 $\pm$ 0.025                  & 0.509 $\pm$ 0.005                  & 0.428 $\pm$  0.004                  & 0.352  $\pm$ 0.008                  \\
        \multirow{2}{*}{7}  & No                                    & 0.459 $\pm$ 0.006                  & 0.561 $\pm$ 0.018                  & 0.505 $\pm$ 0.002                  & 0.428 $\pm$  0.001                  & 0.343  $\pm$ 0.008                  \\
                            & Yes                                   & 0.453 $\pm$ 0.017                  & 0.554 $\pm$ 0.020                  & 0.506 $\pm$ 0.006                  & 0.429 $\pm$  0.002                  & 0.323  $\pm$ 0.061                  \\
        \multirow{2}{*}{8}  & No                                    & 0.463 $\pm$ 0.006                  & 0.571 $\pm$ 0.017                  & 0.500 $\pm$ 0.005                  & 0.431 $\pm$  0.001                  & 0.349  $\pm$ 0.007                  \\
                            & Yes                                   & 0.459 $\pm$ 0.008                  & 0.553 $\pm$ 0.035                  & 0.509 $\pm$ 0.004                  & 0.431 $\pm$  0.005                  & 0.344  $\pm$ 0.013                  \\
        \multirow{2}{*}{9}  & No                                    & 0.460 $\pm$ 0.005                  & 0.562 $\pm$ 0.013                  & 0.503 $\pm$ 0.007                  & 0.430 $\pm$  0.004                  & 0.344  $\pm$ 0.010                  \\
                            & Yes                                   & 0.464 $\pm$ 0.005                  & 0.565 $\pm$ 0.020                  & 0.508 $\pm$ 0.003                  & 0.432 $\pm$  0.001                  & 0.350  $\pm$ 0.007                  \\
        \multirow{2}{*}{10} & No                                    & 0.468 $\pm$ 0.004                  & 0.580 $\pm$ 0.012                  & 0.509 $\pm$ 0.001                  & 0.432 $\pm$  0.001                  & 0.350  $\pm$ 0.006                  \\
                            & Yes                                   & 0.470 $\pm$ 0.003                  & 0.587 $\pm$ 0.004                  & 0.511 $\pm$ 0.001                  & 0.435 $\pm$  0.002                  & 0.348  $\pm$ 0.007                  \\
        \multirow{2}{*}{11} & No                                    & 0.462 $\pm$ 0.002                  & 0.567 $\pm$ 0.012                  & 0.506 $\pm$ 0.005                  & 0.433 $\pm$  0.002                  & 0.344  $\pm$ 0.005                  \\
                            & Yes                                   & 0.469 $\pm$ 0.003                  & 0.580 $\pm$ 0.003                  & 0.511 $\pm$ 0.002                  & 0.433 $\pm$  0.003                  & 0.353  $\pm$ 0.010                  \\
        \multirow{2}{*}{12} & No                                    & 0.468 $\pm$ 0.002                  & 0.583 $\pm$ 0.010                  & 0.508 $\pm$ 0.002                  & 0.431 $\pm$  0.003                  & 0.348  $\pm$ 0.007                  \\
                            & Yes                                   & 0.470 $\pm$ 0.006                  & 0.584 $\pm$ 0.012                  & 0.511 $\pm$ 0.002                  & 0.434 $\pm$  0.002                  & 0.349  $\pm$ 0.015                  \\
        \multirow{2}{*}{13} & No                                    & 0.467 $\pm$ 0.003                  & 0.576 $\pm$ 0.015                  & 0.510 $\pm$ 0.001                  & 0.433 $\pm$  0.001                  & 0.351  $\pm$ 0.008                  \\
                            & Yes                                   & 0.465 $\pm$ 0.005                  & 0.564 $\pm$ 0.019                  & 0.511 $\pm$ 0.002                  & 0.433 $\pm$  0.003                  & 0.351  $\pm$ 0.009                  \\
        \multirow{2}{*}{14} & No                                    & 0.468 $\pm$ 0.003                  & 0.584 $\pm$ 0.009                  & 0.509 $\pm$ 0.003                  & 0.432 $\pm$  0.002                  & 0.346  $\pm$ 0.013                  \\
                            & Yes                                   & 0.470 $\pm$ 0.004                  & 0.580 $\pm$ 0.015                  & 0.510 $\pm$ 0.001                  & 0.434 $\pm$  0.002                  & 0.356  $\pm$ 0.002                  \\
        \multirow{2}{*}{15} & No                                    & 0.466 $\pm$ 0.004                  & 0.572 $\pm$ 0.018                  & 0.510 $\pm$ 0.001                  & 0.432 $\pm$  0.002                  & 0.353  $\pm$ 0.006                  \\
                            & Yes                                   & 0.473 $\pm$ 0.002                  & 0.592 $\pm$ 0.006                  & 0.509 $\pm$ 0.001                  & 0.432 $\pm$  0.003                  & 0.357  $\pm$ 0.003                  \\ \midrule
        Baselines (100\% Train)          &                                       &                                    &                                    &                                    &                                     &                                     \\ \midrule
        $\text{CellViT}^{{\scriptscriptstyle ++}}_\text{SAM-H}$       & No                                    & 0.461 $\pm$ 0.014                  & 0.575 $\pm$ 0.010                  & 0.507 $\pm$ 0.003                  & 0.430 $\pm$  0.002                  & 0.330  $\pm$ 0.048                  \\
        $\text{CellViT}^{{\scriptscriptstyle ++}}_\text{SAM-H}$       & Yes                                   & 0.442 $\pm$ 0.030                  & 0.587 $\pm$ 0.007                  & 0.501 $\pm$ 0.008                  & 0.432 $\pm$  0.002                  & 0.247  $\pm$ 0.107                  \\
        HoVer-Net           & Yes                                   & 0.429                              & 0.596                              & 0.457                              & 0.381                               & 0.281                               \\
        PointNu-Net         & Yes                                   & 0.446                              & 0.596                              & 0.476                              & 0.407                               & 0.307                               \\ \bottomrule
        \end{tabular}%
    }
\end{table*}

\begin{table*}[]
    \centering
    \caption{Performance comparison on the Lizard dataset. We included the best available baseline models, all evaluated on a 3-Fold CV split. For our $\text{CellViT}^{{\scriptscriptstyle ++}}_\text{SAM-H}$ model, we further evaluate the performance when using classical feature engineering for extracting cellular features (histomics), including the 3-layer deep learning classifier and CatBoost. External validation on the test set cannot be conducted, as it remains hidden. HoVer-Net Cerberus refers to the scores published in~\citet{cerberus}, whereas HoVer-Net baseline refers to the original scores published by the Lizard authors.}
    \label{tab:appendix_lizard_results}
    \resizebox{\textwidth}{!}{%
        \begin{tabular}{@{}llrrrr@{}}
        \toprule
        Model                       & Classifier                                & Dice $\pm$ SD      & bPQ $\pm$ SD          & mPQ $\pm$ SD      & $\text{mPQ+}$ $\pm$ SD          \\ \midrule
        HoVer-Net (Baseline)        &                                           & 0.828 $\pm$ 0.008  & 0.624 $\pm$ 0.013     & 0.396 $\pm$ 0.022 & -                      \\
        HoVer-Net (Cerberus)         &                                           & -                  & 0.584 $\pm$ 0.014     & 0.295 $\pm$ 0.018 & 0.409 $\pm$ 0.027      \\
        Cerberus                    &                                           & -                  & 0.612 $\pm$ 0.010     & 0.358 $\pm$ 0.011 & 0.425 $\pm$ 0.019      \\
        CGIS-CPF                    &                                           & 0.889 $\pm$ 0.002  & 0.660 $\pm$ 0.061     & 0.421 $\pm$ 0.013 & -                      \\ \midrule
        $\text{CellViT}^{{\scriptscriptstyle ++}}_{\text{HIPT}_{256}}$                & DL, Token-based                           & 0.767 $\pm$ 0.005  & 0.513 $\pm$ 0.011     & 0.252 $\pm$ 0.003 & 0.272 $\pm$ 0.009      \\
        $\text{CellViT}^{{\scriptscriptstyle ++}}_\text{UNI}$                 & DL, Token-based                           & 0.752 $\pm$ 0.005  & 0.503 $\pm$ 0.009     & 0.279 $\pm$ 0.004 & 0.293 $\pm$ 0.004      \\
        $\text{CellViT}^{{\scriptscriptstyle ++}}_\text{SAM-H}$               & DL, Token-based                           & 0.774 $\pm$ 0.004  & 0.536 $\pm$ 0.009     & 0.294 $\pm$ 0.002 & 0.308 $\pm$ 0.006      \\
        $\text{CellViT}^{{\scriptscriptstyle ++}}_\text{SAM-H}$               & DL, Nuclei Features (Histomics)           & 0.774 $\pm$ 0.004  & 0.536 $\pm$ 0.009     & 0.255 $\pm$ 0.004 & 0.251 $\pm$ 0.003      \\
        $\text{CellViT}^{{\scriptscriptstyle ++}}_\text{SAM-H}$               & CatBoost, Nuclei Features (Histomics)     & 0.774 $\pm$ 0.004  & 0.536 $\pm$ 0.009     & 0.255 $\pm$ 0.009 & 0.255 $\pm$ 0.004      \\ \bottomrule
        \end{tabular}%
    }
\end{table*}

\begin{table*}[]
    \centering
    \caption{NuCLS nuclei amount for the main and super annotation classes within the corrected single annotator dataset used in this study. We excluded ambiguous labelled nuclei, as well as nuclei which center of mass lays outside the annotation field of view.}
    \label{tab:nucls_amount}
    \resizebox{0.75\linewidth}{!}{%
        \begin{tabular}{@{}lllllllllllc@{}}
        \toprule
        \multicolumn{2}{l}{\begin{tabular}[c]{@{}l@{}}Main \\ Annotations\end{tabular}}         & \multicolumn{1}{c}{\begin{tabular}[c]{@{}c@{}}Lympho-\\ cyte\end{tabular}} & \multicolumn{1}{c}{Plasma} &  & \multicolumn{1}{c}{\begin{tabular}[c]{@{}c@{}}Macro-\\ phage\end{tabular}} & \multicolumn{1}{c}{Stromal} &  & \multicolumn{1}{c}{\begin{tabular}[c]{@{}c@{}}Mitotic\\ Tumor\end{tabular}} & \multicolumn{1}{c}{\begin{tabular}[c]{@{}c@{}}Non-Mitotic \\ Tumor\end{tabular}} &  & Uncategorized        \\ \cmidrule(lr){3-4} \cmidrule(lr){6-7} \cmidrule(lr){9-10} \cmidrule(l){12-12} 
        \multicolumn{2}{l}{\begin{tabular}[c]{@{}l@{}}Merged Super \\ Annotations\end{tabular}} & \multicolumn{2}{c}{sTILs}                                                                               &  & \multicolumn{2}{c}{Stromal}                                                                              &  & \multicolumn{2}{c}{Tumor}                                                                                                                                      &  & Uncategorized        \\ \midrule
        \multicolumn{1}{r}{\multirow{2}{*}{Main}}           & \multicolumn{1}{r}{Train}         & \multicolumn{1}{r}{11,162}                                                 & \multicolumn{1}{r}{4,233}  &  & \multicolumn{1}{r}{1,153}                                                  & \multicolumn{1}{r}{7,214}   &  & \multicolumn{1}{r}{167}                                                     & \multicolumn{1}{r}{16,921}                                                       &  & 291                  \\
        \multicolumn{1}{r}{}                                & \multicolumn{1}{r}{Test}          & \multicolumn{1}{r}{1,748}                                                  & \multicolumn{1}{r}{1,065}  &  & \multicolumn{1}{r}{118}                                                    & \multicolumn{1}{r}{1,421}   &  & \multicolumn{1}{r}{44}                                                      & \multicolumn{1}{r}{2,544}                                                        &  & 284                  \\ \midrule
        \multicolumn{1}{r}{\multirow{2}{*}{Super}}          & \multicolumn{1}{r}{Train}         & \multicolumn{2}{c}{15,395}                                                                              &  & \multicolumn{2}{c}{8,367}                                                                                &  & \multicolumn{2}{c}{17,088}                                                                                                                                     &  & 291                  \\
        \multicolumn{1}{r}{}                                & \multicolumn{1}{r}{Test}          & \multicolumn{2}{c}{2,813}                                                                               &  & \multicolumn{2}{c}{1,539}                                                                                &  & \multicolumn{2}{c}{2,588}                                                                                                                                      &  & 284                  \\ \midrule
                                                            &                                   &                                                                            &                            &  &                                                                            &                             &  &                                                                             &                                                                                  &  & \multicolumn{1}{l}{} \\
                                                            &                                   &                                                                            &                            &  &                                                                            &                             &  &                                                                             &                                                                                  &  & \multicolumn{1}{l}{} \\
                                                            &                                   &                                                                            &                            &  &                                                                            &                             &  &                                                                             &                                                                                  &  & \multicolumn{1}{l}{} \\
                                                            &                                   &                                                                            &                            &  &                                                                            &                             &  &                                                                             &                                                                                  &  & \multicolumn{1}{l}{} \\
                                                            &                                   &                                                                            &                            &  &                                                                            &                             &  &                                                                             &                                                                                  &  & \multicolumn{1}{l}{}
        \end{tabular}
    }
\end{table*}

\begin{table*}[]
\centering
\caption{Averaged (SD) 5-Fold CV results on the NuCLS main label set (15 test WSI). The single-rater corrected dataset (correction by pathologists) has been used to assess performance.}
\label{tab:appendix_nucls_total_results_main_split}
\resizebox{\textwidth}{!}{%
\begin{tabular}{@{}llrrrrrrr@{}}
\toprule
                               &               & Lymphocyte        & Macrophage        & Uncategorized     & Plasma Cell       & Stromal           & Tumor (Non-Mitotic) & Tumor Mitotic     \\ \midrule
\multirow{3}{*}{$\text{CellViT}^{{\scriptscriptstyle ++}}_{\text{HIPT}_{256}}$}   & $\text{F}_1$ $\pm$ SD   & 0.654 $\pm$ 0.029 & 0.082 $\pm$ 0.056 & 0.029 $\pm$ 0.033 & 0.606 $\pm$ 0.082 & 0.670 $\pm$ 0.009 & 0.774 $\pm$ 0.011   & 0.008 $\pm$ 0.016 \\
                               & Prec $\pm$ SD & 0.528 $\pm$ 0.038 & 0.174 $\pm$ 0.108 & 0.248 $\pm$ 0.228 & 0.658 $\pm$ 0.016 & 0.593 $\pm$ 0.013 & 0.676 $\pm$ 0.018   & 0.029 $\pm$ 0.057 \\
                               & Rec $\pm$ SD  & 0.861 $\pm$ 0.008 & 0.054 $\pm$ 0.038 & 0.016 $\pm$ 0.019 & 0.578 $\pm$ 0.147 & 0.769 $\pm$ 0.007 & 0.905 $\pm$ 0.009   & 0.005 $\pm$ 0.009 \\
\multirow{3}{*}{$\text{CellViT}^{{\scriptscriptstyle ++}}_\text{UNI}$}   & $\text{F}_1$ $\pm$ SD   & 0.670 $\pm$ 0.022 & 0.239 $\pm$ 0.054 & 0.567 $\pm$ 0.073 & 0.570 $\pm$ 0.103 & 0.689 $\pm$ 0.010 & 0.801 $\pm$ 0.004   & 0.134 $\pm$ 0.049 \\
                               & Prec $\pm$ SD & 0.545 $\pm$ 0.030 & 0.325 $\pm$ 0.060 & 0.837 $\pm$ 0.074 & 0.718 $\pm$ 0.059 & 0.598 $\pm$ 0.028 & 0.714 $\pm$ 0.013   & 0.302 $\pm$ 0.102 \\
                               & Rec $\pm$ SD  & 0.871 $\pm$ 0.017 & 0.212 $\pm$ 0.089 & 0.446 $\pm$ 0.122 & 0.481 $\pm$ 0.121 & 0.816 $\pm$ 0.023 & 0.913 $\pm$ 0.011   & 0.100 $\pm$ 0.053 \\
\multirow{3}{*}{$\text{CellViT}^{{\scriptscriptstyle ++}}_\text{SAM-H}$} & $\text{F}_1$ $\pm$ SD   & 0.661 $\pm$ 0.016 & 0.102 $\pm$ 0.034 & 0.037 $\pm$ 0.027 & 0.544 $\pm$ 0.087 & 0.691 $\pm$ 0.006 & 0.800 $\pm$ 0.008   & 0.092 $\pm$ 0.092 \\
                               & Prec $\pm$ SD & 0.531 $\pm$ 0.021 & 0.336 $\pm$ 0.062 & 0.467 $\pm$ 0.241 & 0.675 $\pm$ 0.049 & 0.601 $\pm$ 0.015 & 0.719 $\pm$ 0.014   & 0.213 $\pm$ 0.198 \\
                               & Rec $\pm$ SD  & 0.878 $\pm$ 0.012 & 0.061 $\pm$ 0.023 & 0.020 $\pm$ 0.015 & 0.463 $\pm$ 0.110 & 0.814 $\pm$ 0.012 & 0.902 $\pm$ 0.003   & 0.059 $\pm$ 0.060 \\ \bottomrule
\end{tabular}%
}
\end{table*}

\begin{table*}[]
    \centering
    \caption{Performance of the single-cell type classifier trained on the automatically generated SegPath cell dataset, compared to NuCLS expert level annotations, with final evaluation on the NuCLS corrected single-rater test set for lymphocytes.}
    \label{tab:appendix_nucls_lympho}
    \resizebox{\textwidth}{!}{%
    \begin{tabular}{@{}lllllllllll@{}}
    \toprule
                                    &         & \multicolumn{1}{c}{}    &  & \multicolumn{3}{c}{Lymphocyte}                                   &  & \multicolumn{3}{c}{Other}                                                             \\ \cmidrule(lr){5-7} \cmidrule(l){9-11} 
        Dataset                     & Network & $\text{mF}_1$ $\pm$ SD            &  & $\text{F}_1$ $\pm$ SD           & Prec  $\pm$ SD      & Rec $\pm$ SD        &  & $\text{F}_1$ $\pm$ SD           & Prec $\pm$ SD     & Rec $\pm$ SD      \\ \cmidrule(r){1-7} \cmidrule(l){9-11} 
        \multirow{3}{*}{NuCLS Base} & SAM-H   & 0.769 $\pm$ 0.012       &  & 0.693 $\pm$ 0.020     & 0.595 $\pm$ 0.047   & 0.839 $\pm$ 0.052   &  & 0.846 $\pm$ 0.006     & 0.793 $\pm$ 0.021 & 0.908 $\pm$ 0.029 \\
                                    & $\text{HIPT}_{256}$  & 0.742 $\pm$ 0.011       &  & 0.652 $\pm$ 0.020     & 0.555 $\pm$ 0.037   & 0.795 $\pm$ 0.047   &  & 0.832 $\pm$ 0.005     & 0.764 $\pm$ 0.018 & 0.916 $\pm$ 0.021 \\
                                    & UNI     & 0.756 $\pm$ 0.004       &  & 0.673 $\pm$ 0.006     & 0.590 $\pm$ 0.018   & 0.787 $\pm$ 0.035   &  & 0.839 $\pm$ 0.003     & 0.767 $\pm$ 0.011 & 0.928 $\pm$ 0.011 \\ \midrule
        \multirow{3}{*}{SegPath}    & SAM-H   & 0.743                   &  & 0.651                 & 0.625               & 0.679               &  & 0.836                 & 0.751             & 0.943             \\
                                    & $\text{HIPT}_{256}$  & 0.713                   &  & 0.603                 & 0.490               & 0.783               &  & 0.824                 & 0.770             & 0.887             \\
                                    & UNI     & 0.738                   &  & 0.642                 & 0.532               & 0.809               &  & 0.834                 & 0.779             & 0.897             \\ \bottomrule
        \end{tabular}%
    }
\end{table*}

\begin{table*}[]
    \centering
    \caption{Performance of the single-cell type classifier trained on the automatically generated SegPath cell dataset, compared to NuCLS expert level annotations, with final evaluation on the NuCLS corrected single-rater test set for plasma cells.}
    \label{tab:appendix_nucls_plasma}
    \resizebox{\textwidth}{!}{%
        \begin{tabular}{@{}lllllllllll@{}}
        \toprule
                                    &         & \multicolumn{1}{c}{} &  & \multicolumn{3}{c}{Plasma Cell}                                &  & \multicolumn{3}{c}{Other}                                 \\ \cmidrule(lr){5-7} \cmidrule(l){9-11} 
        Dataset                     & Encoder & $\text{mF}_1$ $\pm$ SD         &  & $\text{F}_1$ $\pm$ SD       & Prec $\pm$ SD     & Rec $\pm$ SD      &  & $\text{F}_1$ $\pm$ SD       & Prec $\pm$ SD     & Rec $\pm$ SD      \\ \cmidrule(r){1-7} \cmidrule(l){9-11} 
        \multirow{3}{*}{NuCLS Base} & $\text{HIPT}_{256}$  & 0.654 $\pm$ 0.075    &  & 0.486 $\pm$ 0.134 & 0.641 $\pm$ 0.112 & 0.432 $\pm$ 0.195 &  & 0.822 $\pm$ 0.017 & 0.715 $\pm$ 0.026 & 0.968 $\pm$ 0.013 \\
                                    & UNI     & 0.745 $\pm$ 0.042    &  & 0.651 $\pm$ 0.075 & 0.740 $\pm$ 0.073 & 0.594 $\pm$ 0.117 &  & 0.838 $\pm$ 0.009 & 0.738 $\pm$ 0.015 & 0.971 $\pm$ 0.005 \\
                                    & SAM-H   & 0.682 $\pm$ 0.033    &  & 0.524 $\pm$ 0.060 & 0.644 $\pm$ 0.041 & 0.449 $\pm$ 0.090 &  & 0.839 $\pm$ 0.006 & 0.742 $\pm$ 0.012 & 0.967 $\pm$ 0.005 \\ \midrule
        \multirow{3}{*}{SegPath}    & $\text{HIPT}_{256}$  & 0.582                &  & 0.354             & 0.668             & 0.240             &  & 0.811             & 0.692             & 0.980             \\
                                    & UNI     & 0.723                &  & 0.606             & 0.612             & 0.600             &  & 0.841             & 0.747             & 0.961             \\
                                    & SAM-H   & 0.740                &  & 0.632             & 0.653             & 0.613             &  & 0.848             & 0.760             & 0.960             \\ \bottomrule
        \end{tabular}%
    }
\end{table*}

\begin{table*}[h]
    \centering
    \begin{minipage}[t]{0.40\linewidth}
        \caption{Summary of cell annotations in the PanopTILs dataset split by cell type and data split.}
        \label{tab:appendix_panoptils_summary}
        \resizebox{\linewidth}{!}{%
            \begin{tabular}{@{}lrrrr@{}}
            \toprule
                  & TILs    & Stromal & Epithelial & Miscellaneous \\ \midrule
            Train & 197,617 & 237,483 & 338,251    & 41,535        \\
            Test  & 14,237  & 7,986   & 16,704     & 5,946         \\ \bottomrule
            \end{tabular}
        }
    \end{minipage}%
    \hfill
    \begin{minipage}[t]{0.58\linewidth}
        \caption{PanopTILs reference results of our $\text{CellViT}^{{\scriptscriptstyle ++}}$ models, split across cell types.}
        \label{tab:}
        \resizebox{\linewidth}{!}{%
            \begin{tabular}{@{}lrrrr@{}}
            \toprule
                                        & \multicolumn{1}{l}{} & $\text{CellViT}^{{\scriptscriptstyle ++}}_{\text{HIPT}_{256}}$ & $\text{CellViT}^{{\scriptscriptstyle ++}}_\text{UNI}$       & $\text{CellViT}^{{\scriptscriptstyle ++}}_\text{SAM-H}$     \\ \midrule
            \multirow{3}{*}{TILs}       & $\text{F}_1$ $\pm$ SD          & 0.792 $\pm$ 0.005           & 0.800 $\pm$ 0.006 & 0.801 $\pm$ 0.006 \\
                                        & Precision $\pm$ SD   & 0.843 $\pm$ 0.012           & 0.845 $\pm$ 0.008 & 0.846 $\pm$ 0.010 \\
                                        & Recall $\pm$ SD      & 0.747 $\pm$ 0.019           & 0.760 $\pm$ 0.017 & 0.760 $\pm$ 0.016 \\ \midrule
            \multirow{3}{*}{Epithelial} & $\text{F}_1$ $\pm$ SD          & 0.785 $\pm$ 0.009           & 0.787 $\pm$ 0.005 & 0.800 $\pm$ 0.003 \\
                                        & Precision $\pm$ SD   & 0.858 $\pm$ 0.011           & 0.827 $\pm$ 0.007 & 0.868 $\pm$ 0.008 \\
                                        & Recall $\pm$ SD      & 0.723 $\pm$ 0.021           & 0.750 $\pm$ 0.007 & 0.741 $\pm$ 0.010 \\ \midrule
            \multirow{3}{*}{Stromal}    & $\text{F}_1$ $\pm$ SD          & 0.632 $\pm$ 0.003           & 0.634 $\pm$ 0.006 & 0.643 $\pm$ 0.006 \\
                                        & Precision $\pm$ SD   & 0.563 $\pm$ 0.009           & 0.549 $\pm$ 0.017 & 0.584 $\pm$ 0.017 \\
                                        & Recall $\pm$ SD      & 0.721 $\pm$ 0.012           & 0.751 $\pm$ 0.018 & 0.716 $\pm$ 0.017 \\ \midrule
            \multirow{3}{*}{Other}      & $\text{F}_1$ $\pm$ SD          & 0.278 $\pm$ 0.023           & 0.247 $\pm$ 0.026 & 0.242 $\pm$ 0.023 \\
                                        & Precision $\pm$ SD   & 0.545 $\pm$ 0.033           & 0.603 $\pm$ 0.036 & 0.560 $\pm$ 0.024 \\
                                        & Recall $\pm$ SD      & 0.189 $\pm$ 0.024           & 0.157 $\pm$ 0.021 & 0.155 $\pm$ 0.020 \\ \bottomrule
            \end{tabular}%
        }
    \end{minipage}
\end{table*}

\begin{table*}[]
    \centering
    \caption{MIDOG++ precision values.}
    \label{tab:appendix_midog_precision}
    \resizebox{0.70\textwidth}{!}{%
        \begin{tabular}{llrrr}
        \hline
        Models                    &                         & $\text{CellViT}^{{\scriptscriptstyle ++}}_\text{SAM-H}$           & $\text{CellViT}^{{\scriptscriptstyle ++}}_\text{SAM-H}$        & $\text{CellViT}^{{\scriptscriptstyle ++}}_\text{SAM-H}$       \\ \hline
        Organs                    & Origin                  & No Additional Cells     & 20 Additional Cells     & 200 Additional Cells    \\ \hline
        Breast Cancer             & \multirow{3}{*}{Human}  & 0.73 ($\text{SD}$ 0.03) & 0.79 ($\text{SD}$ 0.01) & 0.77 ($\text{SD}$ 0.04) \\
        Neuroendocrine Tumor      &                         & 0.47 ($\text{SD}$ 0.14) & 0.53 ($\text{SD}$ 0.03) & 0.54 ($\text{SD}$ 0.07) \\
        Melanoma                  &                         & 0.74 ($\text{SD}$ 0.21) & 0.75 ($\text{SD}$ 0.04) & 0.83 ($\text{SD}$ 0.03) \\ \hline
        Cutaneous Mast Cell Tumor & \multirow{4}{*}{Canine} & 0.65 ($\text{SD}$ 0.14) & 0.73 ($\text{SD}$ 0.05) & 0.73 ($\text{SD}$ 0.04) \\
        Lung Cancer               &                         & 0.36 ($\text{SD}$ 0.07) & 0.41 ($\text{SD}$ 0.04) & 0.42 ($\text{SD}$ 0.06) \\
        Lymphoma                  &                         & 0.61 ($\text{SD}$ 0.09) & 0.60 ($\text{SD}$ 0.05) & 0.59 ($\text{SD}$ 0.04) \\
        Soft Tissue Sarcoma       &                         & 0.71 ($\text{SD}$ 0.03) & 0.75 ($\text{SD}$ 0.02) & 0.71 ($\text{SD}$ 0.03) \\ \hline
        \end{tabular}%
    }
\end{table*}

\begin{table*}[]
    \centering
    \caption{MIDOG++ recall values.}
    \label{tab:appendix_midog_recall}
    \resizebox{0.70\textwidth}{!}{%
        \begin{tabular}{llrrr}
        \hline
        Models                    &                         & $\text{CellViT}^{{\scriptscriptstyle ++}}_\text{SAM-H}$           & $\text{CellViT}^{{\scriptscriptstyle ++}}_\text{SAM-H}$        & $\text{CellViT}^{{\scriptscriptstyle ++}}_\text{SAM-H}$       \\ \hline
        Organs                    & Origin                  & No Additional Cells     & 20 Additional Cells     & 200 Additional Cells    \\ \hline
        Breast Cancer             & \multirow{3}{*}{Human}  & 0.38 ($\text{SD}$ 0.05) & 0.42 ($\text{SD}$ 0.02) & 0.49 ($\text{SD}$ 0.03) \\
        Neuroendocrine Tumor      &                         & 0.33 ($\text{SD}$ 0.08) & 0.39 ($\text{SD}$ 0.02) & 0.48 ($\text{SD}$ 0.06) \\
        Melanoma                  &                         & 0.56 ($\text{SD}$ 0.10) & 0.60 ($\text{SD}$ 0.03) & 0.62 ($\text{SD}$ 0.03) \\ \hline
        Cutaneous Mast Cell Tumor & \multirow{4}{*}{Canine} & 0.63 ($\text{SD}$ 0.07) & 0.61 ($\text{SD}$ 0.03) & 0.68 ($\text{SD}$ 0.03) \\
        Lung Cancer               &                         & 0.34 ($\text{SD}$ 0.06) & 0.41 ($\text{SD}$ 0.03) & 0.44 ($\text{SD}$ 0.02) \\
        Lymphoma                  &                         & 0.39 ($\text{SD}$ 0.05) & 0.44 ($\text{SD}$ 0.03) & 0.57 ($\text{SD}$ 0.05) \\
        Soft Tissue Sarcoma       &                         & 0.43 ($\text{SD}$ 0.06) & 0.42 ($\text{SD}$ 0.01) & 0.48 ($\text{SD}$ 0.04) \\ \hline
        \end{tabular}%
    }
\end{table*}

\begin{table*}[h]
    \centering
    \begin{minipage}[t]{0.25\linewidth}
        \caption{List of classical machine learning models evaluated using the PyCaret automated machine learning (AutoML) framework for the Lizard dataset. CatBoost classifier (marked bold) was the best performing model among all tested.}
        \label{tab:appendix_pycaret_models}
        \resizebox{\linewidth}{!}{%
            \begin{tabular}{l|l} 
            \toprule
            \multicolumn{1}{l}{} & Model \\ 
            \hline
            1 & Logistic Regression \\
            2 & Linear Discriminant Analysis \\
            3 & Ridge Classifier \\
            4 & Ada Boost Classifier \\
            5 & Gradient Boosting Classifier \\
            \textbf{6} & \textbf{CatBoost Classifier} \\
            7 & Extreme Gradient Boosting \\
            8 & Light Gradient Boosting Machine \\
            9 & Quadratic Discriminant Analysis \\
            10 & Random Forest Classifier \\
            11 & Decision Tree Classifier \\
            12 & Extra Trees Classifier \\
            13 & K Neighbors Classifier \\
            14 & Naive Bayes \\
            15 & SVM - Linear Kernel \\
            \bottomrule
            \end{tabular}
        }
    \end{minipage}%
    \hfill
    \begin{minipage}[t]{0.73\linewidth}
        \caption{Overview of augmentation techniques used in the study. This table lists the augmentation techniques along with their corresponding Albumentations function names, the probability of application, and relevant parameters for each technique.}
        \label{tab:appendix_data_augmentation}
        \resizebox{\linewidth}{!}{%
            \begin{tabular}{@{}llll@{}}
            \toprule
            \textbf{Augmentation Technique} & \textbf{Albumentations Function Name} & \textbf{Probability (p)} & \textbf{Parameters}                                              \\ \midrule
            Random Rotate 90                & RandomRotate90                        & 0.5                      & None                                                             \\
            Horizontal Flip                 & HorizontalFlip                        & 0.5                      & None                                                             \\
            Vertical Flip                   & VerticalFlip                          & 0.5                      & None                                                             \\
            Downscale                       & Downscale                             & 0.15                     & scale\_max = 0.5, scale\_min = 0.5                               \\
            Blur                            & Blur                                  & 0.2                      & blur\_limit = 10                                                 \\
            Gaussian Noise                  & GaussNoise                            & 0.25                     & var\_limit = 50                                                  \\
            Color Jitter                    & ColorJitter                           & 0.2                      & brightness = 0.25, contrast = 0.25, saturation = 0.1, hue = 0.05 \\
            Superpixels                     & Superpixels                           & 0.1                      & p\_replace = 0.1, n\_segments = 200, max\_size = h/2             \\
            Zoom Blur                       & ZoomBlur                              & 0.1                      & max\_factor = 1.05                                               \\
            Random Sized Crop               & RandomSizedCrop                       & 0.1                      & min\_max\_height = (h/2, h), height = h, width = w               \\ \bottomrule
            \end{tabular}
        }
    \end{minipage}
\end{table*}

\begin{table*}[]
    \centering
    \caption{Summary of the cell datasets used in this study. 
    }
    \label{tab:appendix_cell_dataset_overview}
    \resizebox{\textwidth}{!}{%
        \begin{tabular}{|l|l|l|l|l|l|l|l|l|}
        \hline
        \textbf{Dataset}   & \textbf{Cell Classes}                                                                                                                                                                                                    & \textbf{Organs}                                                                                                                                                                                                                                 & \textbf{Nuclei Amount} & \textbf{Patches/Slide Amount}                                                                                                                                                            & \textbf{Source}                                                                                                                & \textbf{\begin{tabular}[c]{@{}l@{}}Resolution/\\ Magnification\end{tabular}} & \textbf{\begin{tabular}[c]{@{}l@{}}Seg. \\ Mask\end{tabular}} & \textbf{Note}                                                                                                                                                                                                                                                                                 \\ \hline
        \textbf{Ocelot}    & \begin{tabular}[c]{@{}l@{}}Tumor, \\ Non-tumor\end{tabular}                                                                                                                                                              & \begin{tabular}[c]{@{}l@{}}Kidney, \\ Head/Neck, \\ Prostate, \\ Stomach, \\ Endometrium, \\ Bladder\end{tabular}                                                                                                                               & 113,026                & \begin{tabular}[c]{@{}l@{}}Total 303 Slides \\ (173 train, 65 val, 65 test), \\ cut into 663 tiles\\ (400 train, 137 val,126 test)\\ with size 1024$\times$1024\end{tabular}                    & TCGA                                                                                                                           & $0.25~\si{\text{\textmu} m \per px}$ / $\times$40                                                        & No                                                            & \begin{tabular}[c]{@{}l@{}}The authors \\ removed 4 test tiles\\ due to missing annotations\end{tabular}                                                                                                                                                                                      \\ \hline
        \textbf{MIDOG++}   & \begin{tabular}[c]{@{}l@{}}Mitotic figures, \\ Non-mitotic figures\end{tabular}                                                                                                                                          & \begin{tabular}[c]{@{}l@{}}Breast (Human), \\ Neuroendocrine Tumor (Human), \\ Melanoma (Human) \\ Cutaneous Mast Cell Tumor (Canine), \\ Neuroendocrine Tumor (Canine), \\ Lymphoma (Canine), \\ Soft Tissue Sarcoma (Canine) \end{tabular}                                                                                                                                                                                                                                               & \begin{tabular}[c]{@{}l@{}}26,289 annotated, \\  7,398,795 total\end{tabular}                      & \begin{tabular}[c]{@{}l@{}}503 images with various sizes, \\  average of 6,804$\times$5,102 pixels\end{tabular}                                                                                                                                                                                        & \begin{tabular}[c]{@{}l@{}} UMC Utrecht, \\ VMU Vienna, \\ FU Berlin, \\ AMC New York \end{tabular}                                                                                                                                 & $0.23~\si{\text{\textmu} m \per px}$ - $0.25~\si{\text{\textmu} m \per px}$                                                                    & No                                                             & -                                                                                                                                                                                                                                                                                             \\ \hline
        \textbf{CoNSeP}    & \begin{tabular}[c]{@{}l@{}}Epithelial, \\ Inflammatory, \\ Spindle-shaped, \\ Miscellaneous\end{tabular}                                                                                                                 & Colon                                                                                                                                                                                                                                           & 24,332                 & \begin{tabular}[c]{@{}l@{}}41 patches (27 train, 14 test)\\ with size $num{1000}\times num{1000}$\end{tabular}                                                                                             & \begin{tabular}[c]{@{}l@{}}University \\ Hospitals \\ Coventry \\ and \\ Warwickshire (UK)\end{tabular}                        & $0.25~\si{\text{\textmu} m \per px}$ / $\times$40                                                        & Yes                                                           & Resized to $num{1024}\times num{1024}$                                                                                                                                                                                                                                                                          \\ \hline
        \textbf{Lizard}    & \begin{tabular}[c]{@{}l@{}}Neutrophils, \\ Lymphocytes, \\ Plasma, \\ Eosinophils, \\ Epithelial, \\ Connective\end{tabular}                                                                                             & Colon                                                                                                                                                                                                                                           & 418,935                & \begin{tabular}[c]{@{}l@{}}270 tiles of various sizes, \\ 1016$\times$917 pixels on average at $0.50~\si{\text{\textmu} m \per px}$\end{tabular}                                                                             & \begin{tabular}[c]{@{}l@{}}University \\ Hospitals \\ Coventry \\ and \\ Warwickshire (UK)\\ TCGA,\\ China (Four)\end{tabular} & $0.50~\si{\text{\textmu} m \per px}$ / $\times$20                                                        & Yes                                                           & \begin{tabular}[c]{@{}l@{}}Resized from 0.5 to 0.25 \\ with Lanczos filter\\ at input and back to $0.50~\si{\text{\textmu} m \per px}$ \\ at the output\end{tabular}                                                                                                                                                      \\ \hline
        \textbf{NuCLS}     & \begin{tabular}[c]{@{}l@{}}Lymphocytes and Plasma (superclass: sTILs), \\ Macrophages and Stromal (superclass: stromal cells), \\ Mitotic and Non-Mitotic Tumor (superclass: tumor cells),\\ Miscellaneous\end{tabular}  & Breast                                                                                                                                                                                                                                          & 48,365                 & \begin{tabular}[c]{@{}l@{}}109 train and 15 test WSI with 1-3 annotated crops, \\ average crop size of 362 $\times$ 362 pixels, \\ average FOV size (annotated area) of 320 pixels\end{tabular} & TCGA                                                                                                                           & $0.20~\si{\text{\textmu} m \per px}$ / $\times$40                                                        & No                                                            & \begin{tabular}[c]{@{}l@{}}Cropped just FOV area \\ and resize to 256$\times$256 pixels \\ to achieve $0.25~\si{\text{\textmu} m \per px}$\end{tabular}                                                                                                                                                                          \\ \hline
        \textbf{PanopTILs} & \begin{tabular}[c]{@{}l@{}}TILs, \\ Stromal, \\ Epithelial,\\ Miscellaneous\end{tabular}                                                                                                                                 & Breast                                                                                                                                                                                                                                          & 859,759                & \begin{tabular}[c]{@{}l@{}}1709 train and 1317 test tiles, \\ with size 1024$\times$1024,\\ but test tiles just annotated in a narrow FOV\end{tabular}                                          & TCGA                                                                                                                           & $0.25~\si{\text{\textmu} m \per px}$ / $\times$40                                                        & No                                                            & -                                                                                                                                                                                                                                                                                             \\ \hline
        \textbf{Segpath}   & \begin{tabular}[c]{@{}l@{}}IF stainings of: \\ Epithelial, \\ Smooth muscle/Myofibroblasts, \\ Lymphocytes, \\ Leukocytes, \\ Blood/lymphatic vessel, \\ Plasma cells, \\ Myeloid cells, \\ Red blood cells\end{tabular} & \begin{tabular}[c]{@{}l@{}}Bladder, \\ Brain, \\ Breast, \\ Colon, \\ Head/Neck, \\ Kidney, \\ Liver, \\ Lung, \\ Oesophagus, \\ Ovary, \\ Pancreas, \\ Prostate, \\ Sarcoma, \\ Skin, \\ Stomach, \\ Testis, \\ Thymus, \\ Uterus\end{tabular} & -                      & \begin{tabular}[c]{@{}l@{}}220 breast tiles for lymphocytes, \\ 2054 for plasma cells\\ with size $984\times 984$\end{tabular}                                                                   & \begin{tabular}[c]{@{}l@{}}University \\ of Tokyo \\ Hospital\end{tabular}                                                     & $0.22~\si{\text{\textmu} m \per px}$ / $\times 40$                                                        & -                                                             & \begin{tabular}[c]{@{}l@{}}No nuclei have been annotated. \\ The dataset consists of registered \\ HE and IHC stainings \\ to automatically derive \\ region-wise annotations. \\ We extract nuclei annotations \\ by transferring \\ binary $\text{CellViT}^{{\scriptscriptstyle ++}}$ results\\  to the IHC mask.\end{tabular} \\ \hline
        \end{tabular}
    }
\end{table*}

\begin{table*}[]
\centering
\caption{Resources used in this study with identifier}
\label{tab:appendix_identifier_overview}
\resizebox{\textwidth}{!}{%
\begin{tabular}{@{}lllr@{}}
\toprule
Resource                        & Source                      & Identifier                                                                                                                                          & Reference                     \\ \midrule
$\text{CellViT}^{{\scriptscriptstyle ++}}$                       &                             &                                                                                                                                                     & -                             \\ \midrule
Albumentations 1.3.0            & Pip                         & \href{https://albumentations.ai/}{albumentations.ai}                                                                                                                          &~\citet{albumentations}        \\
CatBoost 1.2.5                  & Pip                         & \href{https://github.com/catboost/catboost}{github.com/catboost/catboost}                                                                                                                &~\citet{catboost}              \\
cuCIM 24.04.00                  & Rapidsai (conda channel)    & \href{https://github.com/rapidsai/cucim}{github.com/rapidsai/cucim}                                                                                                                   & -                             \\
CuPY 13.2.0                     & Conda-forge (conda channel) & \href{https://cupy.dev/}{cupy.dev}                                                                                                                                   &~\citet{cupy}                  \\
GeoJSON 3.0.0                   & Pip                         & \href{https://python-geojson.readthedocs.io/en/latest/}{python-geojson.readthedocs.io/en/latest}                                                                                                    & -                             \\
huggingface-hub 0.22.2          & Pip                         & \href{https://huggingface.co/}{huggingface.co}                                                                                                                             & -                             \\
Numba 0.59.0                    & Pip                         & \href{https://numba.pydata.org/}{numba.pydata.org}                                                                                                                           &~\citet{numba}                 \\
NumPy 1.23.5                    & Pip                         & \href{https://numpy.org/}{numpy.org}                                                                                                                                  &~\citet{numpy}                 \\
OpenSlide 4.0.0                 & Conda-forge (conda channel) & \href{https://openslide.org/}{openslide.org}                                                                                                                              &~\citet{openslide}             \\
openslide-python 1.3.1          & Pip                         & \href{https://openslide.org/api/python/}{openslide.org/api/python}                                                                                                                  & -                             \\
opencv-python-headless 4.5.4.58 & Pip                         & \href{https://opencv.org/}{opencv.org}                                                                                                                                 &~\citet{opencv}                \\
pandarallel 1.6.5               & Pip                         & \href{https://github.com/nalepae/pandarallel}{github.com/nalepae/pandarallel}                                                                                                              & -                             \\
pandas 1.4.3                    & Pip                         & \href{https://pandas.pydata.org/}{pandas.pydata.org}                                                                                                                          &~\citet{pandas}                \\
PathoPatch 1.0.4b0              & Pip                         & \href{https://github.com/TIO-IKIM/PathoPatcher}{github.com/TIO-IKIM/PathoPatcher}                                                                                                           &~\citet{pathopatcher}          \\
Pillow 10.3.0                   & Conda-forge (conda channel) & \href{https://pillow.readthedocs.io/}{pillow.readthedocs.io}                                                                                                                      & -                             \\
PyCaret 3.3.2                   & Pip                         & \href{https://pycaret.org/}{pycaret.org}                                                                                                                                &~\citet{pycaret}               \\
Python 3.10.14                  & Conda-forge (conda channel) & \href{https://www.python.org/}{www.python.org}                                                                                                                             & -                             \\
Ray 2.9.3                       & Pip                         & \href{https://www.ray.io/}{www.ray.io}                                                                                                                                 &~\citet{ray}                   \\
scikit-base 0.7.8               & Pip                         & \href{https://scikit-learn.org/}{scikit-learn.org}                                                                                                                           &~\citet{scikit-learn}          \\
scikit-image 0.19.8             & Pip                         & \href{https://scikit-image.org/}{scikit-image.org}                                                                                                                           &~\citet{scikit-image}          \\
scikit-learn 1.3.0              & Pip                         & \href{https://scikit-learn.org/}{scikit-learn.org}                                                                                                                           &~\citet{scikit-learn}          \\
scipy 1.8.1                     & Pip                         & \href{https://scipy.org/}{scipy.org}                                                                                                                                  &~\citet{scipy}                 \\
timm 1.0.8                      & Pip                         & \href{https://timm.fast.ai/}{timm.fast.ai}                                                                                                                               &~\citet{timm}                  \\
torch 2.2.1                     & Pip                         & \href{https://pytorch.org/}{pytorch.org}                                                                                                                                &~\citet{pytorch}               \\
torchmetrics 0.11.4             & Pip                         & \href{https://lightning.ai/docs/torchmetrics}{lightning.ai/docs/torchmetrics}                                                                                                              &~\citet{torchmetrics}          \\
torchvision 0.17.1              & Pip                         & \href{https://pytorch.org/}{/pytorch.org}                                                                                                                                & -                             \\
ujson 5.8.0                     & Pip                         & \href{https://pypi.org/project/ujson/}{pypi.org/project/ujson}                                                                                                                     & -                             \\
WandB 0.15.4                    & Pip                         & \href{https://wandb.ai/}{wandb.ai}                                                                                                                                   & -                             \\
Wsidicom 0.20.4                 & Pip                         & \href{https://github.com/imi-bigpicture/wsidicom}{github.com/imi-bigpicture/wsidicom}                                                                                                          & -                             \\
Wsidicomizer 0.13.2             & Pip                         & \href{https://github.com/imi-bigpicture/wsidicomizer}{github.com/imi-bigpicture/wsidicomizer}                                                                                                      & -                             \\
XGBoost 2.1.1                   & Pip                         & \href{https://xgboost.readthedocs.io/}{xgboost.readthedocs.io}                                                                                                                     &~\citet{xgboost}               \\ \midrule
Comparison Methods              &                             &                                                                                                                                                     & -                             \\ \midrule
SoftCTM                         & GitHub                      & \href{https://github.com/lely475/SoftCTM}{github.com/lely475/SoftCTM} (commit 8918beafd7d5a36695d1bbdb5bb8d6139376a4dc)                                                                &~\citet{softctm}               \\
HoVer-Net                       & GitHub                      & \href{https://github.com/vqdang/hover\_net}{github.com/vqdang/hover\_net} (commit 67e2ce5e3f1a64a2ece77ad1c24233653a9e0901)                                                              &~\citet{hovernet}              \\
Cerberus                        & GitHub                      & \href{https://github.com/TissueImageAnalytics/cerberus}{github.com/TissueImageAnalytics/cerberus} (commit 5bcecbb071bebd5911250034c94f3568f23f50bb)                                                  &~\citet{cerberus}              \\
TIAToolBox                      & GitHub                      & \href{https://github.com/TissueImageAnalytics/tiatoolbox}{github.com/TissueImageAnalytics/tiatoolbox} (commit c180566bbe7ec04a9b91924748acf2d03f6302d9)                                                &~\citet{tiatoolbox}            \\
PointNu-Net                     & GitHub                      & \href{https://github.com/Kaiseem/PointNu-Net}{github.com/Kaiseem/PointNu-Net} (commit 747f5019df5f611e81a823e5318a2fa0b60e2571)                                                            &~\citet{pointnunet}            \\
Nucleiio                        & GitHub                      & \href{https://github.com/huangzhii/nuclei.io}{github.com/huangzhii/nuclei.io} (commit 78d52270eaeb05bc26f9b134231431a04a837b22)                                                            &~\citet{nucleiio}              \\
CGIS-CPF                        &                             &                                                                                                                                                     & -                             \\ \midrule
Foundation Models               &                             &                                                                                                                                                     & -                             \\ \midrule
HIPT           & GitHub                      & \href{https://github.com/mahmoodlab/HIPT}{github.com/mahmoodlab/HIPT} (commit 7336ee7)                                                                                                                 &~\citet{hipt}                  \\
UNI                             & Hugging Face                & \href{https://huggingface.co/MahmoodLab/UNI}{huggingface.co/MahmoodLab/UNI} (commit ba5018a94088b378720cd07995efe65a79c6b952)                                                             &~\citet{uni}                   \\
Virchow                         & Hugging Face                & \href{https://huggingface.co/paige-ai/Virchow}{huggingface.co/paige-ai/Virchow} (commit b80411ffe80f1d3070879e512ffb0152d7997377)                                                           &~\citet{virchow}               \\
Virchow2                        & Hugging Face                & \href{https://huggingface.co/paige-ai/Virchow2}{huggingface.co/paige-ai/Virchow2} (commit a8536e8a8dd3cd0b200aa44be674ef95d2ad1598)                                                          &~\citet{virchow2}              \\
Segment Anything                & GitHub                      & \href{https://github.com/facebookresearch/segment-anything}{github.com/facebookresearch/segment-anything}                                                                                                & -                             \\ \midrule
Datasets                        &                             &                                                                                                                                                     & -                             \\ \midrule
Ocelot                          & Zenodo                      & \href{https://zenodo.org/records/8417503}{zenodo.org/records/8417503}                                                                                                                  &~\citet{ocelot}                \\
Midog++                         & Figshare                    & \href{https://doi.org/10.6084/m9.figshare.c.6615571.v1}{doi.org/10.6084/m9.figshare.c.6615571.v1}                                                                                                    &~\citet{midogpp}               \\
CoNSeP                          & HoVer-Net                   & \href{https://warwick.ac.uk/fac/cross\_fac/tia/data/hovernet/}{warwick.ac.uk/fac/cross\_fac/tia/data/hovernet}                                                                                             &~\citet{hovernet}              \\
Lizard                          & -                           & \href{https://warwick.ac.uk/fac/cross\_fac/tia/data/lizard}{warwick.ac.uk/fac/cross\_fac/tia/data/lizard}                                                                                                &~\citet{lizard}                \\
NuCLS                           & -                           & \href{https://sites.google.com/view/nucls/home}{sites.google.com/view/nucls/home}                                                                                                            &~\citet{nucls}                 \\
PanopTILs                       & -                           & \href{https://sites.google.com/view/panoptils/}{sites.google.com/view/panoptils}                                                                                                            &~\citet{panoptils}             \\
SegPath                         & -                           & \begin{tabular}[c]{@{}l@{}}\href{https://dakomura.github.io/SegPath/}{dakomura.github.io/SegPath/}\\ \href{https://zenodo.org/record/7412529}{zenodo.org/record/7412529}\\ \href{https://zenodo.org/record/7412500}{zenodo.org/record/7412500}\end{tabular} &~\citet{segpath}               \\ \bottomrule
\end{tabular}%
}
\end{table*}

\end{document}